\documentclass[10pt,twocolumn,letterpaper]{article}

\usepackage[accsupp]{axessibility}
\usepackage[]{cvpr} 
\usepackage{graphicx}
\usepackage{amsmath}
\usepackage{booktabs}

\usepackage{gensymb} 
\usepackage{multirow}
\usepackage{xcolor}
\usepackage{graphicx}
\usepackage{subcaption}
\usepackage{arydshln}
\usepackage[pagebackref=true,breaklinks=true,colorlinks,bookmarks=false]{hyperref}
\usepackage{nicefrac}  
\usepackage{wasysym}
\usepackage{amssymb}

\usepackage[capitalize]{cleveref}
\crefname{section}{Sec.}{Secs.}
\Crefname{section}{Section}{Sections}
\Crefname{table}{Table}{Tables}
\crefname{table}{Tab.}{Tabs.}

\def\eg{\emph{e.g.}}
\def\etal{\emph{et al.}}

\begin{document}

\title{InfoNeRF: Ray Entropy Minimization for Few-Shot Neural Volume Rendering}

\author{
Mijeong Kim$^1$ \qquad\qquad Seonguk Seo$^1$ \qquad\qquad Bohyung Han$^{1,2}$ \\
$^1$ECE \& $^1$ASRI \& $^{1,2}$IPAI, Seoul National University\\
 {\tt\small \{mijeong.kim, seonguk, bhhan\}@snu.ac.kr}
}
\twocolumn[{
\maketitle
\centering
\includegraphics[width=0.98\linewidth]{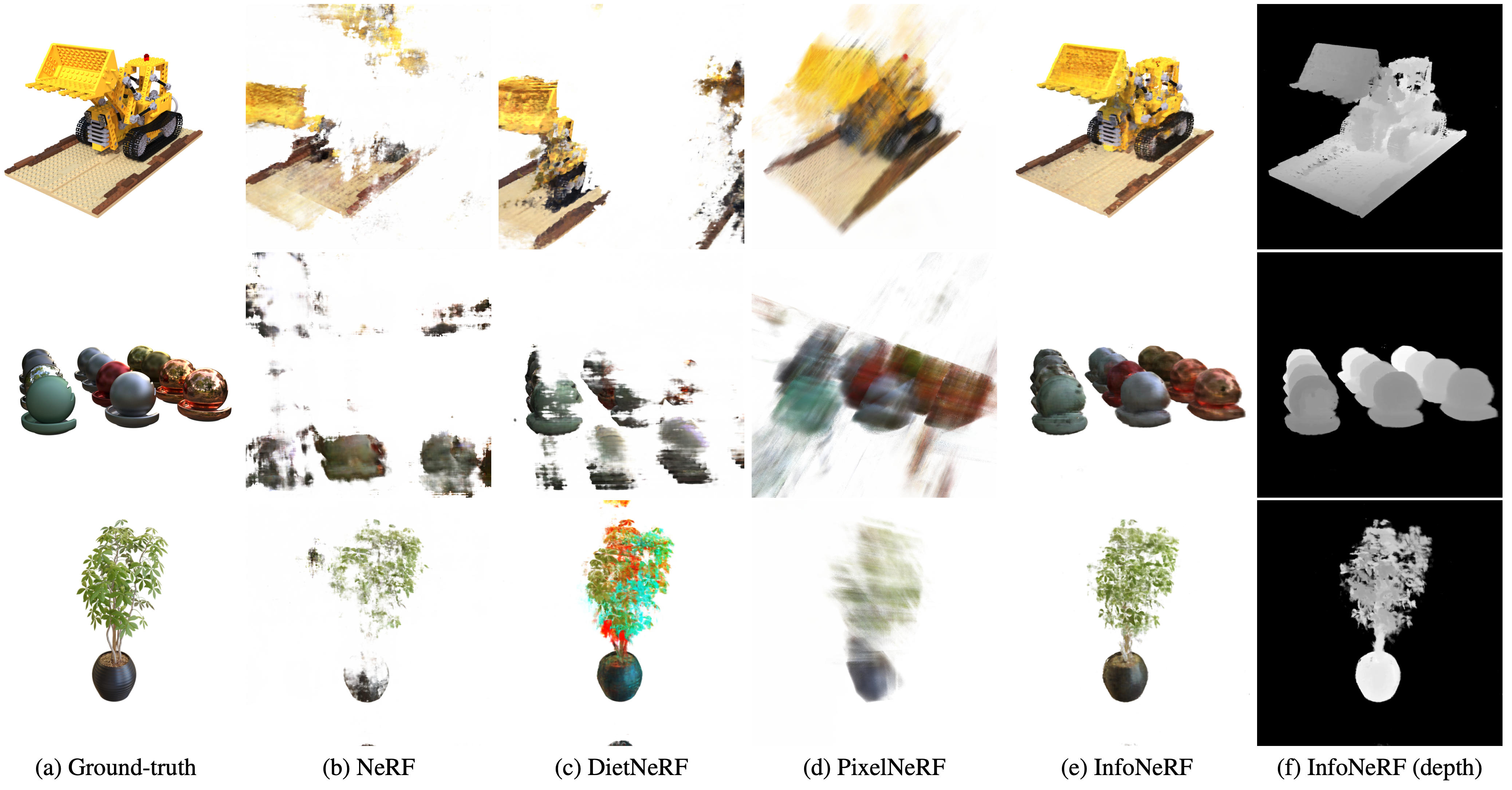}
\vspace{-1mm}
\captionof{figure}
{
Qualitative comparison of InfoNeRF (ours) with other NeRF-based models~\cite{mildenhall2020nerf, jain2021putting, yu2021pixelnerf} on the {\textit{Lego}, \textit{Materials}, and \textit{Ficus}} scenes of Realistic Synthetic 360$^{\circ}$ dataset in 4-view settings.
Existing works often suffer from noise (b), color distortion (c), or blur effect (d), while InfoNeRF (ours) achieves outstanding quality of rendered images with only a few input views.
The last column (f) visualizes depth maps estimated by InfoNeRF (ours), which provide clear boundaries and fine details of the objects.
}
\label{fig:teaser}
\vspace{0.5cm}
}]


\begin{abstract}
We present an information-theoretic regularization technique for few-shot novel view synthesis based on neural implicit representation.
The proposed approach minimizes potential reconstruction inconsistency that happens due to insufficient viewpoints by imposing the entropy constraint of the density in each ray.
In addition, to alleviate the potential degenerate issue when all training images are acquired from almost redundant viewpoints, we further incorporate the spatial smoothness constraint into the estimated images by restricting information gains from additional rays with slightly different viewpoints.
The main idea of our algorithm is to make reconstructed scenes compact along individual rays and consistent across rays in the neighborhood.
The proposed regularizers can be plugged into most of existing neural volume rendering techniques based on NeRF in a straightforward way.
Despite its simplicity, we achieve consistently improved performance compared to existing neural view synthesis methods by large margins on multiple standard benchmarks.
Our codes and models are available in the project website\footnote{\url{http://cvlab.snu.ac.kr/research/InfoNeRF}}.
\end{abstract}



\section{Introduction}
\label{sec:introduction}

Understanding 3D structure of a natural scene is a critical step for various high-level computer vision applications including object recognition, photorealistic rendering, autonomous driving, virtual reality, and many others.
Recent advance of deep learning capacitates high-fidelity 3D reconstruction and recognition, but learning with 3D data is inherently more difficult than its counterpart based on 2D images due to unstructured nature of data format, high memory requirement, and lack of principled architectures.
Hence, many researchers investigate the standard models with appropriate training algorithms and the methods for reducing their computational costs, and attempt to solve various challenging tasks.

Novel view synthesis based on neural implicit representations is one of the 3D learning tasks that draws a lot of attention these days since Neural Radiance Field (NeRF)~\cite{mildenhall2020nerf} has been introduced.
NeRF delivers accurate 3D reconstruction results without explicit modeling of 3D scene structures, but the requirement of many images captured from multiple calibrated cameras hampers the applicability of the method.
Therefore, several recent approaches aim to reduce the high computational cost and alleviate the constraints related to datasets~\cite{deng2021depth, jain2021putting, yu2021pixelnerf}.

In this line of research, we explore the few-shot prior-free novel view synthesis task, where only a limited number of training images are accessible and other prior information, such as object categories and semantic structures of target scenes, are unavailable.
There exist several prior works for this task, but they either work barely with few examples~\cite{jain2021putting} or require narrow baseline assumption to find correspondences using an external module~\cite{deng2021depth}.
Other approaches rely on prior knowledge of scenes such as object classes or features.
For example, PixelNeRF~\cite{yu2021pixelnerf} takes advantage of the features extracted from seen images to compensate for missing information in unseen views while \cite{peng2021neural, kwon2021neural} focus on a particular object class, \eg, human, in novel view synthesis.

We address the fundamental drawbacks of existing few-shot novel view synthesis methods: inconsistent reconstruction, which generates noise, blur, or artifacts in rendered images, and overfitting to seen views, which leads to degenerate or trivial solutions.
The proposed approach, referred to as InfoNeRF, alleviates the reconstruction inconsistency by imposing the sparsity on the estimated scene, which is achieved by entropy minimization in each ray.
The overfitting issues are handled by enforcing the smoothness of the reconstruction with respect to viewpoint changes, which is controlled by minimizing information gains from a pair of slightly different viewpoints.
Figure~\ref{fig:teaser} illustrates the outstanding quality of rendered images and depth maps estimated by our model, which delineates clear object boundaries and fine structures using only 4 input views with wide baselines.

Overall, the main contributions and benefits of our algorithm are summarized as follows:
\begin{itemize}
\item We propose a novel information-theoretic approach, InfoNeRF, for the regularization of the neural implicit representations for volume rendering.
Our method points out key drawbacks of the existing few-shot novel view synthesis techniques, and introduces two effective regularization schemes, ray entropy minimization and ray information gain reduction. 

\item Since InfoNeRF is a generic regularization technique and does not require any other external data structures, \eg, voxels or meshes, or additional learnable parameters, it can be applied to various neural volume rendering algorithms with and without scene prior.

\item The proposed regularization technique turns out to be effective to alleviate reconstruction inconsistency across multiple views and prevent degenerate solutions by overfitting despite its simplicity.
We demonstrate outstanding performance of InfoNeRF on several standard benchmarks for few-shot novel view synthesis.

\item To our knowledge, InfoNeRF is the first NeRF variant that performs few-shot novel view synthesis on wide-baseline image datasets without prior information.
\end{itemize}



\section{Related Work}
\label{sec:related}

\subsection{Novel View Synthesis}
Novel view synthesis aims to render realistic images via geometric and photometric understanding of a 3D scene given a set of training images.
To address this problem, light fields~\cite{levoy1996light,srinivasan2017learning,wood2000surface} or image-based rendering~\cite{debevec1996modeling, buehler2001unstructured,chaurasia2011silhouette,sinha2009piecewise,chaurasia2013depth} approaches have been employed traditionally, and the approaches based on deep learning~\cite{zhou2018stereo,zhou2016view,flynn2019deepview,flynn2016deepstereo, mildenhall2019local} have recently received growing attention.
In particular, NeRF~\cite{mildenhall2020nerf} achieves photo-realistic rendering results by applying multi-layer perceptrons to differentiable volume rendering successfully.
The following works attempt to extend NeRF in various aspects, such as dynamic view synthesis~\cite{li2021neural}, self-calibrated view synthesis~\cite{wang2021nerf}, real-time rendering~\cite{yu2021plenoctrees, wizadwongsa2021nex, neff2021donerf}, relighting~\cite{srinivasan2021nerv}, and anti-aliasing~\cite{barron2021mip}.
Although NeRF-based models have achieved impressive performance, they have a common drawback; they require dense scene sampling, making them difficult to be applied in real-world scenarios.
We address the few-shot volume rendering task to improve applicability by reducing the need for  many images captured by calibrated cameras.

\subsection{Few-shot Novel View Synthesis}

To synthesize novel views of a scene given sparse observations, some algorithms estimate depth maps from images since depth is valuable source for view synthesis and 3D reconstruction.
Depth information plays a crucial role in depth-guided image interpolation~\cite{flynn2016deepstereo, saito2019pifu}, multi-plane image prediction~\cite{tucker2020single}, and learned geometry regularization in complicated scenes~\cite{deng2021depth} for few-shot view synthesis.
However, these methods require training images with depth supervision or external depth estimation modules, \eg, multiview stereo or COLMAP SfM~\cite{schonberger2016structure}, and are susceptible to large projection errors due to incorrect depth predictions.

To overcome the limitation, 
several approaches exploit multi-view feature semantics by introducing an image encoder for NeRF to estimate color and opacity~\cite{yu2021pixelnerf, peng2021neural, kwon2021neural} or achieve semantic consistency between seen images and rendered novel views~\cite{jain2021putting}.
These strategies facilitate learning the semantic prior, and allow us to synthesize novel views with only few-shot images.
Unlike the aforementioned works, the proposed algorithm does not rely on any prior information or additional pretrained encoders.
Our regularization technique is orthogonal to the other methods discussed above and can be integrated into existing few-shot volume rendering approaches straightforwardly.

Some explicit representation methods~\cite{yu2021plenoctrees, lombardi2019neural} incorporate sparsity constraints for neural volume rendering, which may also be useful for few-shot novel-view synthesis, although they do not directly address the task.
Yu~\etal~\cite{yu2021plenoctrees} adopt a variation of the octree structure with a sparsity prior loss to remove a subset of nodes in the tree corresponding to unobserved regions.
On the other hand, Lombardi~\etal~\cite{lombardi2019neural} conduct ray marching through voxel grids for volume rendering and regularize the total variation of voxel opacities by enforcing sparse spatial gradient.
Our method also employ sparsity via entropy constraints, but is more generic than the explicit methods because our algorithm does not require external data structures and suffer from memory bound to store explicit representations.


\section{Preliminaries: NeRF}
\label{sec:preliminaries}
\begin{figure*}[t]
     \centering
     \begin{subfigure}[h]{0.415\linewidth}
         \centering
         \includegraphics[width=0.49\linewidth]{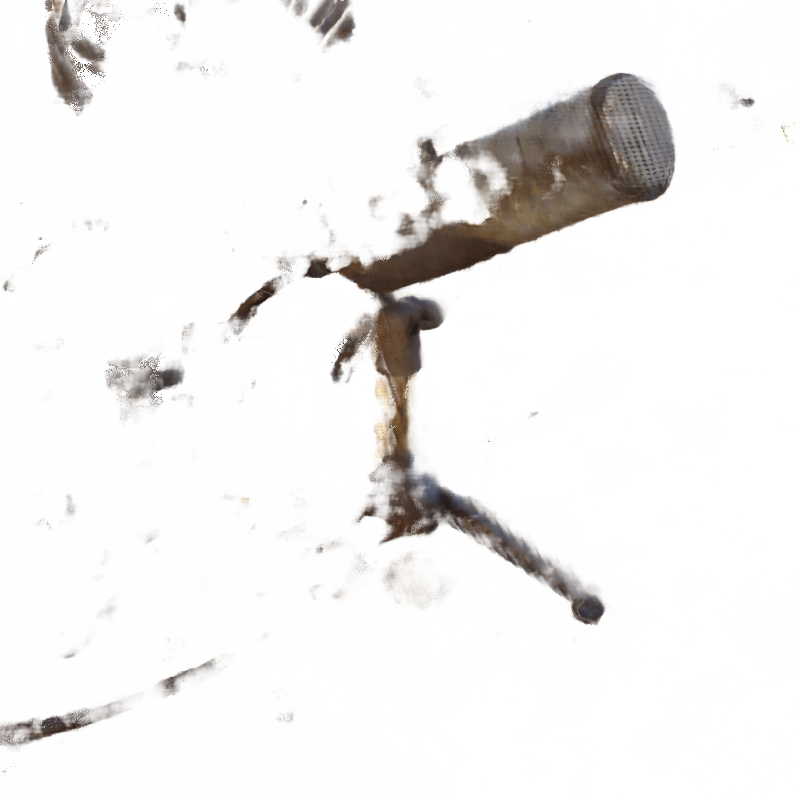} 
         \includegraphics[width=0.49\linewidth]{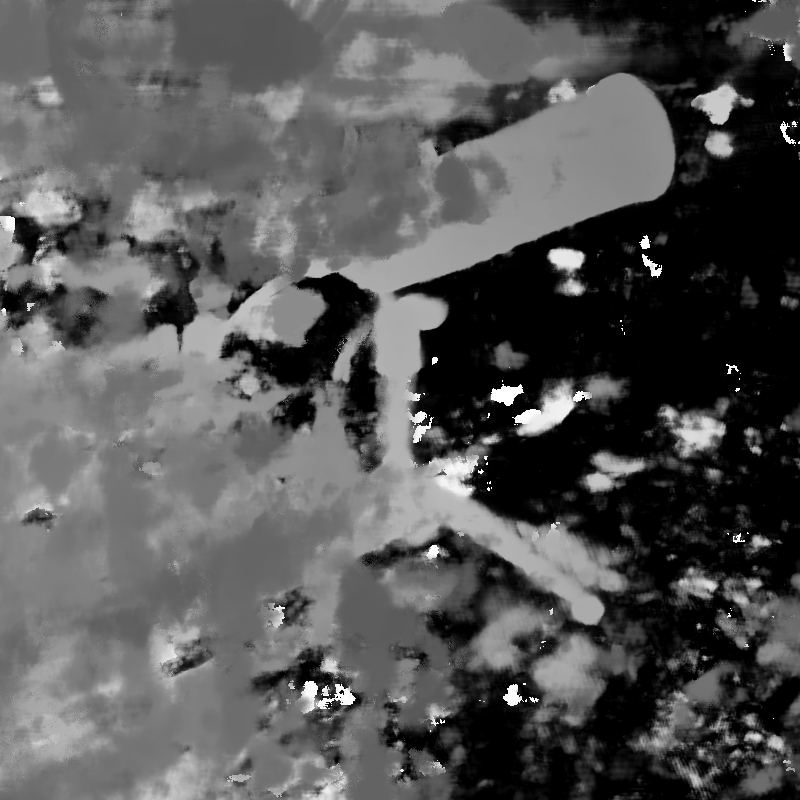}
        \caption{Rendered RGB and depth images estimated by NeRF~\cite{mildenhall2020nerf} on the Realistic Synthetic 360$^{\circ}$ dataset in a 4-view setting.}
        \label{fig:nerf_noise}
     \end{subfigure}
     \hspace{0.3cm}
    \begin{subfigure}[h]{0.555\linewidth}
         \centering
         \includegraphics[width=0.49\linewidth]{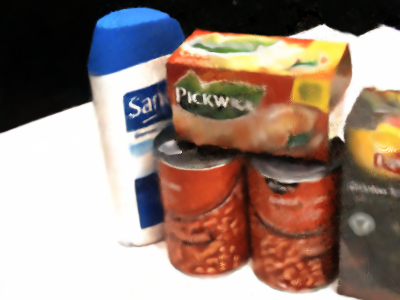}
        \includegraphics[width=0.49\linewidth]{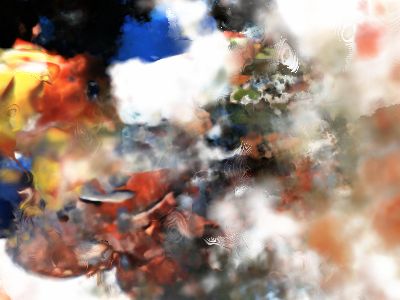}
         \caption{Rendered RGB images for two slightly different viewpoints (left: seen, right: unseen) estimated by NeRF~\cite{mildenhall2020nerf} on the DTU dataset in a 3-view setting.}
        \label{fig:nerf_overfit}
    \end{subfigure}
    \vspace{-1mm}
    \caption{Typical drawbacks of NeRF in the few-shot volume rendering.
    We are motivated by the following two observations and alleviate the limitations by adding two regularizers in this work.
    (a) Few-shot NeRF suffers from significant noise in 3D depth estimation, which leads to noisy rendering.
    (b) Few-shot NeRF provides completely different rendering results between slightly different viewpoints because it extremely overfits to seen input views.
    }
    \vspace{-2mm}
    \label{fig:motivation}
\end{figure*}
Neural Radiance Fields (NeRF)~\cite{mildenhall2020nerf} is a novel framework to represent a 3D scene with a neural implicit function, where a neural network $f(\cdot, \cdot)$, typically given by an MLP, maps a 3D point $\textbf{x}=(x,y,z)$ and a unit viewing direction $\textbf{d}=(\theta, \phi)$ to a volume density $\sigma$ as well as an emitted RGB color $\textbf{c} = (r,g,b)$.
Following the classical volume rendering theory~\cite{max1995optical}, the rendered RGB color of a target pixel is obtained by integrating colors and densities along a ray.
In practice, since the output value $(\mathbf{c}, \sigma)$ of all continuous points on a ray is not observable, a subset of points are sampled and the rendered color of the ray is approximated by using the quadrature rule as follows:
\begin{equation}\label{eq:ray_casting}
\hat{C}(\mathbf{r}) = \sum_{i=1}^{N}T_i(1-\exp(-\sigma_i\delta_i)) \mathbf{c}_i,
\end{equation}
where $\mathbf{r}$ denotes a ray, $N$ is the number of samples, and $\delta_i$ is the distance between the $i^\text{th}$ point and its adjacent sample.
Note that $T_i$ indicates the accumulated transmittance along the ray until the $i^\text{th}$ point, which is given by 
\begin{equation}
T_i = \sum_{j=1}^{i-1} \sigma_j \delta_j
\end{equation}

The points on the ray are sampled in a two-stage hierarchical manner to increase rendering efficiency. 
In the first stage, the points are sampled uniformly while, in the second stage, the importance sampling is performed based on the density estimated in the first stage.
Since all processes are fully differentiable, the neural networks encoding radiance fields are optimized with the following objective:
\begin{equation}\label{eq:seen_ray_loss}
\mathcal{L}_{\text{RGB}} = \frac{1}{\left | \mathcal{R} \right |}\sum_{\mathbf{r} \in \mathcal{R}} \left \| C(\mathbf{r}) - \hat{C}(\mathbf{r}) \right \|^2_2
\end{equation}
where $\mathcal{R}$ denotes a set of rays.
Note that the positional encoding is also employed before the MLP to map an input coordinate $(\textbf{x}, \textbf{d})$ onto a higher dimensional space, which is helpful to represent high-frequency scenes.
 
Although NeRF~\cite{mildenhall2020nerf} achieves outstanding photorealistic view synthesis results, it requires a lot of images densely captured by calibrated cameras in general.
To alleviate this data acquisition issue, we propose a simple yet effective few-shot novel view synthesis approach.


\section{Proposed Method}
\label{sec:method}

Our approach learns a robust neural volume rendering model based only on a few input images without any prior knowledge about a scene.
We focus on how to alleviate the reconstruction inconsistency given by lack of input views and the degeneracy induced by overfitting.
This section discusses the proposed information-theoretic regularizations imposed on NeRF-based models to achieve our goals.

\subsection{Motivation}
\label{sub:motivation}

Due to the small number of views available, few-shot 3D reconstruction and volume rendering are inherently prone to result in noisy estimations and degenerate solutions.
For example, Figure~\ref{fig:nerf_noise} shows that NeRF fails to reconstruct the 3D scene accurately, mainly due to insufficient viewpoints.
In addition, the learned model is severely overfit to seen images and converges to a degenerate solution, especially when the viewpoints of training images are similar to each other.
Consequently, the rendering fails with only a slight change of view as illustrated in Figure~\ref{fig:nerf_overfit}.  

\subsection{Regularization by Ray Entropy Minimization}
\label{sub:entropy}

To alleviate the reconstruction inconsistency, we impose the sparsity constraint on the reconstructed scene, which is achieved by minimizing the entropy of each ray density function using additional regularization terms.
This constraint is reasonable because only a small subset of sampled points along a ray hit objects or background in a scene and the rest of points are likely to observe noise.

\paragraph{Ray density}
Before discussing the ray entropy, we first define the normalized ray density denoted by $p(\mathbf{r})$ as follows:
\begin{align}
p(\mathbf{r}_i) = \frac{\alpha_i}{\sum_j \alpha_j} = \frac{1-\exp{(-\sigma_i\delta_i})}{\sum_j 1-\exp{(-\sigma_j \delta_j})},
\label{eq:ray_density}
\end{align}
where $\mathbf{r}_i$ ($i = 1, \dots, N$) is a sampled point in a ray, $\sigma_i$ is the observed density at $\mathbf{r}_i$, $\delta_i$ is a sampling interval around $\mathbf{r}_i$, and $\alpha_i \equiv 1 - \exp(-\sigma_i \delta_i)$ is the opacity at $\mathbf{r}_i$.
The points on the ray are sampled in a two-stage manner, which are drawn from the uniform distribution followed by from the distribution of opacity as in~\cite{mildenhall2020nerf}.
Note that we actually utilize the opacity $\alpha_i$ to compute $p(\mathbf{r}_i)$ in~\eqref{eq:ray_density}, instead of the density $\sigma_i$, to consider the irregular sampling interval $\delta_i$. 
\paragraph{Ray entropy}
Following Shannon Entropy~\cite{shannon1948mathematical}, we define the entropy of a discrete ray density function given by
\begin{align}
H(\mathbf{r}) = -\sum_{i=1}^{N}p(\mathbf{r}_i) \log{p(\mathbf{r}_i)}.
\label{eq:ray_entropy}
\end{align} 
Because $\sigma$ and $\delta$ values are already calculated to conduct volume rendering procedure in \eqref{eq:ray_casting}, the computation of ray entropy incurs only negligible additional cost.

\paragraph{Disregarding non-hitting rays}
One issue in ray entropy minimization is that some rays are enforced to have low entropy though they do not hit any objects in the scene.
To prevent the potential artifacts induced by this issue, we simply disregard the rays with low density for the entropy minimization.
Formally, we employ a mask variable $M(\cdot$) to indicate the rays that have sufficient observations of the scene, which is based on the opacity as follows:
\begin{equation}\label{eq:ray_mask}
M(\mathbf{r}) = 
\begin{cases}
1  & \text{if~} Q(\mathbf{r}) > \epsilon \\
0  & \text{otherwise}
\end{cases},
\end{equation}
\vspace{-0.1cm}
where
\vspace{-0.1cm}%
\begin{align}
Q(\mathbf{r}) =  \sum_{i=1}^{N} 1-\exp{(-\sigma_i\delta_i})
\end{align}
denotes the cumulative ray density.

\paragraph{Ray entropy loss}
Based on the ray entropy computed in \eqref{eq:ray_entropy}, our ray entropy minimization loss is defined as follows:
\begin{equation}\label{eq:entropy_loss}
\mathcal{L}_{\text{entropy}} = \frac{1}{\left | \mathcal{R}_s \right | + \left | \mathcal{R}_{u} \right |}\sum_{\mathbf{r} \in {\mathcal{R}_{s} \cup \mathcal{R}_{u}}} M(\mathbf{r}) \odot H(\mathbf{r}),
\end{equation}
where $\mathcal{R}_s$ denotes a set of rays from training images, $\mathcal{R}_{u}$ denotes a set of rays from randomly sampled unseen images, {and $\odot$ indicates element-wise multiplication.}
Note that, NeRF-based models are unavailable to use rays from the unseen images due to the lack of their ground-truths of pixel colors, while our model can utilize them since entropy regularization does not require the ground-truths.
We observe that it is beneficial to utilize the rays even from the unobserved viewpoint for better scene reconstruction.

\paragraph{Comparison with existing methods}
There exist only a few prior works that impose constraints on scene representations or models, but their objectives are different from ours~\cite{lombardi2019neural,yu2021plenoctrees}, \eg, focusing on improving reconstruction quality and/or achieving real-time processing without considering few-shot training scenarios.
In addition, since they rely on 3D volume entropy based on voxel representations, they have to draw a large number of samples to estimate 3D density or occupancy map, resulting in heavy computational cost in terms of both space and time complexities.
On the other hand, InfoNeRF employs the entropy minimization along a ray via 1D sampling and consequently, it runs very efficiently compared to the methods based on 3D volume entropy.

\subsection{Regularization by Information Gain Reduction}
\label{sub:information}

According to our observations, when the training images have sufficiently diverse viewpoints, the proposed entropy regularization is very helpful to improve the quality of both rendered images and 3D depth estimation in the few-shot setting.
However, if all the training images have similar viewpoints to each other, the models are prone to overfit to seen images and fail to generalize to unseen views.  
This is probably because the lack of diverse observations makes the trained models find degenerate and trivial solutions.

To alleviate the aforementioned limitations, we introduce an additional regularization term to assure the consistent density distribution across rays in the neighborhood.
Given an observed ray $\mathbf{r}$, we sample another ray with a slightly different viewpoint, denoted by $\Tilde{\mathbf{r}}$, and minimize the KL-divergence between the density functions of the two rays.
The motivation of this objective is to make the observations from two similar viewpoints sufficiently consistent so that the model is generalized to a nearby viewpoint, which is achieved by enforcing smoothness to reconstruction results over spatial view perturbations.

The regularization loss for information gain reduction is given by%
\vspace{-0.2cm}
\begin{align}\label{eq:info_gain}
\mathcal{L}_{\text{KL}} &= D_{\text{KL}}\Big(P(\mathbf{r}) || {P}(\Tilde{\mathbf{r}})\Big) = \sum_{i=1}^{N}p(\mathbf{r}_i) \log\frac{p(\mathbf{r}_i)}{p(\Tilde{\mathbf{r}_i})}, 
\end{align}
where $\Tilde{\mathbf{r}}_i$ is a sampled point for observation in ray $\Tilde{\mathbf{r}}$.
{In our implementation, we obtain $\Tilde{\mathbf{r}}$ by slightly rotating the camera pose of $\mathbf{r}$ in the range from $-5^{\circ}$ to $5^{\circ}$.}

\subsection{Overall Objective}
The total loss function to train a neural implicit model for few-shot neural volume rendering is given by
\begin{equation}
\mathcal{L}_{\text{total}} = \mathcal{L}_{\text{RGB}} +
\lambda_1 \mathcal{L}_{\text{entropy}} + \lambda_2  \mathcal{L}_{\text{KL}},
\end{equation}
where $\lambda_1$ and $\lambda_2$ are balancing terms for our regularization terms.
As mentioned in~\eqref{eq:seen_ray_loss}, the reconstruction loss, denoted by $\mathcal{L}_\text{RGB}$, is given by
\begin{equation}
\mathcal{L}_{\text{RGB}} = \frac{1}{\left | \mathcal{R}_{s} \right |}\sum_{\mathbf{r}\in \mathcal{R}_s} \left \| C(\mathbf{r}) - \hat{C}(\mathbf{r}) \right \|^2_2,
\end{equation}
which uses only a set of rays from training images with pixel-level ground-truth, unlike $\mathcal{L}_{\text{entropy}}$ and $\mathcal{L}_{\text{KL}}$ which utilize the rays even from unobserved viewpoints.


\begin{table*}[t]
\centering
        \caption{Experimental results of few-shot novel view synthesis on the Realistic Synthetic 360$^{\circ}$ dataset in the 4-view setting.
Our approach outperforms all other existing methods by significant margins in all image quality metrics.
The asterisk ($\ast$) denotes that the model is pretrained on an external training dataset with dense input views and finet-uned on this dataset with 4 input views.
We run all experimental five times with different viewpoint samples and the same hyperparameters, and compute the average scores and their standard deviations.
}
\vspace{-1mm}
\scalebox{0.9}{
    \setlength\tabcolsep{12pt} \hspace{-0.25cm}
    \begin{tabular}{ccccccc}
    \toprule
    Method  & PSNR  $\uparrow$ & SSIM $\uparrow$ & LPIPS $\downarrow$ & FID $\downarrow$ & KID $\downarrow$
    \\ 
    \hline 
    \hline
    NeRF, 100 views  &  31.01 &0.947 & 0.081 & 42.83 & 0.002  \\ \hline
   PixelNeRF$^\ast$~\cite{yu2021pixelnerf} & 16.09\footnotesize{$\pm$0.78}& 0.738\footnotesize{$\pm$0.012}&0.390\footnotesize{$\pm$0.030}	&265.25\footnotesize{$\pm$6.73} 	&0.127\footnotesize{$\pm$0.006}\\          \hdashline
    NeRF~\cite{mildenhall2020nerf}  &15.93\footnotesize{$\pm$1.06} & 0.780\footnotesize{$\pm$0.014} & 0.320\footnotesize{$\pm$0.049} & 215.16\footnotesize{$\pm$2.32}&	0.074\footnotesize{$\pm$0.012}\\
    DietNeRF~\cite{jain2021putting} &16.06\footnotesize{$\pm$1.13} & 0.793\footnotesize{$\pm$0.019} & 0.306\footnotesize{$\pm$0.050} & \ 197.02\footnotesize{$\pm$12.87}&	0.065\footnotesize{$\pm$0.004}\\
    InfoNeRF (ours)  & \textbf{18.65\footnotesize{$\pm$0.18}} & \textbf{0.811\footnotesize{$\pm$0.008}} & \textbf{0.230\footnotesize{$\pm$0.008}} & \textbf{181.47\footnotesize{$\pm$4.97}}&	\textbf{0.062\footnotesize{$\pm$0.004}}\\ \bottomrule
    \end{tabular}}
    \label{tab:synthetic_results}
 \end{table*}

\begin{table*}
\centering
    \caption{Average PSNRs and standard deviations of individual scenes on the Realistic Synthetic 360$^{\circ}$ dataset in the 4-view setting.}
          \vspace{-1mm}
\scalebox{0.9}{
    \setlength\tabcolsep{4.5pt} \hspace{-0.25cm}
    \begin{tabular}{ccccccccc|c}
    \toprule
    Method & Lego & Chair & Drums & Ficus & Hotdog & Materials& Mic& Ship & Avg.
    \\ 
    \hline \hline
    NeRF, 100 views				&32.54&	33.00&	25.01&	30.13&	36.18&	29.62&	32.91&	28.65&	31.01\\ 
   \hline
      PixelNeRF$^\ast$~\cite{yu2021pixelnerf} 	&	 15.14\footnotesize{$\pm$0.75}&	18.87\footnotesize{$\pm$1.38}&	15.10\footnotesize{$\pm$0.63}&	16.60\footnotesize{$\pm$0.70}&	19.37\footnotesize{$\pm$1.78}&	12.31\footnotesize{$\pm$1.02}&	16.35\footnotesize{$\pm$0.97}&	14.96\footnotesize{$\pm$0.75}&	16.09\footnotesize{$\pm$0.78} \\ \hdashline
    NeRF~\cite{mildenhall2020nerf} 	&15.61\footnotesize{$\pm$4.53}&	18.57\footnotesize{$\pm$1.64}&	12.50\footnotesize{$\pm$0.98}&	16.37\footnotesize{$\pm$2.24}&	19.64\footnotesize{$\pm$2.26}&	15.65\footnotesize{$\pm$4.16}&	14.78\footnotesize{$\pm$2.37}&	14.30\footnotesize{$\pm$4.04}&	15.93\footnotesize{$\pm$1.06}\\    	
    DietNeRF~\cite{jain2021putting} 		&17.13\footnotesize{$\pm$4.77}&	19.37\footnotesize{$\pm$3.12}&	13.74\footnotesize{$\pm$1.55}&	15.76\footnotesize{$\pm$3.56}&	18.24\footnotesize{$\pm$5.28}&	15.00\footnotesize{$\pm$5.18}&	17.71\footnotesize{$\pm$1.55}&	11.51\footnotesize{$\pm$4.27}&	16.06\footnotesize{$\pm$1.13}\\
 InfoNeRF (ours) 					&\textbf{18.92\footnotesize{$\pm$0.51}}&	\textbf{20.06\footnotesize{$\pm$1.11}}&	\textbf{14.33\footnotesize{$\pm$0.62}}&	\textbf{19.41\footnotesize{$\pm$0.07}}&	\textbf{21.30\footnotesize{$\pm$2.31}}&	\textbf{18.34\footnotesize{$\pm$0.88}}&	\textbf{18.55\footnotesize{$\pm$1.71}}&	\textbf{18.27\footnotesize{$\pm$0.71}}&	\textbf{18.65\footnotesize{$\pm$0.18}}\\
   \bottomrule
    \end{tabular}}
    \label{tab:synthetic_psnr}
          \vspace{-0.1cm}
 \end{table*}
\section{Experiments}
\label{sec:experiments}

We demonstrate the effectiveness of the proposed approach, referred to as InfoNeRF, on the standard benchmarks.
This section also discusses the characteristics of our algorithm based on the experiment results.

\subsection{Datasets}

We describe the details of three benchmarks employed to evaluate our algorithm, which include the Realistic Synthetic 360$^{\circ}$~\cite{mildenhall2020nerf}, ZJU-MoCap~\cite{peng2021neural}, and DTU~\cite{jensen2014large} datasets.

\vspace{-0.2cm}
\paragraph{Realistic Synthetic 360$^{\circ}$}
This benchmark is common for neural volume rendering, which contains 8 synthetic scenes with view-dependent light transport effects.
Each scene has an object at the center and 400 rendered images from inward-facing virtual cameras with different viewpoints.
For few-shot training, we randomly sample 4 viewpoints out of 100 training images in each scene, and use the 200 testing images for evaluation.

\vspace{-0.2cm}
\paragraph{ZJU-MoCap}
This dataset consists of multi-view videos capturing human motion from 23 calibrated cameras.
Following~\cite{peng2021neural}, we sample 4 uniformly distributed viewpoints to construct a training set and use the remaining images for testing.

\vspace{-0.2cm}
\paragraph{DTU MVS Dataset (DTU)}
The images in this dataset contain complex and real-world scenes captured by the calibrated cameras in controlled environments.
All the collected images have similar viewpoints and face only one side of a scene.
We conduct experiments on 15 scenes, where we optimize the model with 3 images out of 49 views while testing with the remaining 46 views.

\subsection{Implementation and Evaluation}

\paragraph{Implementation details}
Our implementation is based on PyTorch~\cite{paszke2019pytorch}.
{We use Adam optimizer~\cite{kingma2014adam} with the initial learning rate of $5 \times 10^{-4}$, which decays exponentially by a factor of 10 at every 250,000 iterations.}
{The balancing term for $\mathcal{L}_\text{KL}$ is decayed by a factor of 2 at every 5,000 iterations.} 
We set the number of rays from seen and unobserved views, denoted respectively by $|\mathcal{R}_s|$ and $|\mathcal{R}_u|$, identically to 1024, and our experiments are conducted with a single NVIDIA Titan XP GPU.

\vspace{-0.2cm}
\paragraph{Metrics}
We evaluate the novel view rendering quality based on the standard image quality metrics, peak signal-to-noise ratio (PSNR) and structural similarity (SSIM)~\cite{wang2004image}.
We also use perceptual metrics, learned perceptual image patch similarity (LPIPS)~\cite{zhang2018unreasonable}, Frèchet inception distance (FID)~\cite{heusel2017gans}, and kernel inception distance (KID)~\cite{binkowski2018demystifying}.
LPIPS estimates normalized features distance between image pair while FID and KID compute the distance in Inception representations~\cite{szegedy2016rethinking} between two sets of images.

\subsection{Results}

\subsubsection{Realistic Synthetic 360$^{\circ}$}
\label{sec:results_synthetic}

We compare our approach with NeRF~\cite{mildenhall2020nerf}, DietNeRF~\cite{jain2021putting}, and PixelNeRF~\cite{yu2021pixelnerf} on Realistic Synthetic 360$^{\circ}$ dataset.
NeRF, DietNeRF, and InfoNeRF (ours) are trained with randomly sampled 4 views from scratch.
Unlike others, PixelNeRF is pretrained on the DTU~\cite{jensen2014large} dataset with dense input views, and we fine-tune the model with 4 sampled views to handle the domain shift issue between the two datasets.

Table~\ref{tab:synthetic_results} presents overall quantitative results, where InfoNeRF consistently outperforms the baseline algorithms in terms of all metrics with considerable margins while having lower standard deviations.
Table~\ref{tab:synthetic_psnr} breaks down the PSNU scores into 8 individual scenes, where InfoNeRF obviously achieves significant gains for all scenes.
Refer to our supplementary document for the results of SSIM and LPIPS, which have the same tendencies.

Figure~\ref{fig:teaser} demonstrates the qualitative results on the novel viewpoints, where InfoNeRF shows outstanding quality in the rendered images compared to all the compared methods.
As illustrated in Figure~\ref{fig:teaser}(f), the quality of the depth maps estimated by InfoNeRF looks impressive while we notice that all the compared algorithms often fail to reconstruct 3D structures accurately and DietNeRF even has color distortion due to its high-level semantic consistency loss.

\begin{figure*}[t]
    \centering
    \begin{subfigure}[h]{0.16\linewidth}
         \centering
         \includegraphics[width=\textwidth]{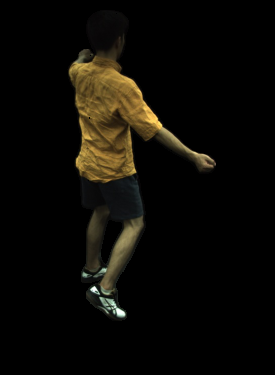}
         \caption{Ground-truth}
     \end{subfigure} \hspace{0.3cm}
    \begin{subfigure}[h]{0.16\linewidth}
         \centering
         \includegraphics[width=\textwidth]{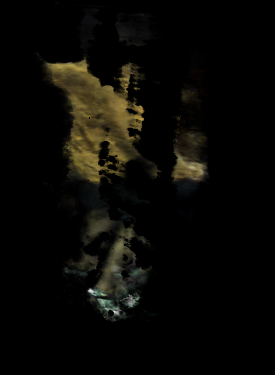}
         \caption{NeRF~\cite{mildenhall2020nerf}}
     \end{subfigure} \hspace{0.3cm}
     \begin{subfigure}[h]{0.16\linewidth}
         \centering
      \includegraphics[width=\textwidth]{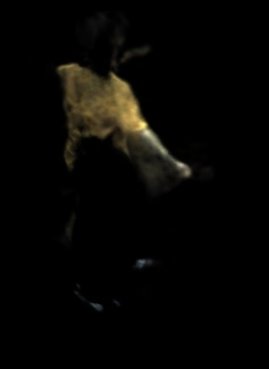}
         \caption{NV~\cite{lombardi2019neural}}
     \end{subfigure} \hspace{0.3cm}
    \begin{subfigure}[h]{0.16\linewidth}
         \centering
         \includegraphics[width=\textwidth]{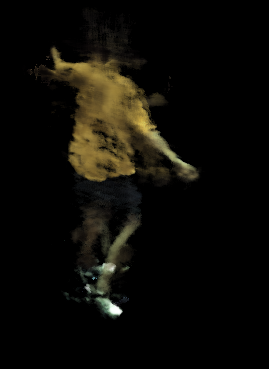}
         \caption{InfoNeRF (ours)}
     \end{subfigure} \hspace{0.3cm}
    \begin{subfigure}[h]{0.16\linewidth}
         \centering
         \includegraphics[width=\textwidth]{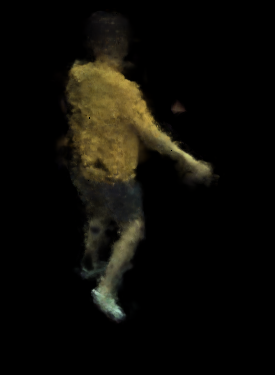}
         \caption{NB~\cite{peng2021neural}}
     \end{subfigure}
       \vspace{-0.1cm}
     \caption{Qualitative comparison on the ZJU-MoCap dataset in 4-view setting.
    We visualize the rendering results of prior-free algorithms (b-d) including ours, and prior-based algorithm (e).
    While existing prior-free algorithms (b-c) often suffer from inconsistent reconstruction and missing parts of the human body, InfoNeRF manages to render most of the human body comparable to prior-based algorithm (e).}
        \label{fig:mocap}
       \vspace{-0.1cm}
\end{figure*}

\subsubsection{ZJU-MoCap}
\begin{table}[t]
     \caption{Quantitative comparison on the ZJU-MoCap dataset in the 4-view setting.
NB~\cite{peng2021neural} has the geometric prior by exploiting the pretrained human body model (SMPL).}
\vspace{-0.1cm}
\centering
\scalebox{0.9}{
   \setlength\tabcolsep{8pt} \hspace{-0.25cm}
    \begin{tabular}{cccccc}
    \toprule
    Method & Prior & PSNR $\uparrow$  & SSIM $\uparrow$ & LPIPS $\downarrow$  
    \\ 
    \hline
    \hline
    NB~\cite{peng2021neural}& \checkmark & 24.18 & 0.888& 0.182\\
    \hdashline 
    NeRF~\cite{mildenhall2020nerf} &\multirow{3}{*}{}&  20.19&0.794&0.309\\
    NV~\cite{lombardi2019neural}&& 21.74 & 0.827& 0.253\\
    InfoNeRF (ours) && \textbf{22.88} & \textbf{0.838} &\textbf{0.242}\\
    \bottomrule
    \end{tabular}}
    \label{tab:human_results}
    \vspace{-0.2cm}
\end{table}

For the ZJU-MoCap dataset, InfoNeRF is evaluated in comparison with NeRF~\cite{mildenhall2020nerf}, Neural Volume (NV)~\cite{lombardi2019neural}, and Neural Body (NB)~\cite{peng2021neural}, where all algorithms are trained with 4 images.
Note that, since NB employs a pretrained human body model denoted by SMPL~\cite{loper2015smpl} as its prior, the performance of NB can be regarded as the upper-bound of all other methods.

Table~\ref{tab:human_results} summarizes the experimental results on the ZJU-MoCap dataset, where InfoNeRF achieves the best performance among the methods without using the prior in terms of all the tested metrics.
Figure~\ref{fig:mocap} demonstrates the qualitative results of all the compared methods, and the reconstruction result given by InfoNeRF is particularly accurate.

\subsubsection{DTU MVS Dataset (DTU)}

Contrary to the other two datasets, DTU has substantially different characteristics because the images in each scene have similar viewpoints.
PixelNeRF takes advantage of this property and learns the scene-agnostic model successfully while the original NeRF exhibits poor generalization performance in this dataset with few-shot learning. 

We compare our algorithms with NeRF and PixelNeRF in this dataset.
We train InfoNeRF and NeRF from scratch without exploiting any scene prior, so it is not possible to reconstruct invisible parts of the scene.
Therefore, the na\"ive evaluation of the algorithms without scene prior is not desirable and we use the mask corresponding to the visible parts of each scene for the evaluation performance. 
Table~\ref{tab:dtu_results} presents the experimental results on the DTU dataset, where our algorithm achieves outstanding performance compared to NeRF.
Note that PixelNeRF achieves the highest performance because it exploits the dataset prior by pretraining on the training split of DTU {with dense input views.}
\begin{table}[t]
\centering
\caption{Quantitative comparison on the DTU dataset in 3-view setting.
PixelNeRF~\cite{yu2021pixelnerf} has the dataset prior by pretraining on other scenes of DTU with dense input views.
}
\vspace{-0.1cm}
\scalebox{0.9}{
\setlength\tabcolsep{8pt} \hspace{-0.25cm}
\begin{tabular}{ccccccc}
\toprule
Method & Prior & PSNR $\uparrow$  & SSIM $\uparrow$ & LPIPS $\downarrow$  
\\ 
\hline 
\hline
PixelNeRF~\cite{yu2021pixelnerf} & \checkmark & 19.55&	0.724&	0.286 \\
\hdashline
NeRF~\cite{mildenhall2020nerf} && 				\ \ 8.50&	0.426&	0.611 \\
InfoNeRF (ours)&&							\textbf{11.23}&	\textbf{0.445}&	\textbf{0.543} \\
\bottomrule
\end{tabular}}
\vspace{-0.2cm}
\label{tab:dtu_results}
\end{table}

\subsection{Analysis}

\begin{figure}[t]
     \centering
     \begin{subfigure}[h]{0.45\linewidth}
         \centering
         \includegraphics[width=\textwidth]{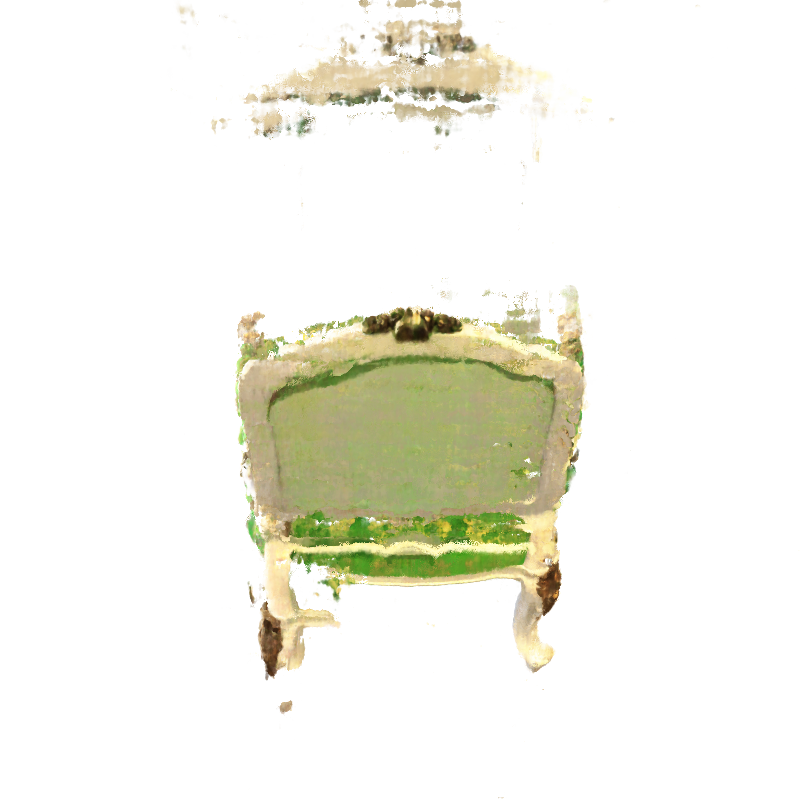}
        \vspace{-8mm}
        \caption{seen view}
        \label{fig:seen_sampling}
     \end{subfigure}
     \begin{subfigure}[h]{0.45\linewidth}
         \centering
         \includegraphics[width=\textwidth]{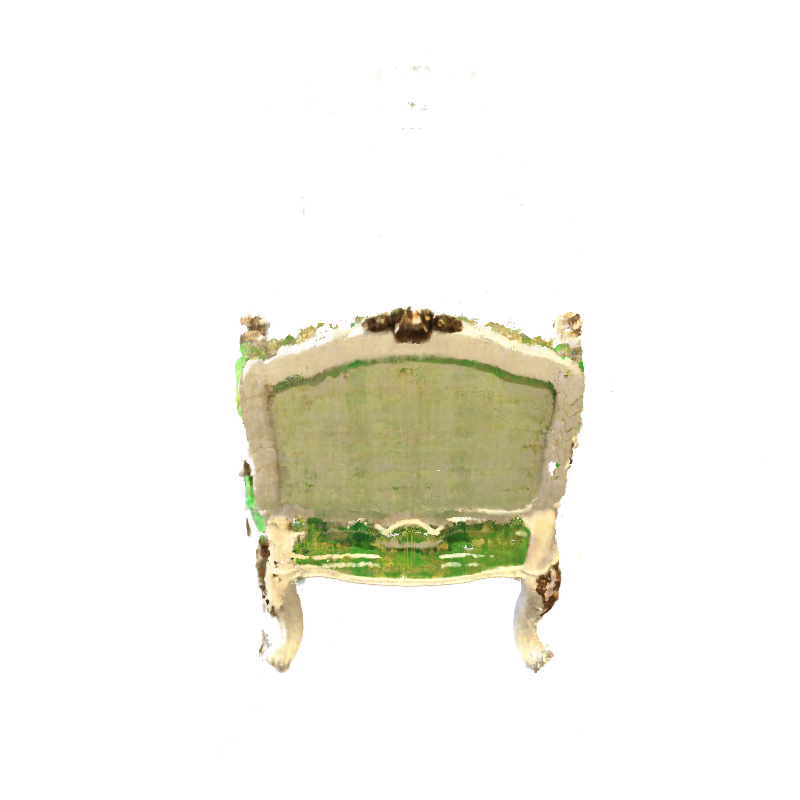}
        \vspace{-8mm}
         \caption{seen view + unseen view}
        \label{fig:unseen_sampling}
     \end{subfigure}
	\vspace{-0.1cm}
             \caption{Benefit of sampling rays for unseen viewpoints.
             We visualize the rendered images on the \textit{Chair} scene of Realistic Synthetic 360$^{\circ}$ dataset, where the noise in the background area is removed completely after adding the rays from unseen viewpoints.}
        \label{fig:ray_num_ablation}
        \vspace{-0.4cm}
\end{figure}

\paragraph{Effect of unseen view sampling}
To verify the effectiveness of sampling from unseen viewpoints for entropy minimization, we run our algorithm by varying the number of rays from unseen views on the \textit{Chair} scene of Realistic Synthetic 360$^{\circ}$ with the images reduced to half.
Table~\ref{tab:ablation_ray_num} presents that increasing the number of rays from the unseen views achieves gradual improvement by alleviating the reconstruction inconsistency but its benefit is saturated when the number of rays is larger than 1,024.
Figure~\ref{fig:ray_num_ablation} visualizes the benefit of sampling additional rays for unseen views; the use of the rays is helpful for noise reduction in this example.

\begin{table}[t]
\centering
\caption{Impact of the number of rays sampled from unseen viewpoints on the \textit{Chair} scene of Realistic Synthetic 360$^{\circ}$ in a 4-view setting.
We fix the number of rays for seen views to 1,024 and vary the number of rays for unseen views.
Bold and underline indicate the first and second place among the results, respectively.
}
\vspace{-0.1cm}
\scalebox{0.9}{
   \setlength\tabcolsep{4pt} \hspace{-0.25cm}
\begin{tabular}{c|c|ccccc}
\toprule
$\#$ of seen rays & $\#$ of unseen rays & PSNR $\uparrow$  & SSIM $\uparrow$ & LPIPS $\downarrow$  
\\ 
\hline 
\hline
\multirow{5}{*}{1024} & 0 & 20.14 & 0.834& 0.225 \\
& 256 & 20.97 & 0.844 &  0.197\\
& 512 &  21.11&  0.851 &  {0.188}\\
& 1024 & \textbf{21.37} &  \underline{0.853} &  \underline{0.185} \\
& 2048 & \underline{21.33} & \textbf{0.855} & \textbf{0.167} \\
\bottomrule
\end{tabular}}
\label{tab:ablation_ray_num}
\vspace{-0.1cm}
\end{table}
\begin{figure*}[t]
    \centering
          \begin{subfigure}[h]{0.137\linewidth}
         \includegraphics[width=\textwidth]{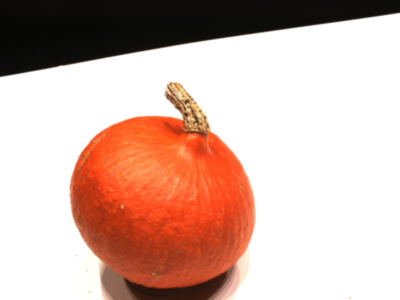}
     \end{subfigure}
    \begin{subfigure}[h]{0.28\linewidth}
         \centering
         \includegraphics[width=0.49\textwidth]{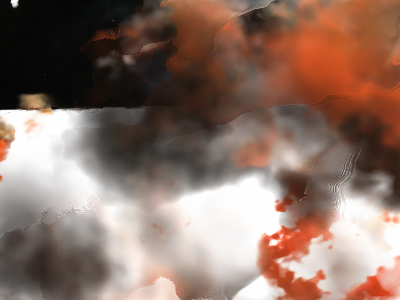}
         \includegraphics[width=0.49\textwidth]{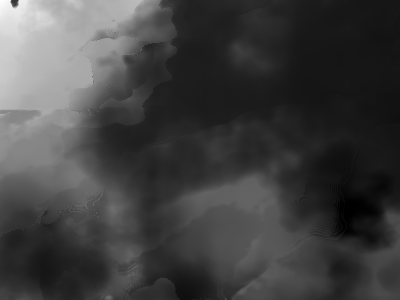}
     \end{subfigure}        
    \begin{subfigure}[h]{0.28\linewidth}
         \centering
         \includegraphics[width=0.49\textwidth]{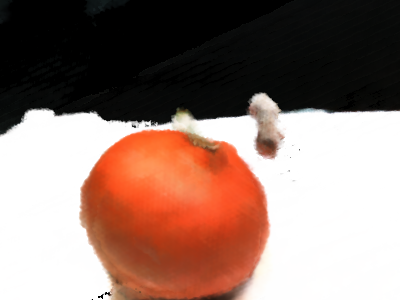}
         \includegraphics[width=0.49\textwidth]{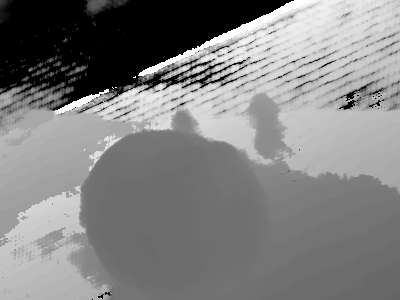}
     \end{subfigure}        
    \begin{subfigure}[h]{0.28\linewidth}
         \centering
         \includegraphics[width=0.49\textwidth]{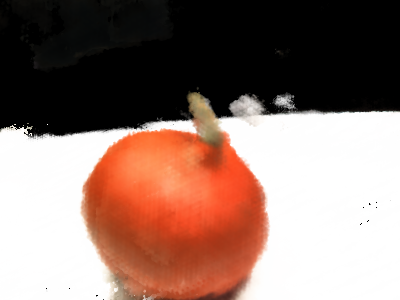}
         \includegraphics[width=0.49\textwidth]{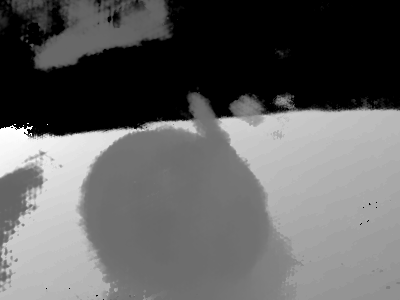}
     \end{subfigure} \\         
\hspace{0.1mm}
          \begin{subfigure}[h]{0.1355\linewidth}
         \centering
         \includegraphics[width=\textwidth]{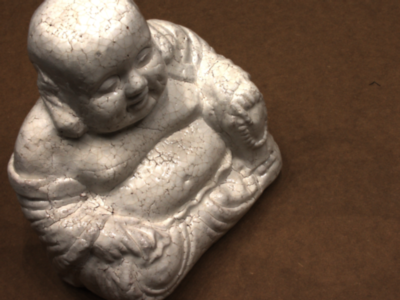}
          \caption{Ground-truth}
     \end{subfigure}
    \begin{subfigure}[h]{0.28\linewidth}
         \centering
         \includegraphics[width=0.49\textwidth]{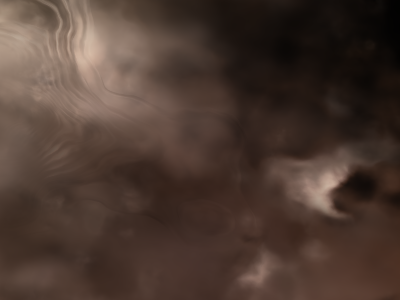}
         \includegraphics[width=0.49\textwidth]{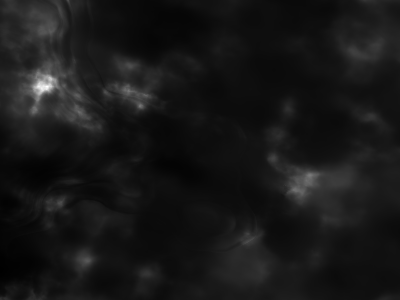}
    	 \caption{NeRF~\cite{mildenhall2020nerf}}
     \end{subfigure}        
    \begin{subfigure}[h]{0.28\linewidth}
         \centering
         \includegraphics[width=0.49\textwidth]{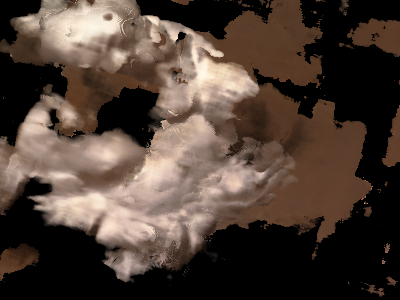}
         \includegraphics[width=0.49\textwidth]{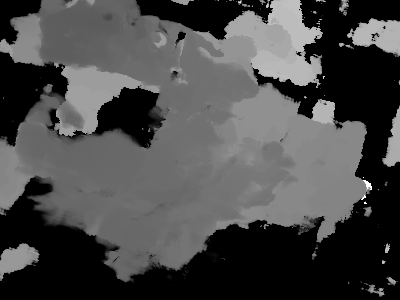}
              \caption{InfoNeRF w/o $\mathcal{L}_\text{KL}$}
     \end{subfigure}        
    \begin{subfigure}[h]{0.2805\linewidth}
         \centering
         \includegraphics[width=0.49\textwidth]{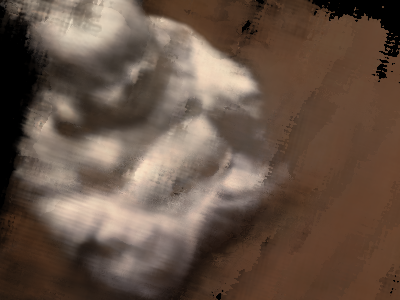}
         \includegraphics[width=0.49\textwidth]{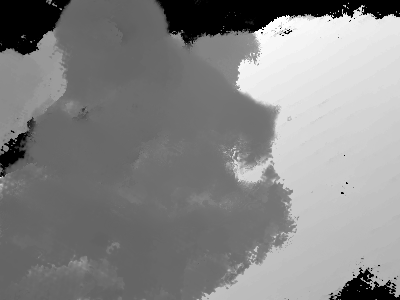}
              \caption{InfoNeRF}
     \end{subfigure}    
     \vspace{-0.1cm}
        \caption{
	Qualitative comparisons of our method with its ablative models on the DTU dataset in the 3-view setting.
        We visualize the image synthesis (left) and the depth estimation (right) results for each algorithm.
        While NeRF suffers from noises, blurs, and artifacts to reconstruct 3D structure, our two loss terms, $\mathcal{L}_\text{entropy}$ and $\mathcal{L}_\text{KL}$, contribute to outstanding rendering quality and fine depth estimation results.
        }
     \vspace{-0.1cm}     
     \label{fig:dtu}
\end{figure*}

\vspace{-0.2cm}
\paragraph{Benefit of regularization}
We analyze the impact of the proposed regularization schemes, ray entropy minimization loss, $\mathcal{L}_\text{entropy}$, and ray information gain reduction loss, $\mathcal{L}_\text{KL}$.
Table~\ref{tab:ablation_dtu_results} shows the ablative results of InfoNeRF on the DTU dataset.
The entropy minimization loss successfully improves PSNR, but the SSIM value gets worse on this dataset.
{This is because the models are prone to overfit to the seen images when all the training images have similar viewpoints to each other.}
However, thanks to our information gain reduction loss that enforces smoothness to reconstruction results over spatial view perturbations, our full model helps alleviate the overfitting issue and prevent degenerate solutions.
Note that, in the other two datasets with substantial viewpoint variations, the entropy minimization loss works well while the information gain reduction loss makes minor contribution in general.
Figure~\ref{fig:dtu} visualizes the rendering quality of InfoNeRF in comparison to its ablative models on the DTU dataset.
Although InfoNeRF without the information gain reduction loss tends to generate crisp images, there exists a lot of noise in the outputs and inconsistency in the depth maps.
On the other hand, our full algorithm manages to reconstruct the overall shape and geometry of the scene more accurately.
Refer to Table C of the supplementary document for more detailed ablation results on the Realistic Synthetic 360$^{\circ}$ dataset.

\begin{table}[t] 
    	\centering
		\caption{Ablative results of our regularization schemes on the DTU dataset in a 3-view setting.}
	\vspace{-0.2cm}
	\scalebox{0.9}{
	 \setlength\tabcolsep{3pt} 
	 \hspace{-3mm}
    	\begin{tabular}{c|cc|ccc}
    	\toprule
	Method & $\mathcal{L}_\text{entropy}$ & $\mathcal{L}_\text{KL}$ & PSNR $\uparrow$  & SSIM $\uparrow$  & LPIPS $\downarrow$\\
	\hline
	\hline
	NeRF & &  										&8.50	&0.426	&0.611\\
	InfoNeRF w/o $\mathcal{L}_\text{entropy}$ &&\checkmark 	&8.91	&0.439	&0.581\\
	InfoNeRF w/o $\mathcal{L}_\text{KL}$ & \checkmark  & 		&10.54	&0.418	&0.561\\	
InfoNeRF & \checkmark& \checkmark						&\textbf{11.23}	&\textbf{0.445}	&\textbf{0.543}\\
	\bottomrule
    \end{tabular}}
		\label{tab:ablation_dtu_results}
\end{table}

\begin{figure}
	\centering
	\includegraphics[width=1\linewidth]{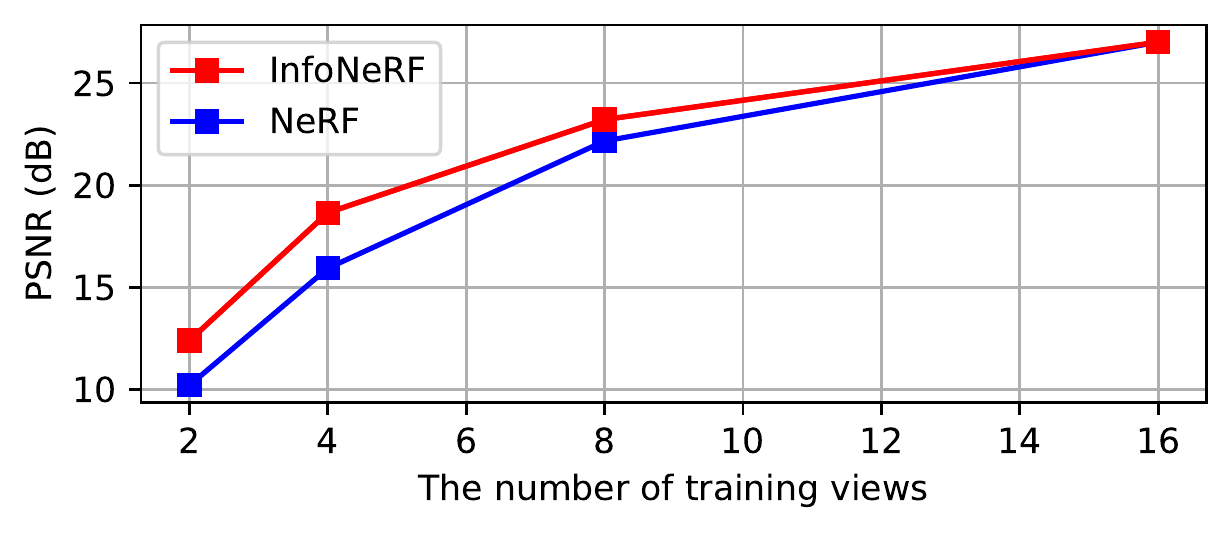}
	\vspace{-0.8cm}
	\caption{PSNR with respect to the number of views for training on the Realistic Synthetic 360$^\circ$ dataset.}
	\label{fig:psnr_img_num}
			\vspace{-0.4cm}
\end{figure}

\vspace{-0.2cm}
\paragraph{Robustness to the number of views}

Figure~\ref{fig:psnr_img_num} illustrates performance of InfoNeRF by varying the number of views for training on the Realistic Synthetic 360$^\circ$.
Compared to NeRF, InfoNeRF illustrates improved results in terms of all metrics until 8 views, but its merit is saturated as the number of views increases.
This is partly because the uncertainty of the reconstructed scene decreases as the number of training views increases, weakening the importance of entropy regularization.
See Table~D of our supplementary document for SSIM and LPIPS results.

\begin{figure}[t]
     \centering
     \begin{subfigure}[h]{0.45\linewidth}
         \centering
         \includegraphics[width=\textwidth]{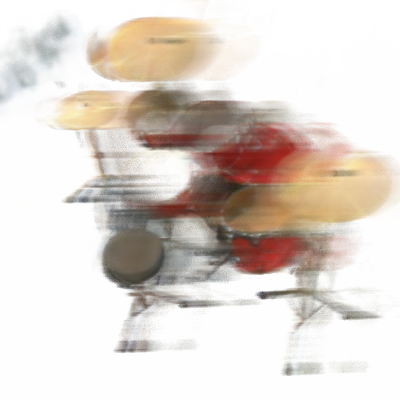}
     \end{subfigure}
     \begin{subfigure}[h]{0.45\linewidth}
         \centering
         \includegraphics[width=\textwidth]{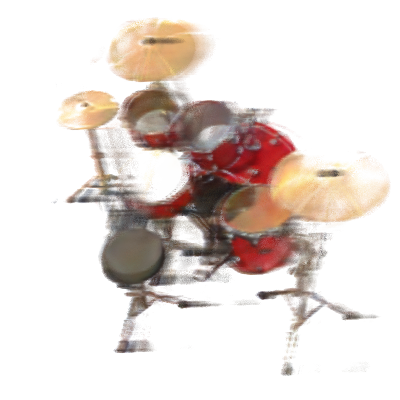}
     \end{subfigure}
             \vspace{-0.4cm}
        \caption{
        Qualitative comparison on the \textit{Drums} scene in the Realistic Synthetic 360$^{\circ}$ dataset. 
        InfoPixelNeRF (right) reduces blur significantly compared to PixelNeRF (left), which demonstrates the effectiveness of our regularization schemes.
        }
        \vspace{-0.2cm}
	\label{fig:compare_pixelnerf}
\end{figure}

\vspace{-0.2cm}
\paragraph{Integration into PixelNeRF}
To demonstrate the generality of our method, we incorporate the proposed regularization method to PixelNeRF~\cite{yu2021pixelnerf}, and refer to this version of our model as InfoPixelNeRF.
Figure~\ref{fig:compare_pixelnerf} illustrates qualitative comparisons between both models, where InfoPixelNeRF reduces blur in the rendered image significantly.
Table E of our supplementary file presents detailed comparisons.

\section{Conclusion}
\label{sec:conclusion}
We proposed an information-theoretic regularization technique for few-shot novel view synthesis.
Existing few-shot view synthesis methods suffer from inconsistent reconstruction, which often generates noise, blur, or artifacts in rendered images, and overfitting to seen views, which leads to degenerate solutions.
To address these issues, we introduced two effective regularization schemes, ray entropy minimization and ray information gain reduction.
Despite its simplicity, the proposed method turns out to be effective to alleviate reconstruction inconsistency across views.
We demonstrated outstanding performance of our method on multiple standard benchmarks, and also conducted a detailed analysis of our approach via extensive analysis.

\vspace{-2mm}
\paragraph{Acknowledgments}
This work was supported in part by the IITP grants [2021-0-02068, AI Innovation Hub; 2021-0-01343, Interdisciplinary Program in AI (SNU)] and the Bio \& Medical Technology Development Program of NRF [2021M3A9E4080782] funded by the Korean government (MSIT) and by Samsung Advanced Institute of Technology.%



{\small
\bibliographystyle{ieee_fullname}
\bibliography{egbib}
}


\onecolumn 

\setcounter{section}{0}
\setcounter{table}{0}
\setcounter{figure}{0}
\renewcommand\thesection{\Alph{section}}
\renewcommand\thetable{\Alph{table}}
\renewcommand\thefigure{\Alph{figure}}

\newpage
\section{Validation for $\mathcal{L}_\text{entropy}$ and $\mathcal{L}_\text{KL}$}

\begin{figure*}[h]
\centering
    \begin{subfigure}[h]{0.35\linewidth}
        \centering
        \includegraphics[width=\textwidth]{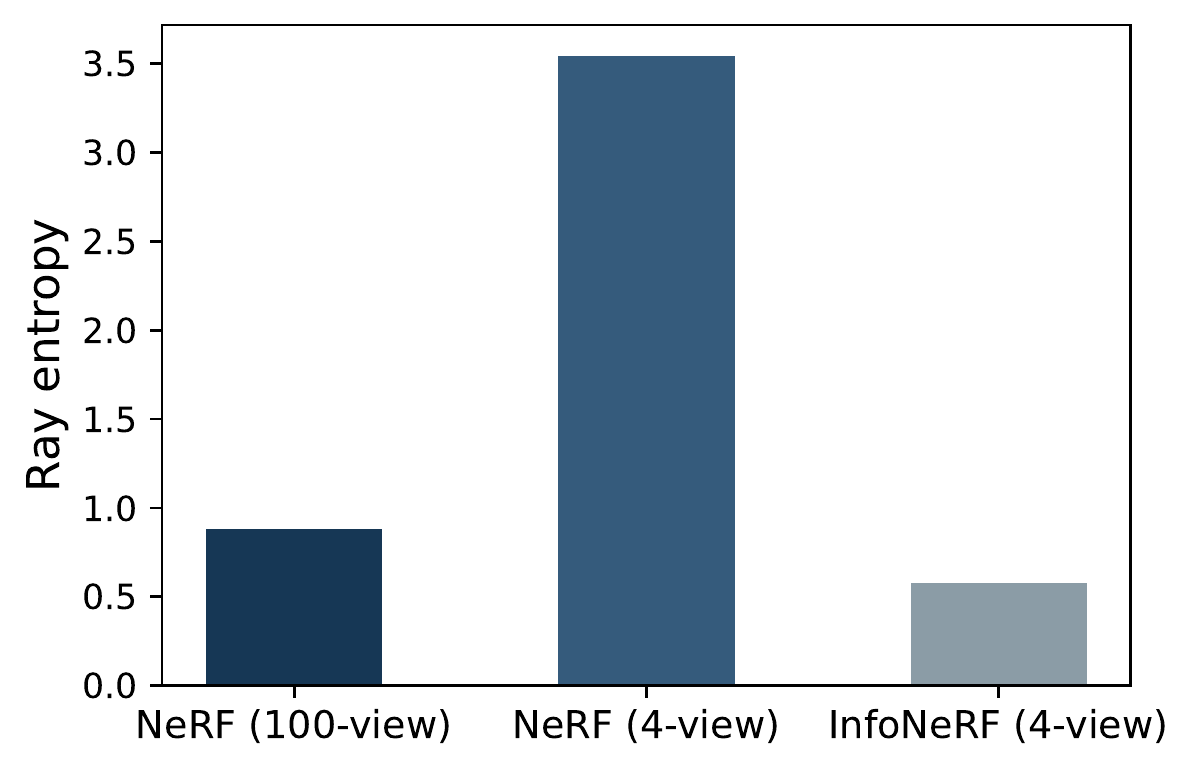}
        \caption{Ray entropy values after training}
    \end{subfigure}
    \hspace{1.8cm}
    \begin{subfigure}[h]{0.35\linewidth}
         \centering
         \includegraphics[width=\textwidth]{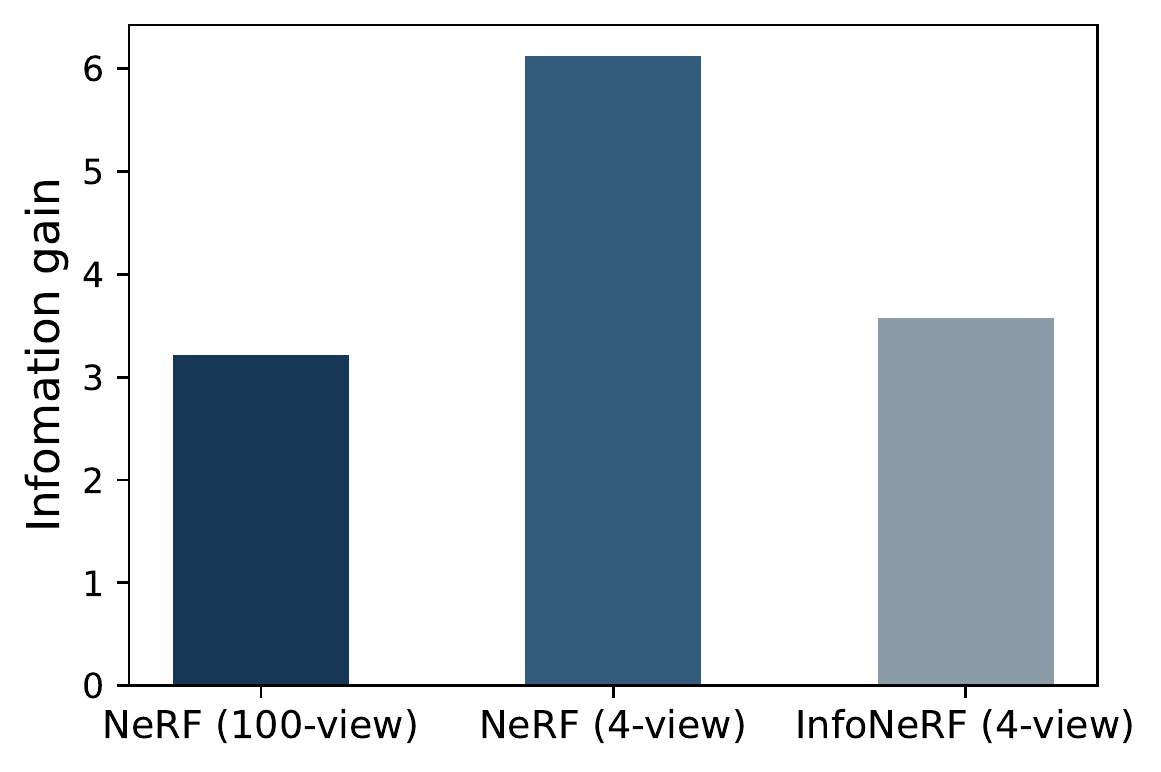}
         \caption{Information gain values after training}
     \end{subfigure}
	\caption{
	Quantitative comparisons of ray entropy and information gain computed in three different models.
	Although the few-shot NeRF model gives higher ray entropy and information gain than the NeRF with 100 views, the proposed few-shot InfoNeRF model reduces both values, which motivates the two regularization loss employed in our approach. 
	}
	\label{fig:validation_of_loss}
\end{figure*}

To analyze the behavior of our two regularization losses, we measure the values of ray entropy and information gain of NeRF models after training with different settings.
Figure~\ref{fig:validation_of_loss} illustrates that both ray entropy and information gain values of the few-shot NeRF model are much higher than those of the NeRF model with 100 views.
This observation implies that the rendered images by the few-shot NeRF are noisier and the reduction of the two losses is desirable.
Note that InfoNeRF with 4 views has almost similar values to the original NeRF model with 100 views, which presents that the proposed algorithm provides compact and consistent reconstruction results even with a small number of views.

\section{Experiments on More Complex Scenarios} 
\begin{figure*}[h]
\centering
    \begin{subfigure}[h]{0.162\linewidth}
        \centering
        \includegraphics[width=\textwidth]{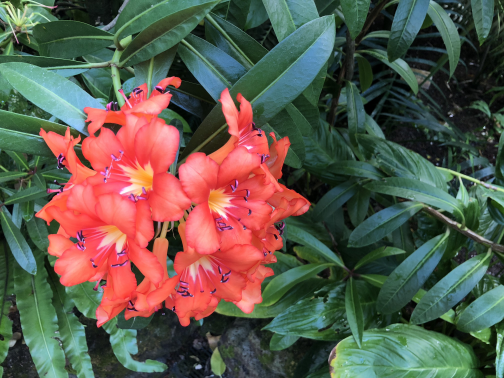}
    \end{subfigure}
    \begin{subfigure}[h]{0.162\linewidth}
         \centering
         \includegraphics[width=\textwidth]{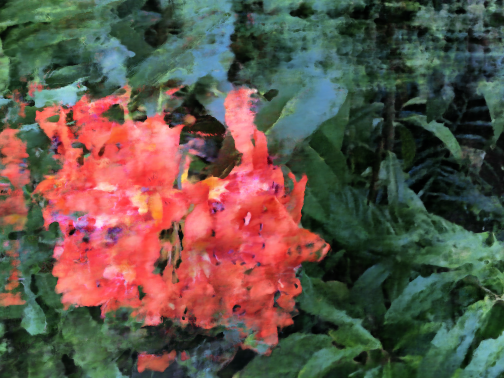}
    \end{subfigure}
    \begin{subfigure}[h]{0.162\linewidth}
        \centering
        \includegraphics[width=\textwidth]{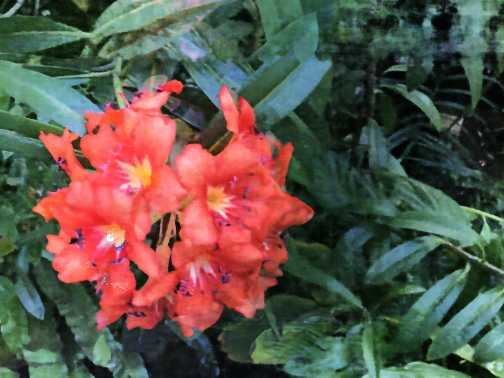}
    \end{subfigure} 
    \centering
    \begin{subfigure}[h]{0.162\linewidth}
        \centering
        \includegraphics[width=\textwidth]{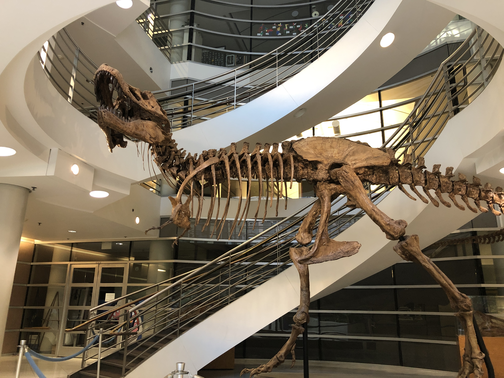}
    \end{subfigure}
    \begin{subfigure}[h]{0.162\linewidth}
         \centering
         \includegraphics[width=\textwidth]{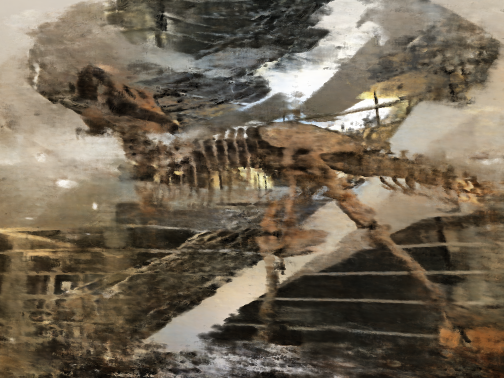}
    \end{subfigure}
    \begin{subfigure}[h]{0.162\linewidth}
        \centering
        \includegraphics[width=\textwidth]{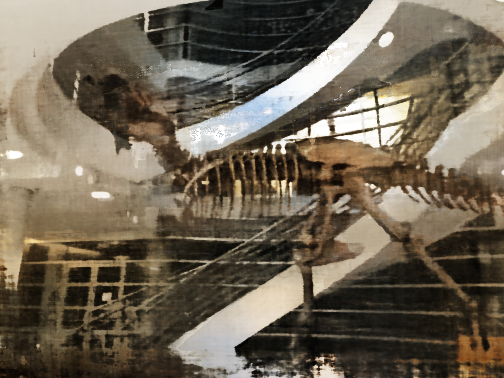}
    \end{subfigure}
    \\
    \begin{subfigure}[h]{0.162\linewidth}
        \centering
        \includegraphics[width=\textwidth]{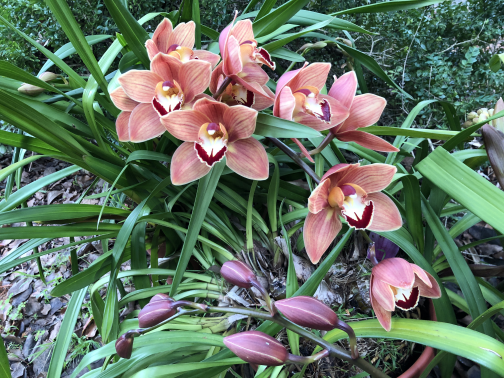}
        \caption{Ground-truth}
    \end{subfigure}
    \begin{subfigure}[h]{0.162\linewidth}
         \centering
         \includegraphics[width=\textwidth]{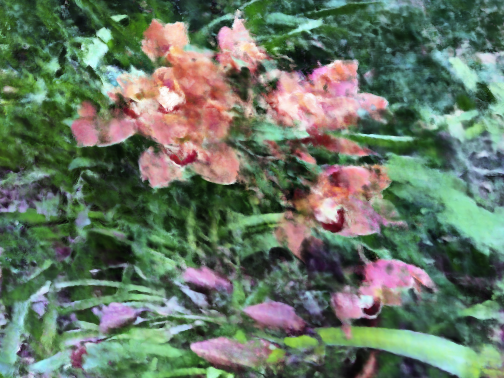}
         \caption{NeRF~\cite{mildenhall2020nerf}}
    \end{subfigure}
    \begin{subfigure}[h]{0.162\linewidth}
        \centering
        \includegraphics[width=\textwidth]{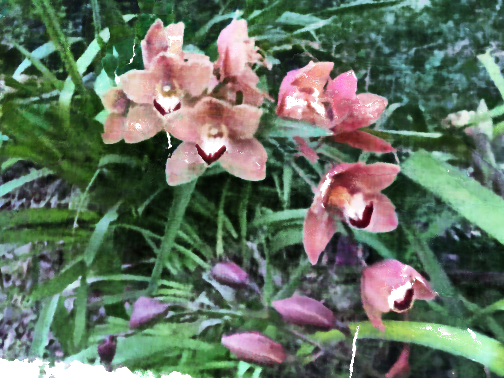}
        \caption{InfoNeRF}
    \end{subfigure} 
    \centering
    \begin{subfigure}[h]{0.162\linewidth}
        \centering
        \includegraphics[width=\textwidth]{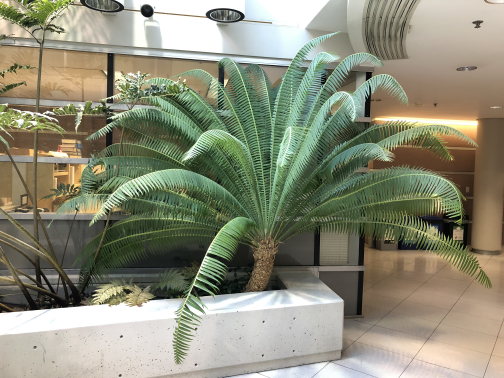}
        \caption{Ground-truth}
    \end{subfigure}
    \begin{subfigure}[h]{0.162\linewidth}
         \centering
         \includegraphics[width=\textwidth]{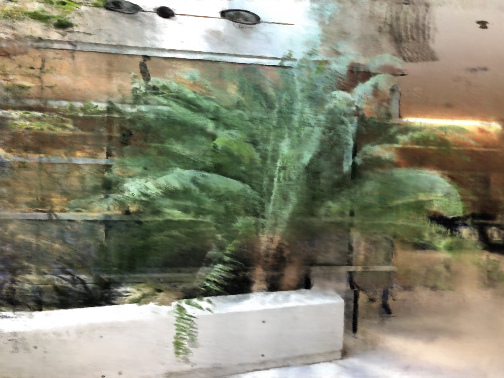}
         \caption{NeRF~\cite{mildenhall2020nerf}}
    \end{subfigure}
    \begin{subfigure}[h]{0.162\linewidth}
        \centering
        \includegraphics[width=\textwidth]{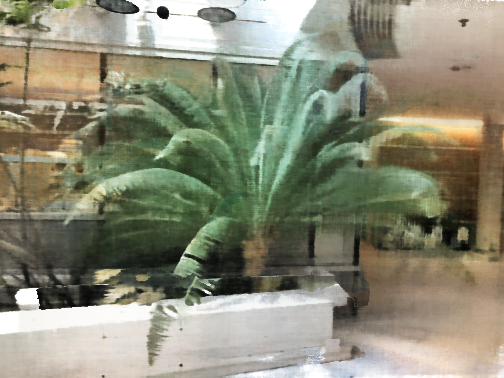}
        \caption{InfoNeRF}
    \end{subfigure} 
\caption{Qualitative comparisons on the \textit{Flower}, \textit{T-Rex}, \textit{Orchids}, and \textit{Fern} scenes of the LLFF dataset in the 2-view setting.}
\label{fig:llff_results}
\end{figure*}

To validate our algorithm in more challenging scenarios, we test InfoNeRF on a more complex real-world dataset, LLFF~\cite{mildenhall2019local}, which contains 8 natural scenes captured by a handheld cellphone.
Each scene has 20 to 62 images, and we hold out $\nicefrac{1}{8}$ of the images as test sets following the standard protocol~\cite{mildenhall2019local, mildenhall2020nerf}.
For training, we sample 2 views in each scene.
Table~\ref{tab:llff_more_complex} presents that our algorithm still outperforms the baseline by large margins and Figure~\ref{fig:llff_results} illustrates InfoNeRF provides better visual quality with less blurring artifact.
\begin{table}[h] 
	\centering
    	\caption{Results on a complex dataset, LLFF~\cite{mildenhall2019local} in the 2-view setting.}
   	 \scalebox{0.9}{
   	\setlength\tabcolsep{12pt} 
    	\begin{tabular}{c|ccc}
    	\toprule
	 &PSNR $\uparrow$  & SSIM $\uparrow$  & LPIPS $\downarrow$ \\
    	\hline
   	NeRF~\cite{mildenhall2020nerf}				&12.93	&0.267 	&0.554\\
    	InfoNeRF 				&\textbf{14.37}	&\textbf{0.349}	&\textbf{0.457}\\
    	\bottomrule
    	\end{tabular}}    
     	\label{tab:llff_more_complex}
	\vspace{-0.2cm}
\end{table}

\section{Role of $\mathcal{L}_\text{KL}$ in Narrow-Baseline Data}

\begin{figure}[h]
     \centering
     \begin{subfigure}[h]{0.49\linewidth}
         \centering
         \includegraphics[width=0.24\linewidth]{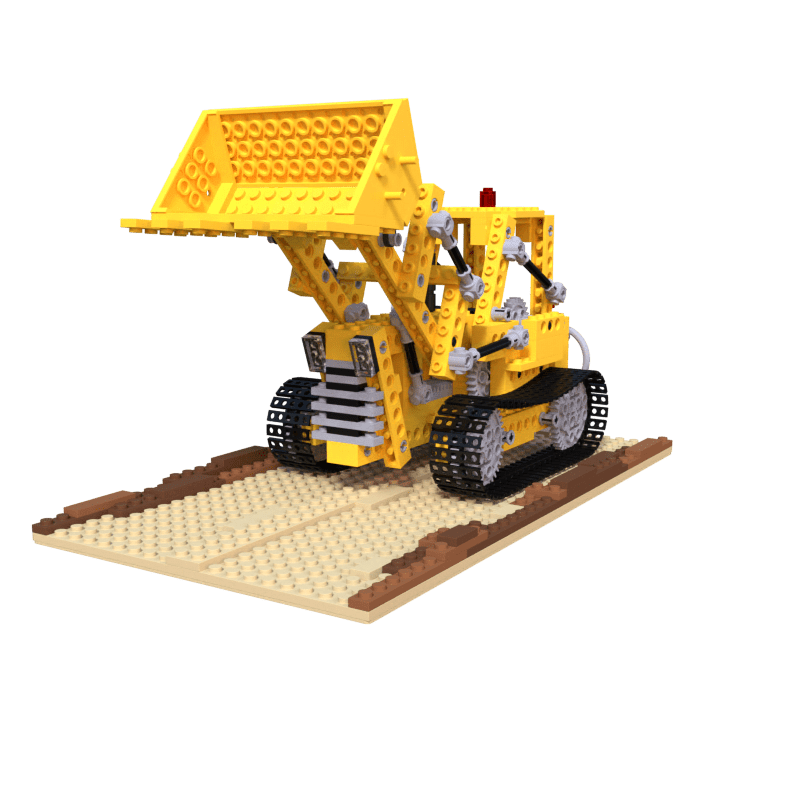} \hspace{-3mm}
         \includegraphics[width=0.24\linewidth]{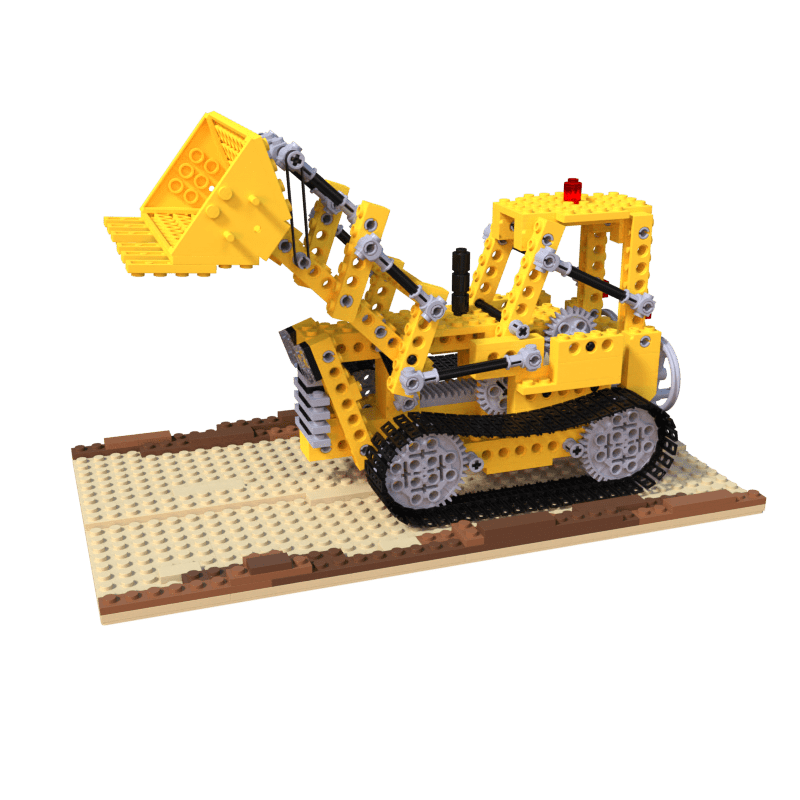} \hspace{-3mm} 
         \includegraphics[width=0.24\linewidth]{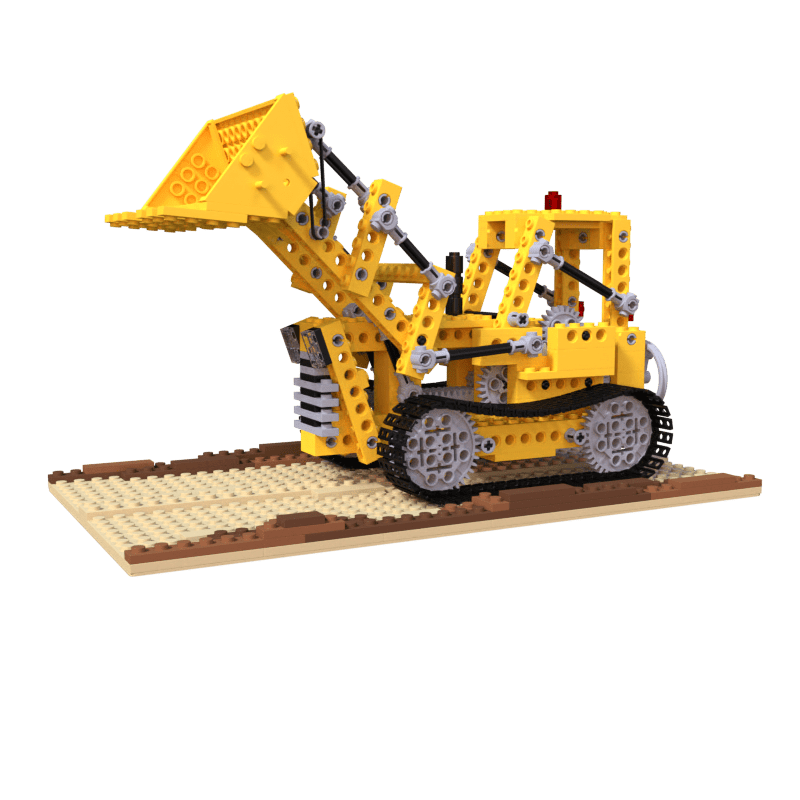} \hspace{-3mm} 
         \includegraphics[width=0.24\linewidth]{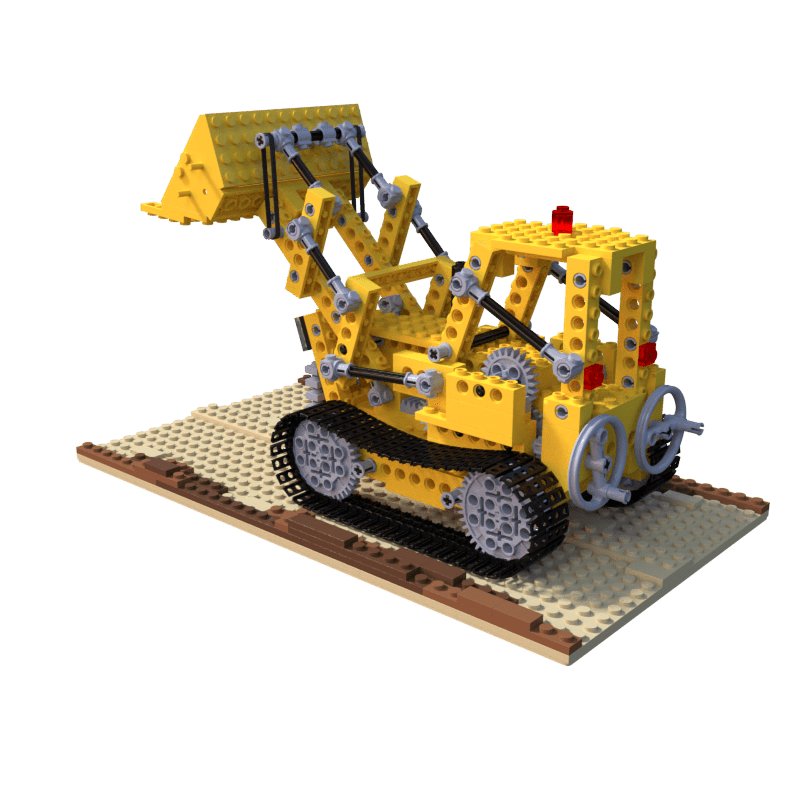} \vspace{-2mm} 
        \caption{{Sampled images in the narrow-baseline training set.}}
     \end{subfigure}
     \begin{subfigure}[h]{0.49\linewidth}
         \centering
         \includegraphics[width=0.24\linewidth]{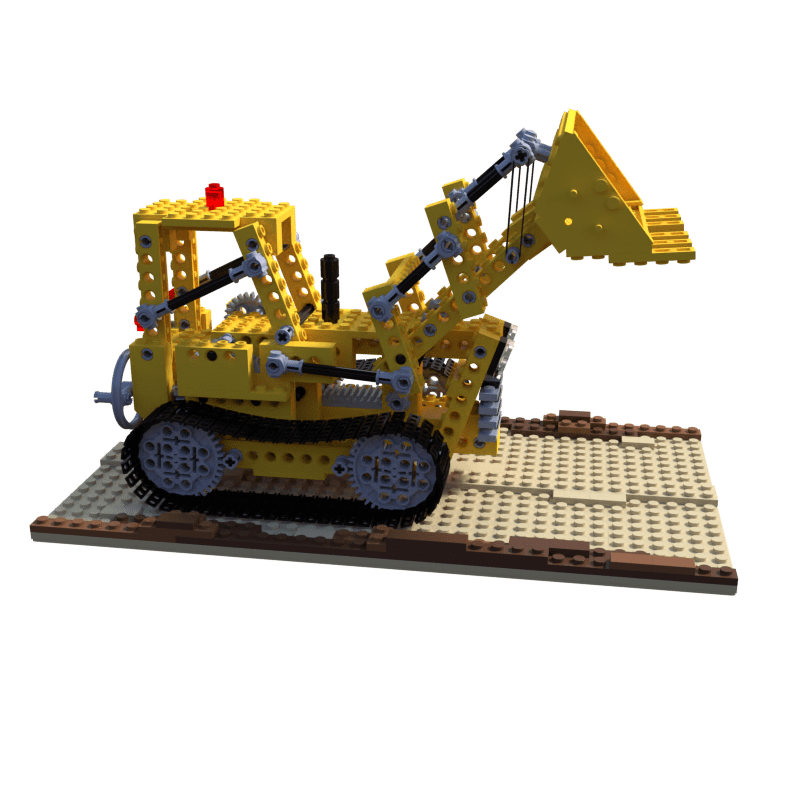} \hspace{-3mm} 
         \includegraphics[width=0.24\linewidth]{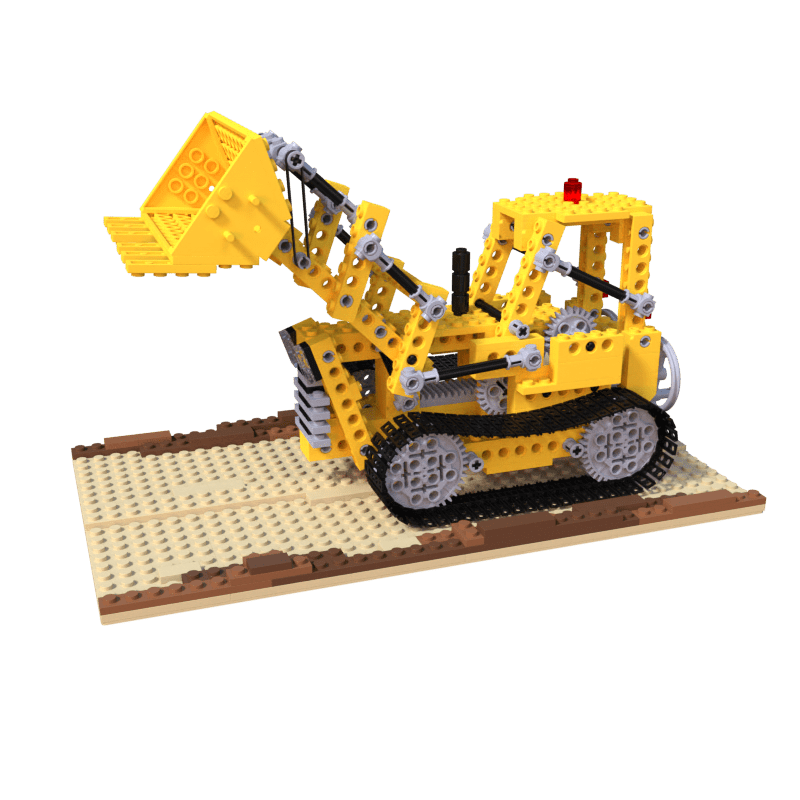} \hspace{-3mm} 
         \includegraphics[width=0.24\linewidth]{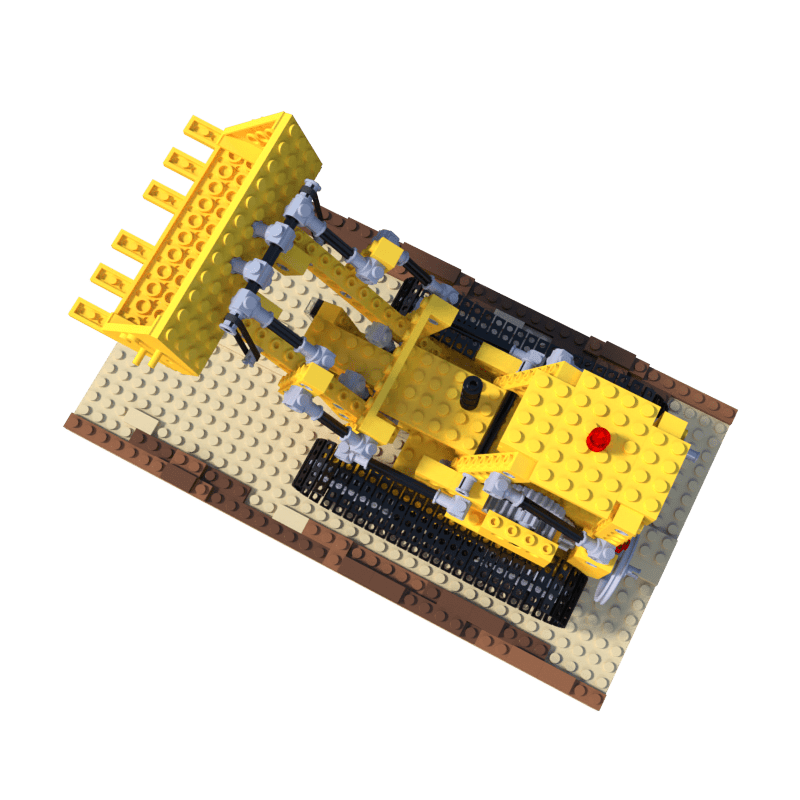} \hspace{-3mm} 
         \includegraphics[width=0.24\linewidth]{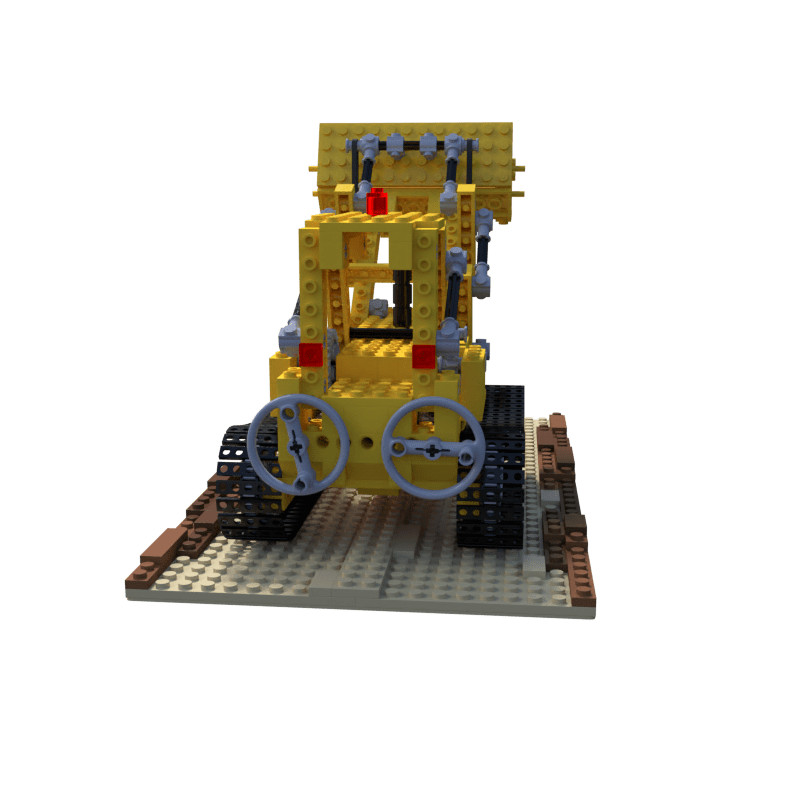} \vspace{-2mm}
        \caption{Sampled images in the wide-baseline training set.}
      \end{subfigure}
    \vspace{-2mm}
    \caption{Training set with the narrow and wide baselines sampled from the \textit{Lego} scene of the Realistic Synthetic 360$^{\circ}$ dataset in the 4-view setting.
    }
    \label{fig:sampling}
\end{figure}

InfoNeRF employs $\mathcal{L}_\text{KL}$ to prevent overfitting, especially for training with narrow-baseline images, and this loss term is particularly helpful for DTU.
To confirm the proposition, we create two sets of views in the \textit{Lego} scene of Realistic Synthetic 360$^{\circ}$ with narrow and wide baselines as illustrated in Figure~\ref{fig:sampling}.
Table~\ref{tab:viewpoint_variation} presents the full results in both settings; the combination of $\mathcal{L}_\text{KL}$ and $\mathcal{L}_\text{entropy}$ is effective for the narrow-baseline setting, as in the results on the DTU dataset in the Table 6 of the main paper, while degrading accuracy slightly for the wide-baseline setting.
Also, $\mathcal{L}_\text{KL}$ does not work well without $\mathcal{L}_\text{entropy}$ because incorporating the smoothness constraint between neighborhood rays is not helpful for noisy reconstructed scenes using the model trained without $\mathcal{L}_\text{entropy}$.

\begin{table}[h] 
	\centering
    	\caption{{Effects of the $\mathcal{L}_\text{KL}$ with respect to viewpoint variations on the \textit{Lego} scene of Realistic Synthetic 360$^{\circ}$ in the 4-view setting.}}
	\scalebox{0.9}{
   	\setlength\tabcolsep{8pt}
	\hspace{-0.2cm} 
    	\begin{tabular}{c|cc|ccc|ccc}
    	\toprule
    	\multirow{2}{*}{Method} &\multirow{2}{*}{$\mathcal{L}_\text{entropy}$} & \multirow{2}{*}{$\mathcal{L}_\text{KL}$}	&\multicolumn{3}{c|}{{narrow-baseline}}  	& \multicolumn{3}{c}{wide-baseline}   \\
 & &  &PSNR  & SSIM & LPIPS &PSNR  & SSIM & LPIPS \\
	\hline
   	NeRF~\cite{mildenhall2020nerf}	& &										& 13.16 	& 0.753 	&0.350 	&18.03	&0.763	&0.248\\
    	InfoNeRF w/o $\mathcal{L}_\text{entropy}$	&&\checkmark & 13.33 	& 0.750 	&0.349	&17.96	&0.753	&0.254\\
    	InfoNeRF w/o $\mathcal{L}_\text{KL}$		&\checkmark &			& 16.74 	& 0.767	&0.241	&\textbf{19.51}	&\textbf{0.792}	&\textbf{0.189}\\
    	InfoNeRF 					 			&\checkmark &\checkmark 			& \textbf{18.41}	& \textbf{0.775} &\textbf{0.199} 	&19.28 	&0.789	&0.190\\
    	\bottomrule
    	\end{tabular}}    
	\vspace{-3mm}
     	\label{tab:viewpoint_variation}
\end{table}

\begin{figure*}[h]
\centering
    \begin{subfigure}[h]{0.19\linewidth}
        \centering
        \includegraphics[width=1.0\textwidth]{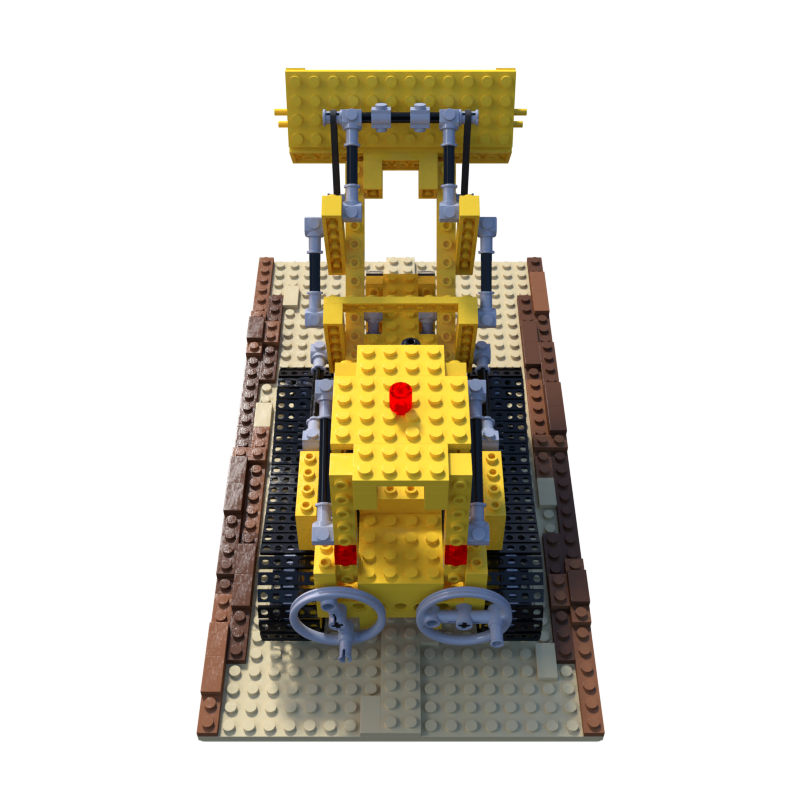}
        \caption{Ground-truth}
    \end{subfigure}~~
    \begin{subfigure}[h]{0.19\linewidth}
        \centering
        \includegraphics[width=1.0\textwidth]{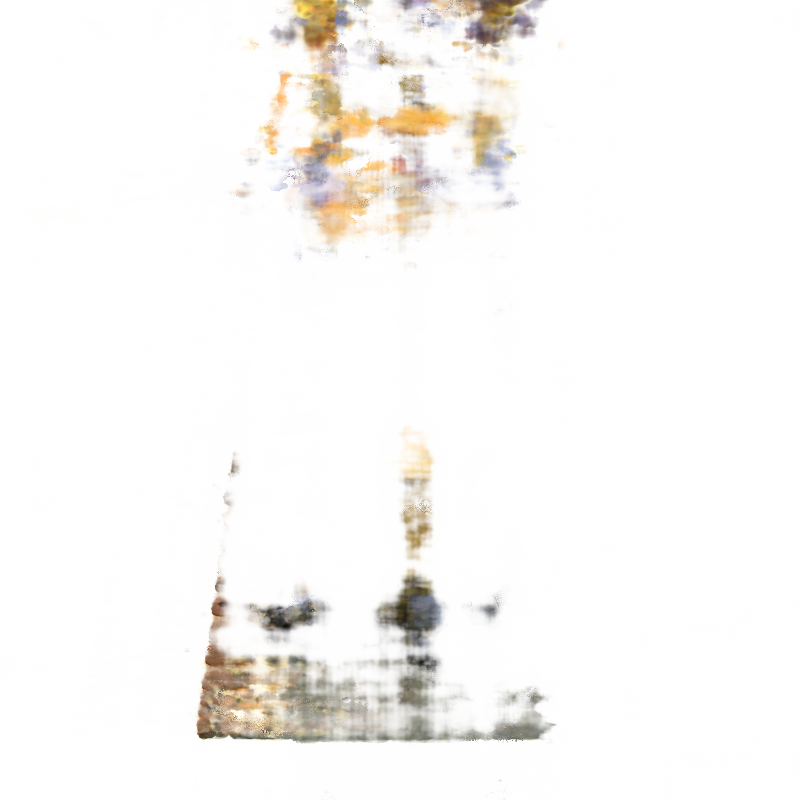}
        \caption{NeRF~\cite{mildenhall2020nerf}}
    \end{subfigure}~~
    \begin{subfigure}[h]{0.19\linewidth}
         \centering
         \includegraphics[width=1.0\textwidth]{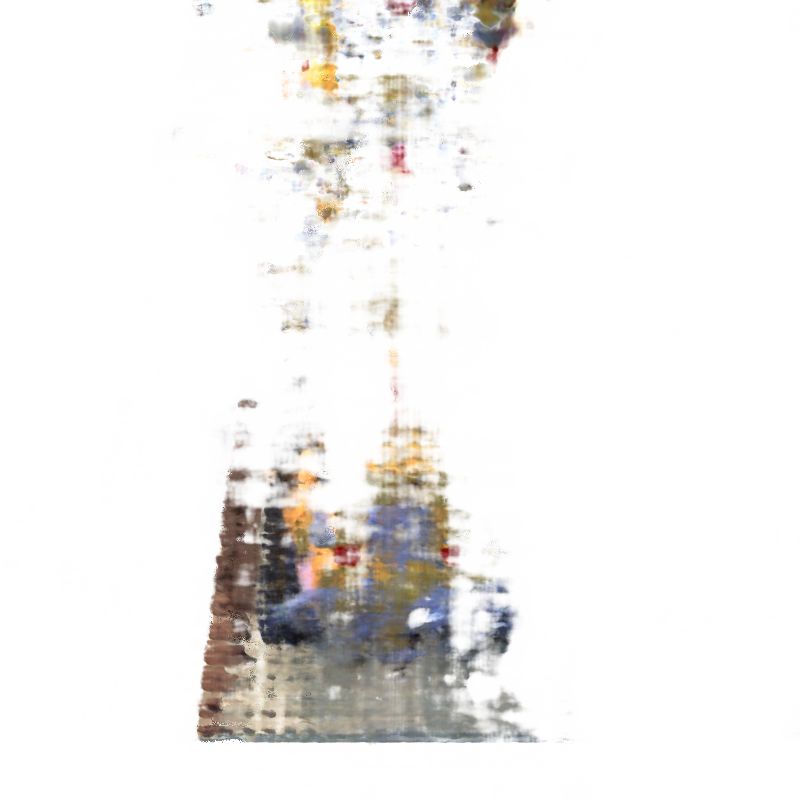}
         \caption{InfoNeRF w/o $\mathcal{L}_\text{entropy}$}
    \end{subfigure}~~
    \begin{subfigure}[h]{0.19\linewidth}
        \centering
        \includegraphics[width=1.0\textwidth]{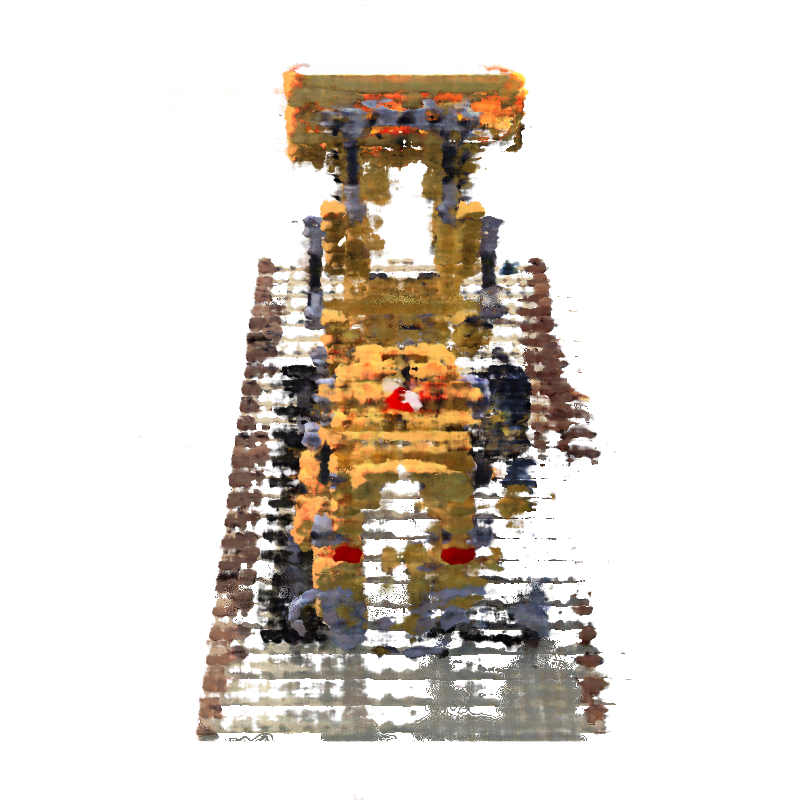}
        \caption{InfoNeRF w/o $\mathcal{L}_\text{KL}$}
    \end{subfigure}~~
    \begin{subfigure}[h]{0.19\linewidth}
        \centering
        \includegraphics[width=1.0\textwidth]{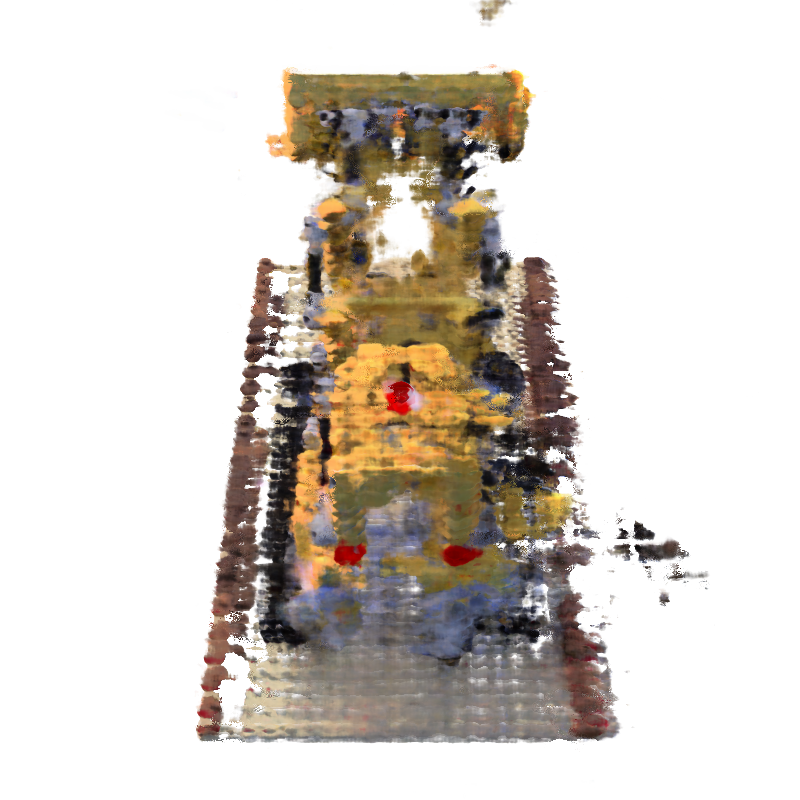}
        \caption{InfoNeRF}
    \end{subfigure} 
\caption{{Qualitative comparison between the reconstructed images for the  \textit{Lego} scene of the Realistic Synthetic 360$^{\circ}$ dataset in the narrow-baseline setting with 4 views.}}
\label{fig:sampling_results}
\end{figure*}

\clearpage
\section{Robustness to the Number of Training Views}
\begin{figure*}[h]
\centering
    \begin{subfigure}[h]{0.33\linewidth}
        \centering
        \includegraphics[width=\textwidth]{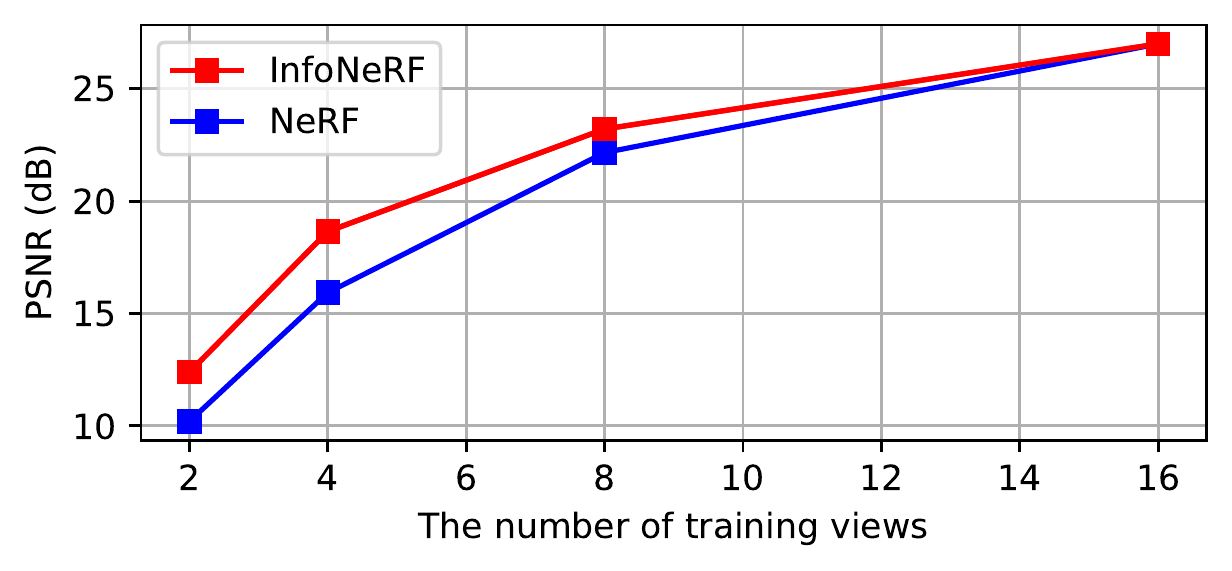}
        \caption{PSNR}
    \end{subfigure}~~
    \begin{subfigure}[h]{0.33\linewidth}
         \centering
         \includegraphics[width=\textwidth]{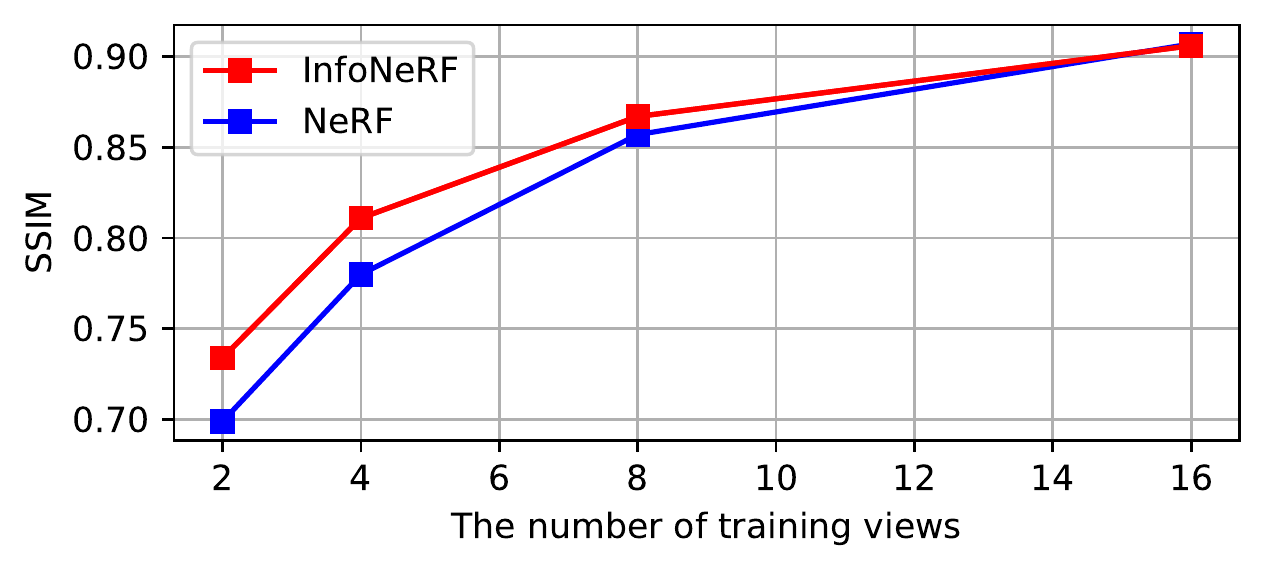}
         \caption{SSIM}
    \end{subfigure}~~
    \begin{subfigure}[h]{0.33\linewidth}
        \centering
        \includegraphics[width=\textwidth]{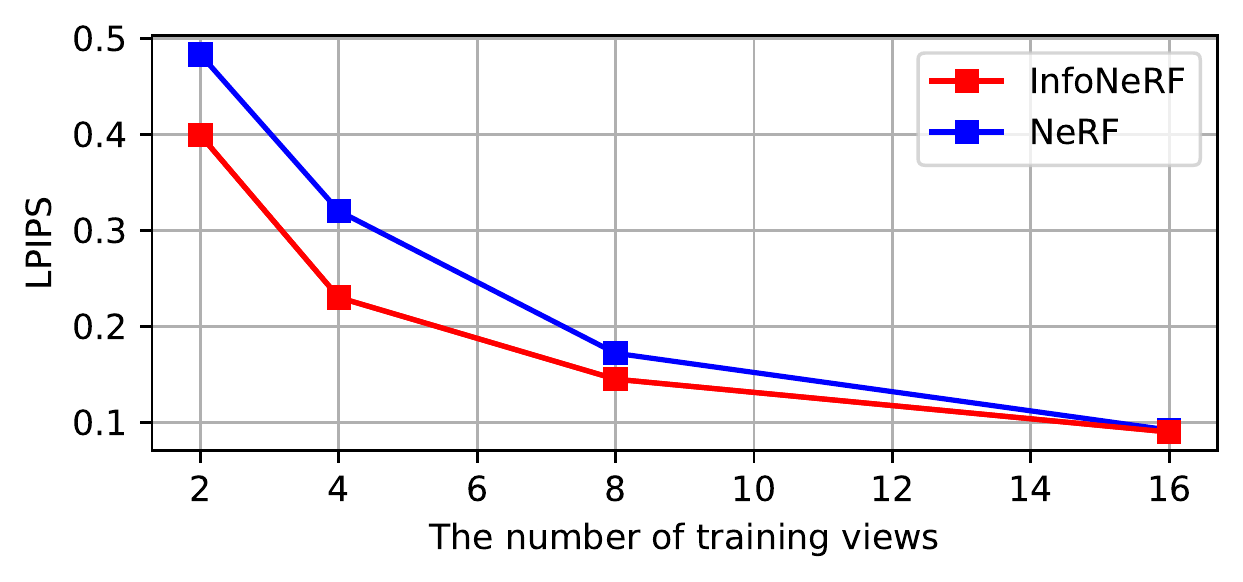}
        \caption{LPIPS}
    \end{subfigure} 
        \vspace{-0.2cm}
\caption{PSNR, SSIM and LPIPS results with respect to the number of training views on the Realistic Synthetic 360$^\circ$ dataset.}
\label{fig:varying_training_image_number}
\end{figure*}

{
Figure~\ref{fig:varying_training_image_number} demonstrates PSNR, SSIM, and LPIPS results of our model and the baseline by varying the number of training views on the Realistic Synthetic 360$^{\circ}$ dataset.
These graphs supplement Figure 6 of the main paper, showing the consistent tendency for all metrics.}

\begin{figure*}[h]
\centering
    \begin{subfigure}[h]{0.25\linewidth}
        \centering
        \includegraphics[width=\textwidth]{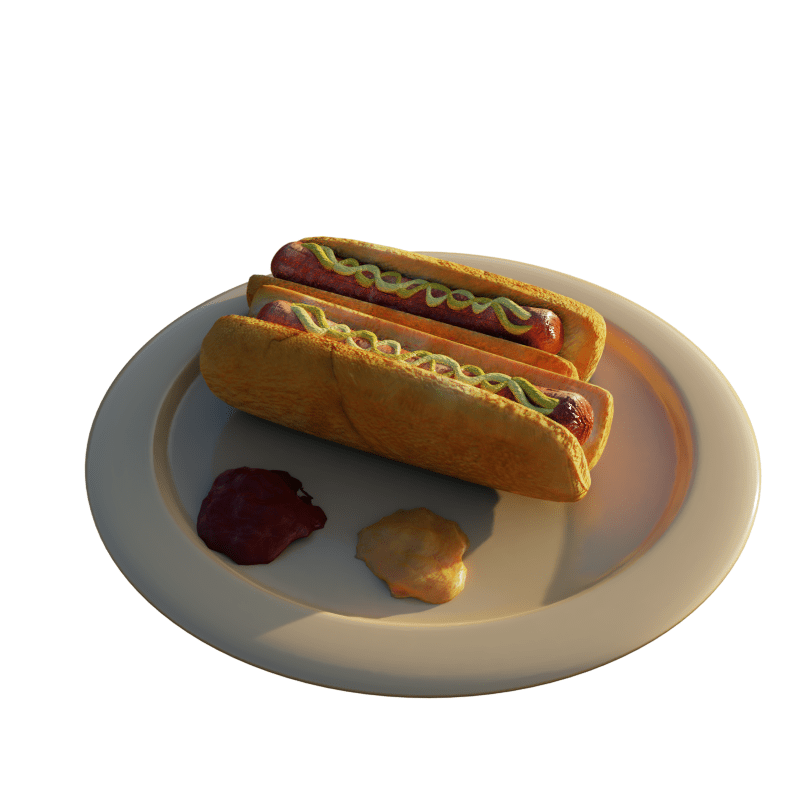}
        \caption{Ground-truth}
    \end{subfigure}
        \hspace{1cm}
    \begin{subfigure}[h]{0.25\linewidth}
         \centering
         \includegraphics[width=\textwidth]{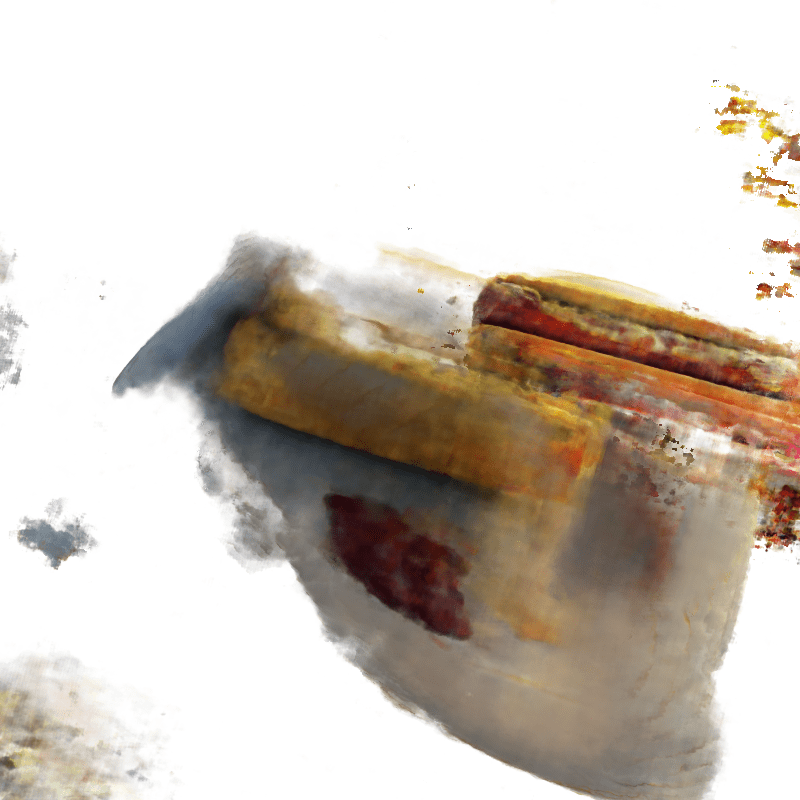}
         \caption{NeRF~\cite{mildenhall2020nerf}}
    \end{subfigure}
        \hspace{1cm}
    \begin{subfigure}[h]{0.25\linewidth}
        \centering
        \includegraphics[width=\textwidth]{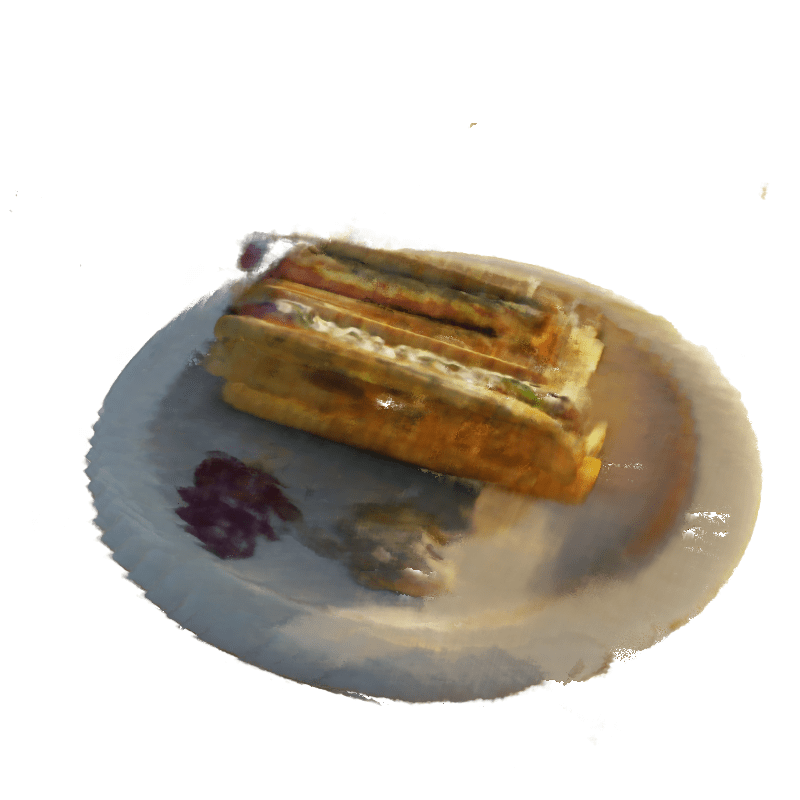}
        \caption{InfoNeRF}
    \end{subfigure} 
\caption{Qualitative comparison with the  \textit{HotDog} scene of the Realistic Synthetic 360$^{\circ}$ dataset in a 2-view setting.}
\label{fig:qual_results_synthetic_2views}
\end{figure*}

\section{Integration into Other NeRF-based Models}
{
To demonstrate the generality of our method, we incorporate the proposed regularization technique to PixelNeRF~\cite{yu2021pixelnerf} and mipNeRF~\cite{barron2021mip}, and refer to these versions of our model as InfoPixelNeRF and InfoMipNeRF, respectively.
Tables~\ref{tab:plug_in_synthetic_results} presents the quantitative results of both models together with their baselines, where we train all models from scratch with 4 views.
PixelNeRF and mipNeRF outperform NeRF, partly due to their semantic prior and cone-based positional encoding, respectively, but InfoPixelNeRF and InfoMipNeRF still achieve relative gains in terms of all metrics compared to their baselines.}

\begin{table*}[h]
\centering
        \caption{{Experimental results of InfoPixelNeRF and InfoMipNeRF in comparisons with their baselines in the 4-view setting on the Realistic Synthetic 360$^{\circ}$ dataset, where all results are based on 5 runs.}
    The asterisk ($\ast$) denotes that the model is pretrained on an external training dataset with dense input views and fine-tuned on this dataset with 4 views.}
\scalebox{0.9}{
    \setlength\tabcolsep{12pt} \hspace{-0.25cm}
    \begin{tabular}{ccccccc}
    \toprule
    Method  & PSNR  $\uparrow$ & SSIM $\uparrow$ & LPIPS $\downarrow$ & FID $\downarrow$ & {KID $\downarrow$}
    \\ 
    \hline 
    \hline
    NeRF~\cite{mildenhall2020nerf}  &15.93\footnotesize{$\pm$1.06} & 0.780\footnotesize{$\pm$0.014} & 0.320\footnotesize{$\pm$0.049} & 215.16\footnotesize{$\pm$2.32}&	0.0740\footnotesize{$\pm$0.0123}\\
    \hline
        PixelNeRF$^\ast$~\cite{yu2021pixelnerf} & 16.09\footnotesize{$\pm$0.78}&0.738\footnotesize{$\pm$0.012}&0.390\footnotesize{$\pm$0.030} &265.25\footnotesize{$\pm$6.73} 	&0.1274\footnotesize{$\pm$0.0063}\\ 
    InfoPixelNeRF$^\ast$ (ours) &\textbf{16.30\footnotesize{$\pm$0.80}}&	\textbf{0.745\footnotesize{$\pm$0.015}}&	\textbf{0.372\footnotesize{$\pm$0.030}} & \textbf{264.33\footnotesize{$\pm$6.21}} & \textbf{0.1263\footnotesize{$\pm$0.0058}}\\
        \hdashline
    mipNeRF~\cite{barron2021mip}  &   19.04\footnotesize{$\pm$0.02} & 0.814\footnotesize{$\pm$0.000} & 0.219\footnotesize{$\pm$0.000} &142.96\footnotesize{$\pm$1.89}	&{0.0422\footnotesize{$\pm$0.0010} }\\
    InfoMipNeRF (ours) & \textbf{19.33\footnotesize{$\pm$0.04}} & \textbf{0.818\footnotesize{$\pm$0.000}} & \textbf{0.213\footnotesize{$\pm$0.001}}  &\textbf{142.95\footnotesize{$\pm$0.54}	}&\textbf{0.0418\footnotesize{$\pm$0.0004}}\\    \bottomrule
    \end{tabular}}
       \vspace{-0.2cm}
    \label{tab:plug_in_synthetic_results}
   \vspace{+0.4cm}
\end{table*}
 
 \newpage
\section{Per-Scene Breakdown on the Realistic Synthetic 360$^\circ$ Dataset}
Supplementing Table 1 of the main paper, we show the experimental results from individual scenes in terms of PSNR, SSIM, and LPIPS in Table~\ref{tab:synthetic_psnr}, \ref{tab:synthetic_ssim}, and \ref{tab:synthetic_lpips}, respectively.
Table~\ref{tab:synthetic_results} is the same as Table 1 of the main paper, added for reference.
As shown in the tables, InfoNeRF achieves consistent and non-trivial improvement over its baselines.

\vspace{0.4cm}

\begin{table*}[h]
\centering
        \caption{Experimental results of few-shot novel view synthesis on the Realistic Synthetic 360$^{\circ}$ dataset in the 4-view setting.
Our approach outperforms all other existing methods by significant margins in all image quality metrics.
The asterisk ($\ast$) denotes that the model is pretrained on an external training dataset with dense input views and finet-uned on this dataset with 4 input views.
We run all experimental five times with different viewpoint samples and the same hyperparameters, and compute the average scores and their standard deviations.}
\vspace{-2mm}
\scalebox{0.9}{
    \setlength\tabcolsep{12pt} \hspace{-0.25cm}
    \begin{tabular}{ccccccc}
    \toprule
    Method  & PSNR  $\uparrow$ & SSIM $\uparrow$ & LPIPS $\downarrow$ & FID $\downarrow$ & KID $\downarrow$
    \\ 
    \hline 
    \hline
        NeRF, 100 views  &  31.01 &0.947 & 0.081 & 42.83 & 0.002  \\ \hline
            PixelNeRF$^\ast$~\cite{yu2021pixelnerf} & 16.09\footnotesize{$\pm$0.78}&0.738\footnotesize{$\pm$0.012}&0.390\footnotesize{$\pm$0.030}  &265.25\footnotesize{$\pm$6.73} 	&0.127\footnotesize{$\pm$0.006}\\  \hdashline
    NeRF~\cite{mildenhall2020nerf}  &15.93\footnotesize{$\pm$1.06} & 0.780\footnotesize{$\pm$0.014} & 0.320\footnotesize{$\pm$0.049} & 215.16\footnotesize{$\pm$2.32}&	0.074\footnotesize{$\pm$0.012}\\
    DietNeRF~\cite{jain2021putting} &16.06\footnotesize{$\pm$1.13} & 0.793\footnotesize{$\pm$0.019} & 0.306\footnotesize{$\pm$0.050} & \ 197.02\footnotesize{$\pm$12.87}&	0.065\footnotesize{$\pm$0.004}\\
    InfoNeRF (ours)  & \textbf{18.65\footnotesize{$\pm$0.18}} & \textbf{0.811\footnotesize{$\pm$0.008}} & \textbf{0.230\footnotesize{$\pm$0.008}} & \textbf{181.47\footnotesize{$\pm$4.97}}&	\textbf{0.062\footnotesize{$\pm$0.004}}\\ 
    \bottomrule
    \end{tabular}}
    \vspace{5mm}
    \label{tab:synthetic_results}

\centering
    \caption{Average PSNRs and standard deviations of individual scenes on the Realistic Synthetic 360$^{\circ}$ dataset in the 4-view setting.} 
\scalebox{0.9}{
    \setlength\tabcolsep{4.5pt} \hspace{-0.25cm}
    \begin{tabular}{ccccccccc|c}
    \toprule
    Method & Lego & Chair & Drums & Ficus & Hotdog & Materials& Mic& Ship & Avg.
    \\ 
    \hline \hline
    NeRF, 100 views				&32.54&	33.00&	25.01&	30.13&	36.18&	29.62&	32.91&	28.65&	31.01\\ 
   \hline
       PixelNeRF$^\ast$~\cite{yu2021pixelnerf} 	&15.14\footnotesize{$\pm$0.75}&	18.87\footnotesize{$\pm$1.38}&	15.10\footnotesize{$\pm$0.63}&	16.60\footnotesize{$\pm$0.70}&	19.37\footnotesize{$\pm$1.78}&	12.31\footnotesize{$\pm$1.02}&	16.35\footnotesize{$\pm$0.97}&	14.96\footnotesize{$\pm$0.75}&	16.09\footnotesize{$\pm$0.78}  \\
       \hdashline
    NeRF~\cite{mildenhall2020nerf} 	&15.61\footnotesize{$\pm$4.53}&	18.57\footnotesize{$\pm$1.64}&	12.50\footnotesize{$\pm$0.98}&	16.37\footnotesize{$\pm$2.24}&	19.64\footnotesize{$\pm$2.26}&	15.65\footnotesize{$\pm$4.16}&	14.78\footnotesize{$\pm$2.37}&	14.30\footnotesize{$\pm$4.04}&	15.93\footnotesize{$\pm$1.06}\\    	
    DietNeRF~\cite{jain2021putting} 		&17.13\footnotesize{$\pm$4.77}&	19.37\footnotesize{$\pm$3.12}&	13.74\footnotesize{$\pm$1.55}&	15.76\footnotesize{$\pm$3.56}&	18.24\footnotesize{$\pm$5.28}&	15.00\footnotesize{$\pm$5.18}&	17.71\footnotesize{$\pm$1.55}&	11.51\footnotesize{$\pm$4.27}&	16.06\footnotesize{$\pm$1.13}\\
  InfoNeRF (ours) 					&\textbf{18.92\footnotesize{$\pm$0.51}}&	\textbf{20.06\footnotesize{$\pm$1.11}}&	\textbf{14.33\footnotesize{$\pm$0.62}}&	\textbf{19.41\footnotesize{$\pm$0.07}}&	\textbf{21.30\footnotesize{$\pm$2.31}}&	\textbf{18.34\footnotesize{$\pm$0.88}}&	\textbf{18.55\footnotesize{$\pm$1.71}}&	\textbf{18.27\footnotesize{$\pm$0.71}}&	\textbf{18.65\footnotesize{$\pm$0.18}}\\
   \bottomrule
    \end{tabular}
   }
   \vspace{5mm}
    \label{tab:synthetic_psnr}
\centering
    \caption{Average SSIMs and standard deviations of individual scenes on the Realistic Synthetic 360$^{\circ}$ dataset in the 4-view setting.}
\scalebox{0.9}{
    \setlength\tabcolsep{3pt} \hspace{-0.25cm}
    \begin{tabular}{ccccccccc|c}
    \toprule
    Method & Lego & Chair & Drums & Ficus & Hotdog & Materials& Mic& Ship & Avg.
    \\ 
    \hline \hline
    NeRF, 100 views 	&	0.961&	0.967&	0.925&	0.964&	0.974&	0.949&	0.980&	0.856&	0.947\\ \hline     
            PixelNeRF$^\ast$~\cite{yu2021pixelnerf} 	&0.703\footnotesize{$\pm$0.014}&	0.802\footnotesize{$\pm$0.026}&	0.699\footnotesize{$\pm$0.016}&	0.802\footnotesize{$\pm$0.017}&	0.836\footnotesize{$\pm$0.023}&	0.644\footnotesize{$\pm$0.027}&	0.767\footnotesize{$\pm$0.021}&	0.655\footnotesize{$\pm$0.014}&	0.738\footnotesize{$\pm$0.012}\\ \hdashline
    NeRF~\cite{mildenhall2020nerf} 	&	0.739\footnotesize{$\pm$0.065}&	0.818\footnotesize{$\pm$0.019}&	0.721\footnotesize{$\pm$0.022}&	0.833\footnotesize{$\pm$0.030}&	0.863\footnotesize{$\pm$0.024}&	0.768\footnotesize{$\pm$0.070}&	0.826\footnotesize{$\pm$0.043}&	0.675\footnotesize{$\pm$0.047}&	0.780\footnotesize{$\pm$0.014}\\    	
    DietNeRF~\cite{jain2021putting} 		&0.766\footnotesize{$\pm$0.079}&	\textbf{0.846\footnotesize{$\pm$0.022}}&	\textbf{0.750\footnotesize{$\pm$0.021}}&	0.812\footnotesize{$\pm$0.046}&	0.851\footnotesize{$\pm$0.070}&	0.789\footnotesize{$\pm$0.050}&	0.879\footnotesize{$\pm$0.028}&	0.649\footnotesize{$\pm$0.057}&	0.793\footnotesize{$\pm$0.019}\\
    InfoNeRF (ours) 			&\textbf{0.788\footnotesize{$\pm$0.008}}&	0.840\footnotesize{$\pm$0.011}&	0.730\footnotesize{$\pm$0.015}&	\textbf{0.851\footnotesize{$\pm$0.001}}&	\textbf{0.871\footnotesize{$\pm$0.027}}&\textbf{	0.799\footnotesize{$\pm$0.052}}&	\textbf{0.883\footnotesize{$\pm$0.012}}&	\textbf{0.723\footnotesize{$\pm$0.012}}&	\textbf{0.811\footnotesize{$\pm$0.008}}\\ 
    \bottomrule
    \end{tabular}}
    \vspace{5mm}
    \label{tab:synthetic_ssim}
\centering
    \caption{Average LPIPS's and standard deviations of individual scenes on the Realistic Synthetic 360$^{\circ}$ dataset in the 4-view setting.}
\scalebox{0.9}{
    \setlength\tabcolsep{3pt} \hspace{-0.25cm}
    \begin{tabular}{ccccccccc|c}
    \toprule
    Method & Lego & Chair & Drums & Ficus & Hotdog & Materials& Mic& Ship & Avg.
    \\ 
    \hline \hline
    NeRF (100 views)		&0.050&	0.046&	0.091&	0.044&	0.121&	0.063&	0.028&	0.206&	0.081	\\ \hline     
                    PixelNeRF$^\ast$~\cite{yu2021pixelnerf} 	&0.410\footnotesize{$\pm$0.031}&	0.325\footnotesize{$\pm$0.065}&	0.462\footnotesize{$\pm$0.026}&	0.326\footnotesize{$\pm$0.039}&	0.266\footnotesize{$\pm$0.050}&	0.490\footnotesize{$\pm$0.034}&	0.395\footnotesize{$\pm$0.033}&	0.446\footnotesize{$\pm$0.037}&	0.390\footnotesize{$\pm$0.030} \\ 
     \hdashline
    NeRF~\cite{mildenhall2020nerf} 	&	0.318\footnotesize{$\pm$0.149}&	0.284\footnotesize{$\pm$0.047}&	0.452\footnotesize{$\pm$0.057}&	0.224\footnotesize{$\pm$0.089}&	0.232\footnotesize{$\pm$0.049}&	0.294\footnotesize{$\pm$0.178}&	0.351\footnotesize{$\pm$0.094}&	0.412\footnotesize{$\pm$0.095}&	0.320\footnotesize{$\pm$0.049}\\    	
    DietNeRF~\cite{jain2021putting} 	&0.285\footnotesize{$\pm$0.178}&	0.314\footnotesize{$\pm$0.152}&	\textbf{0.304\footnotesize{$\pm$0.094}}&	0.315\footnotesize{$\pm$0.173}&	0.229\footnotesize{$\pm$0.113}&	0.324\footnotesize{$\pm$0.200}&	0.210\footnotesize{$\pm$0.059}&	0.464\footnotesize{$\pm$0.137}&	0.306\footnotesize{$\pm$0.050}\\
    InfoNeRF (ours) 			&\textbf{0.182\footnotesize{$\pm$0.020}}&	\textbf{0.196\footnotesize{$\pm$0.016}}&	0.374\footnotesize{$\pm$0.026}&	\textbf{0.148\footnotesize{$\pm$0.011}}&	\textbf{0.188\footnotesize{$\pm$0.044}}&\textbf{	0.218\footnotesize{$\pm$0.073}}&	\textbf{0.207\footnotesize{$\pm$0.038}}&	\textbf{0.324\footnotesize{$\pm$0.015}}&	\textbf{0.230\footnotesize{$\pm$0.008}}\\
    \bottomrule
    \end{tabular}}
    \vspace{5mm}
    \label{tab:synthetic_lpips}
 \end{table*}

\clearpage
\section{Additional Qualitative Results}
\vspace{0.3cm}
\subsection{Realistic Synthetic 360$^{\circ}$}
\vspace{0.3cm}
\begin{figure*}[h]
    \centering
    \begin{subfigure}[h]{0.162\linewidth}
        \centering
        \includegraphics[width=\textwidth]{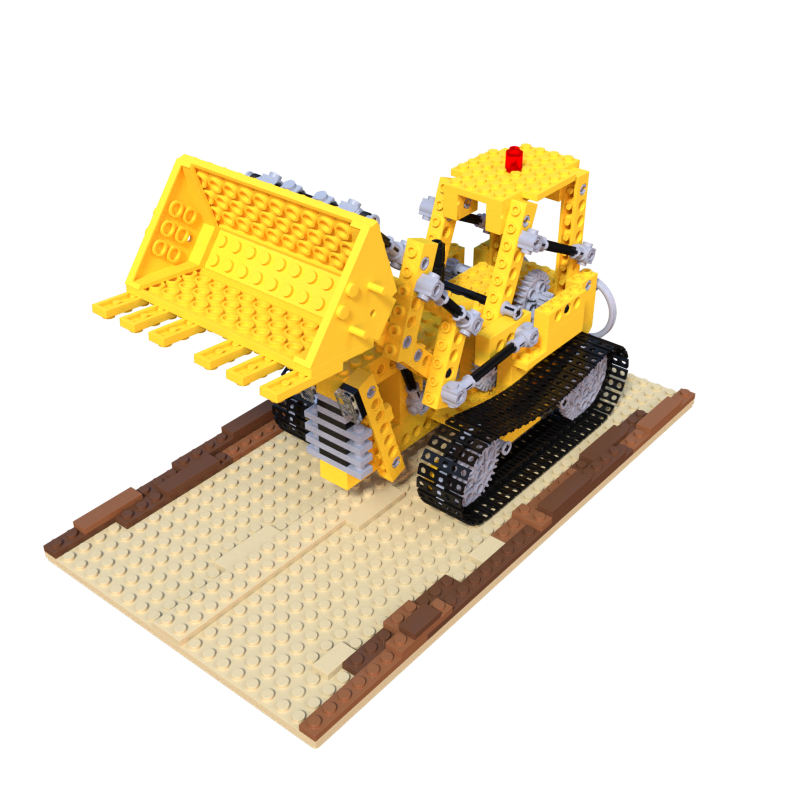}
    \end{subfigure}
    \begin{subfigure}[h]{0.162\linewidth}
        \centering
        \includegraphics[width=\textwidth]{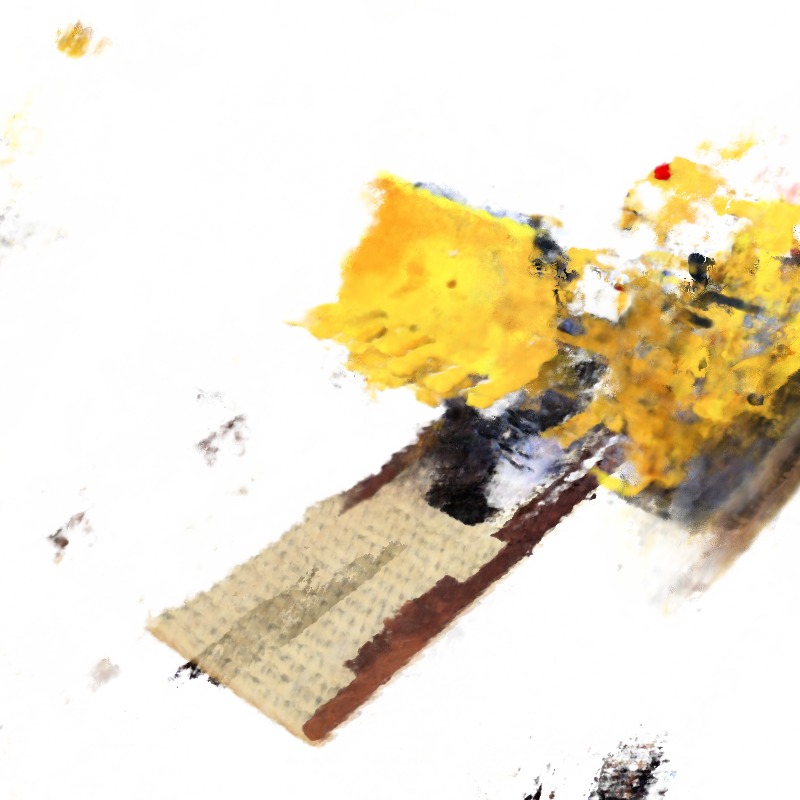}
    \end{subfigure}
    \begin{subfigure}[h]{0.162\linewidth}
        \centering
        \includegraphics[width=\textwidth]{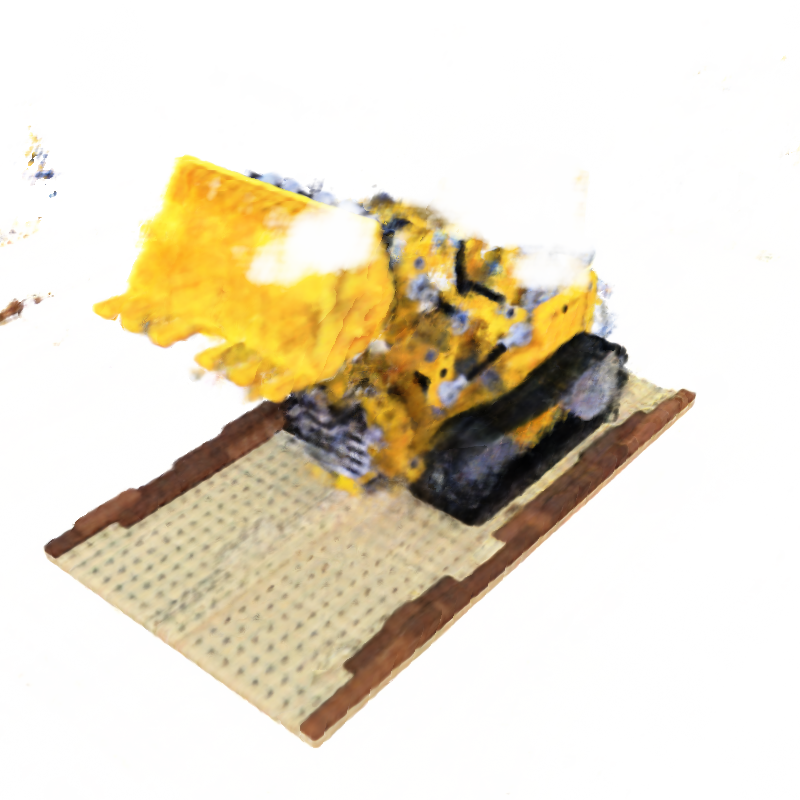}
    \end{subfigure}
    \begin{subfigure}[h]{0.162\linewidth}
        \centering
        \includegraphics[width=\textwidth]{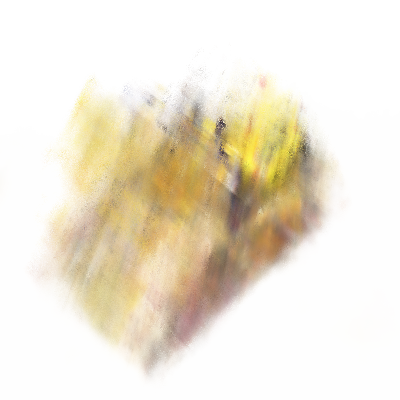}
    \end{subfigure}
    \begin{subfigure}[h]{0.162\linewidth}
        \centering
        \includegraphics[width=\textwidth]{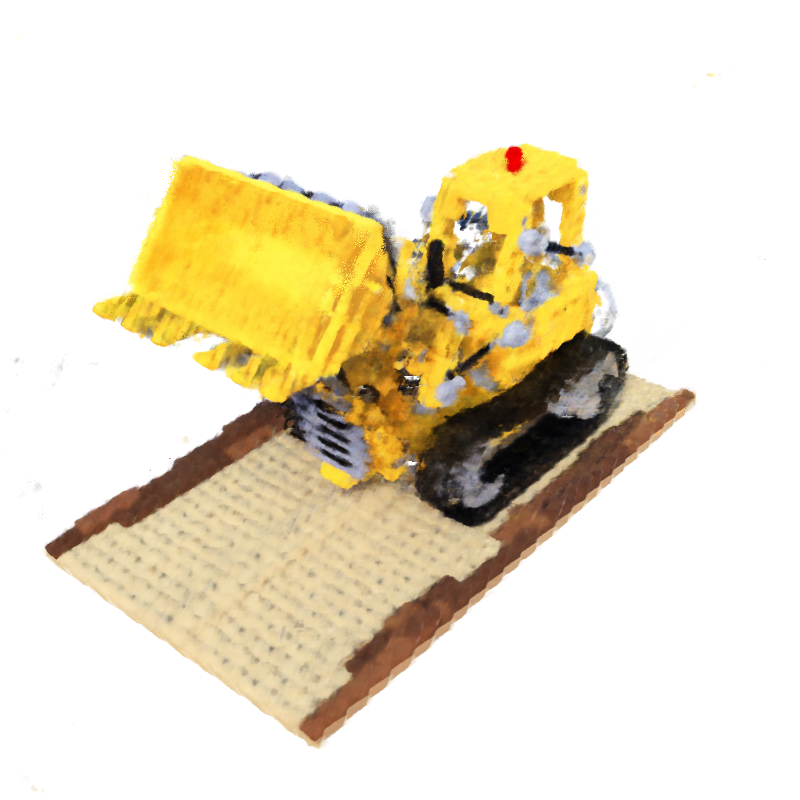}
    \end{subfigure}
    \begin{subfigure}[h]{0.162\linewidth}
        \centering
        \includegraphics[width=\textwidth]{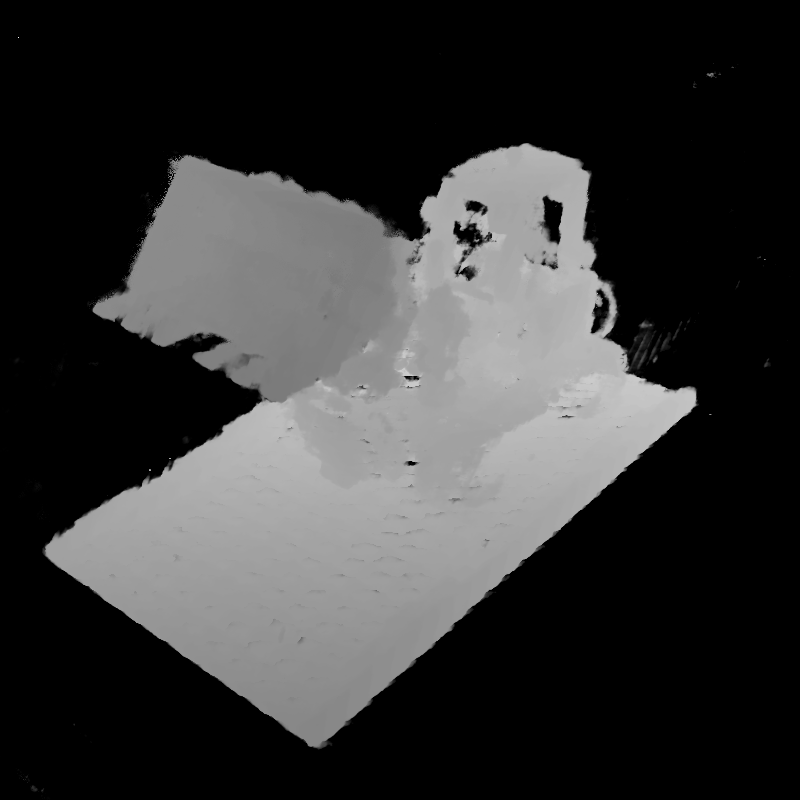}
    \end{subfigure}
        \begin{subfigure}[h]{0.162\linewidth}
        \centering
        \includegraphics[width=\textwidth]{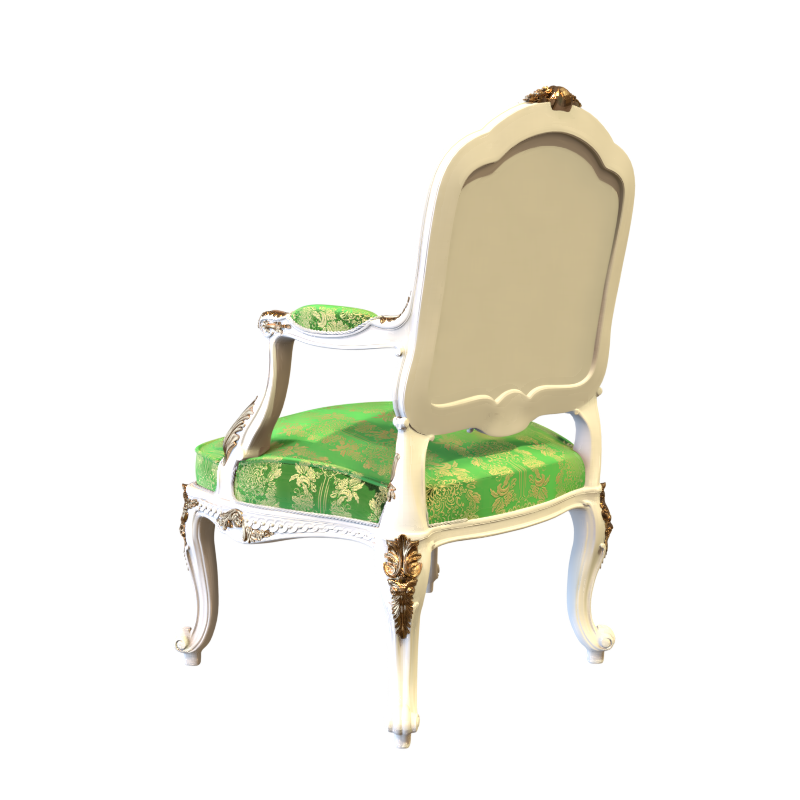}
    \end{subfigure}
    \begin{subfigure}[h]{0.162\linewidth}
        \centering
        \includegraphics[width=\textwidth]{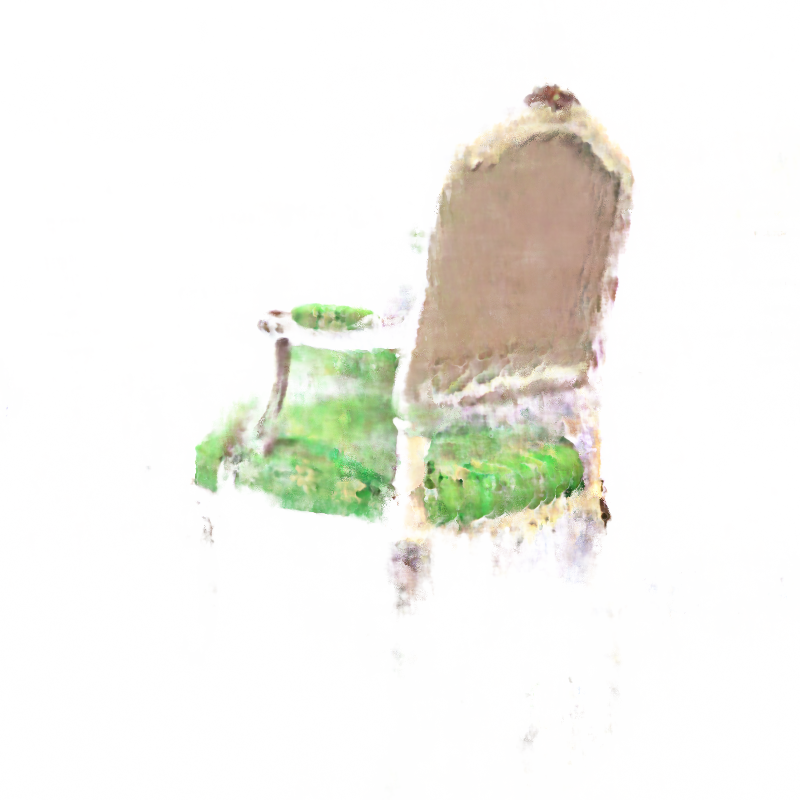}
    \end{subfigure}
    \begin{subfigure}[h]{0.162\linewidth}
        \centering
        \includegraphics[width=\textwidth]{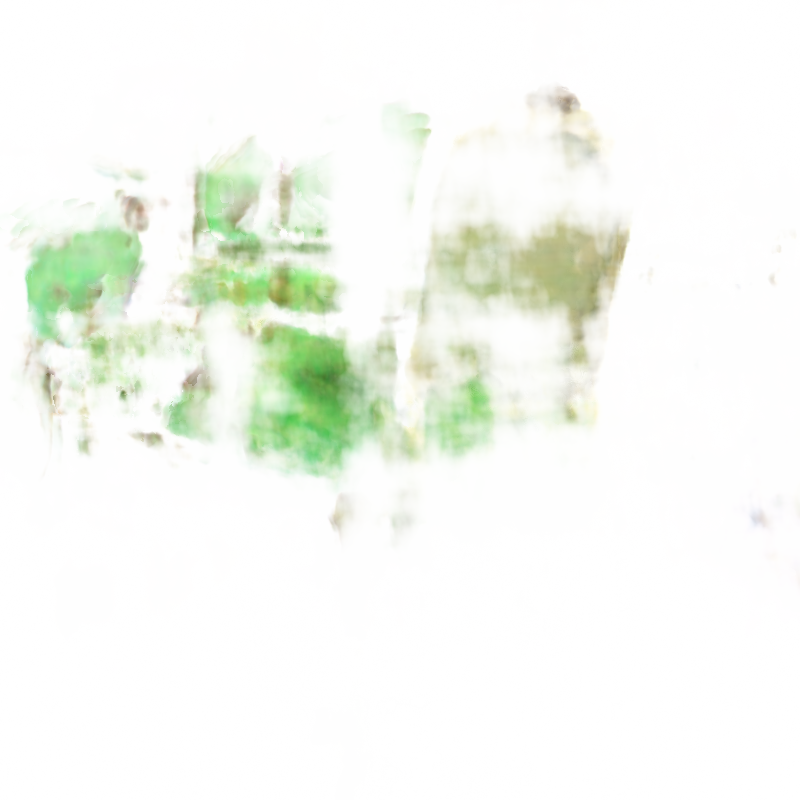}
    \end{subfigure}
    \begin{subfigure}[h]{0.162\linewidth}
        \centering
        \includegraphics[width=\textwidth]{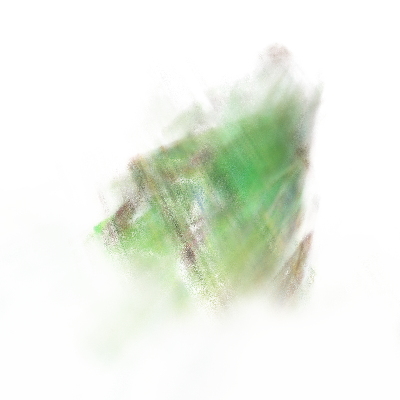}
    \end{subfigure}
    \begin{subfigure}[h]{0.162\linewidth}
        \centering
        \includegraphics[width=\textwidth]{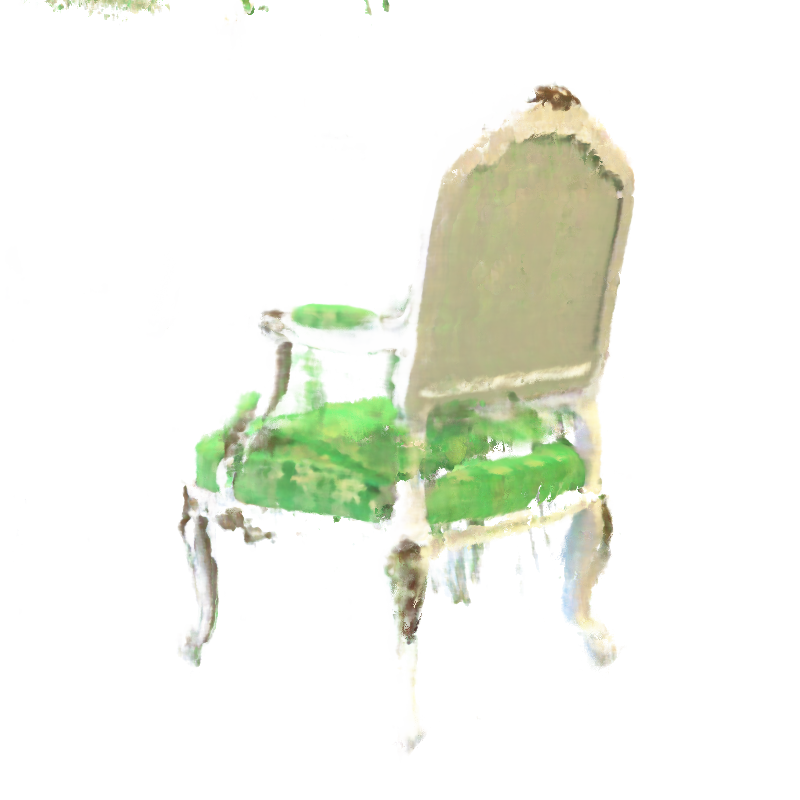}
    \end{subfigure}
    \begin{subfigure}[h]{0.162\linewidth}
        \centering
        \includegraphics[width=\textwidth]{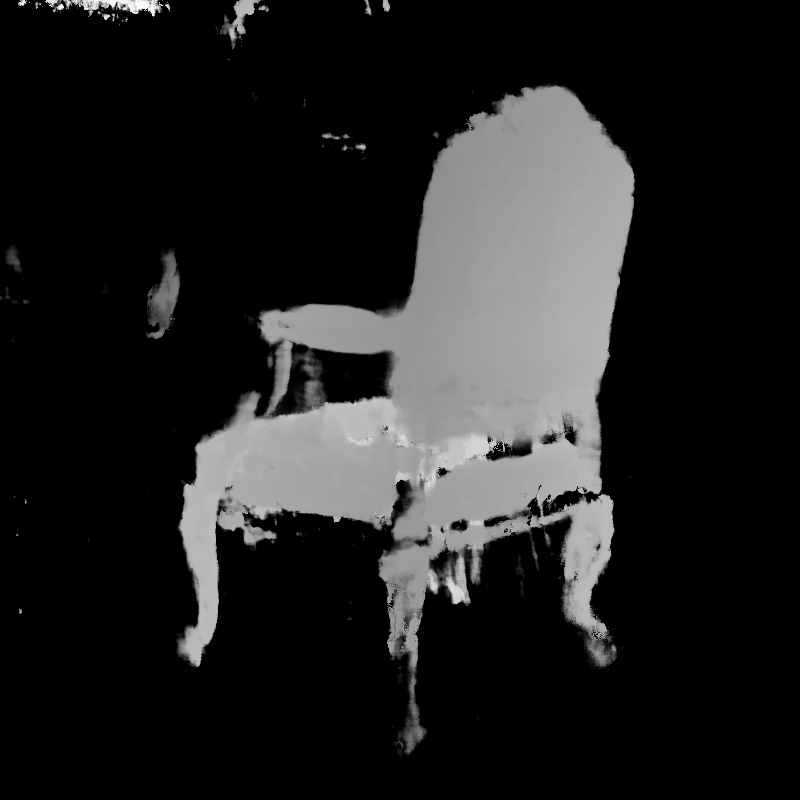}
    \end{subfigure}
            \begin{subfigure}[h]{0.162\linewidth}
        \centering
        \includegraphics[width=\textwidth]{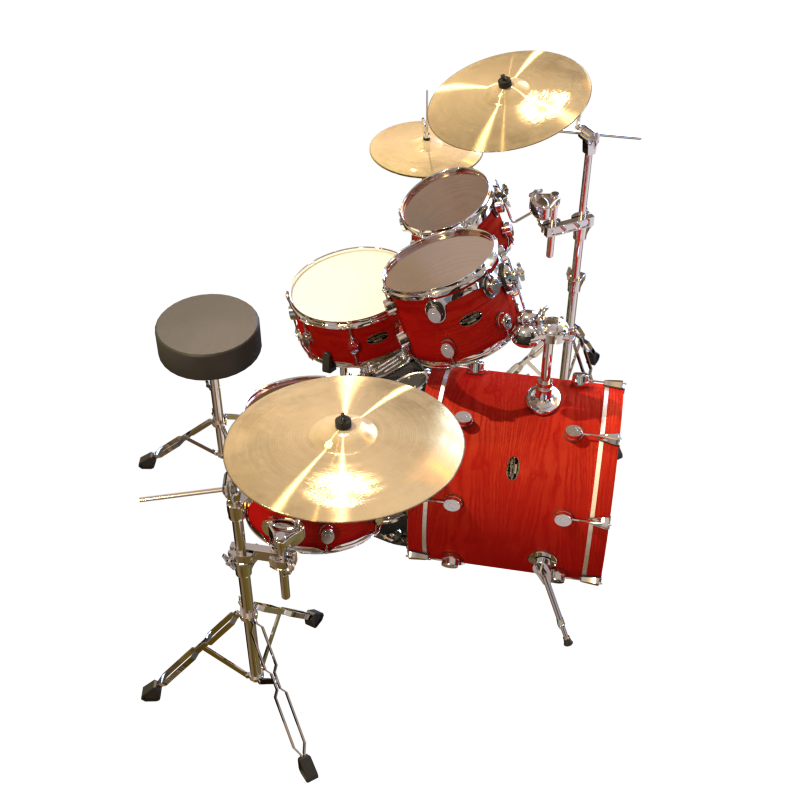}
    \end{subfigure}
    \begin{subfigure}[h]{0.162\linewidth}
        \centering
        \includegraphics[width=\textwidth]{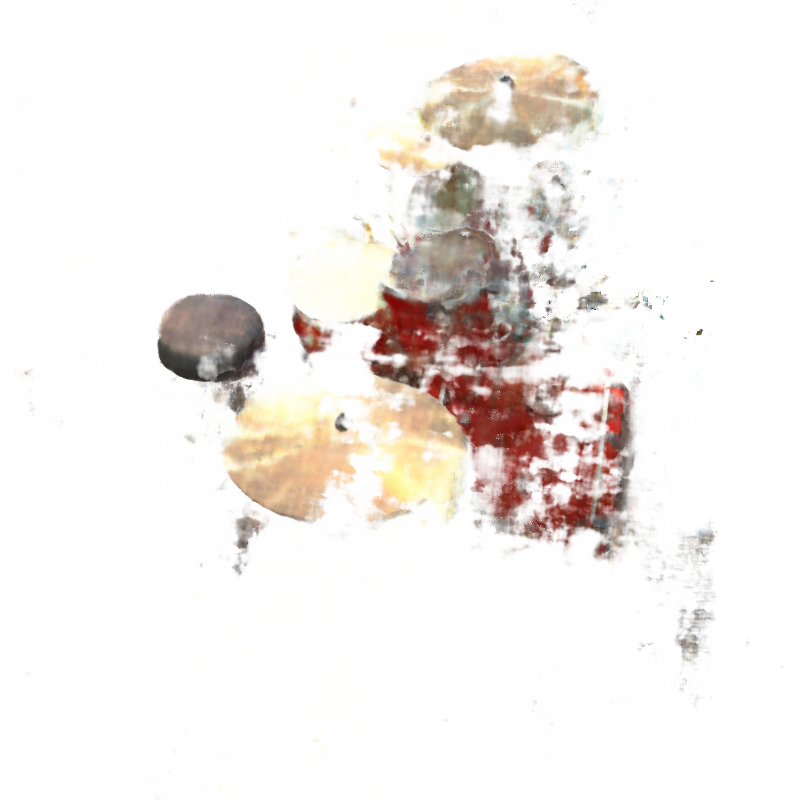}
    \end{subfigure}
    \begin{subfigure}[h]{0.162\linewidth}
        \centering
        \includegraphics[width=\textwidth]{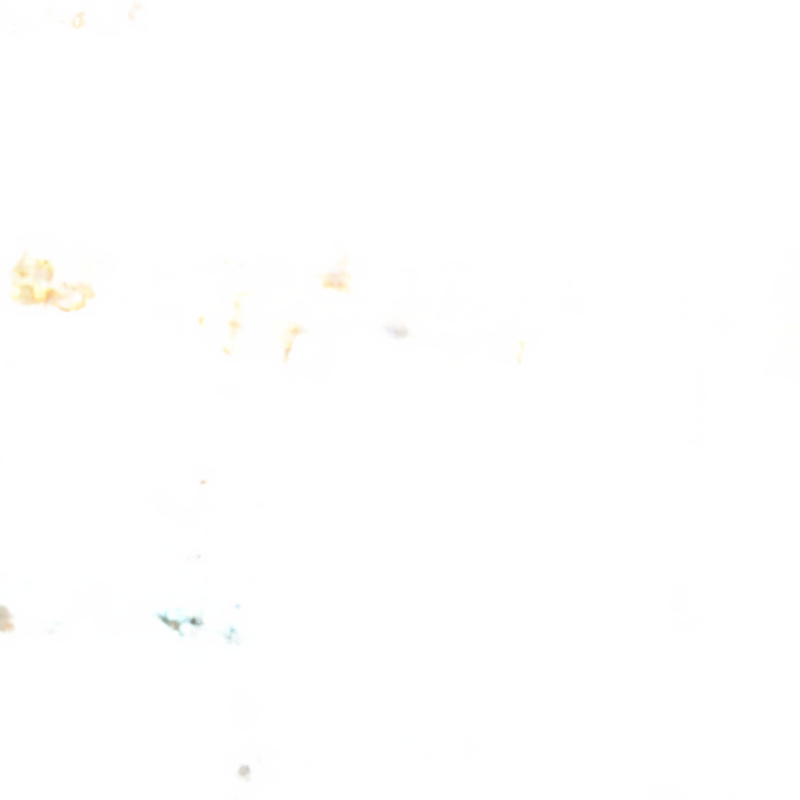}
    \end{subfigure}
    \begin{subfigure}[h]{0.162\linewidth}
        \centering
        \includegraphics[width=\textwidth]{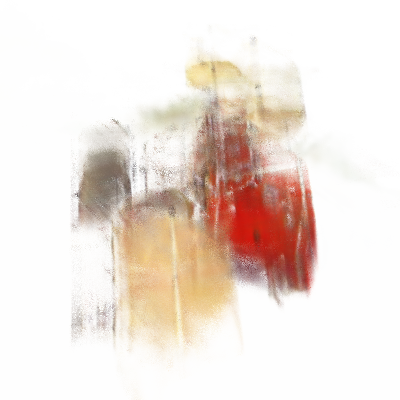}
    \end{subfigure}
    \begin{subfigure}[h]{0.162\linewidth}
        \centering
        \includegraphics[width=\textwidth]{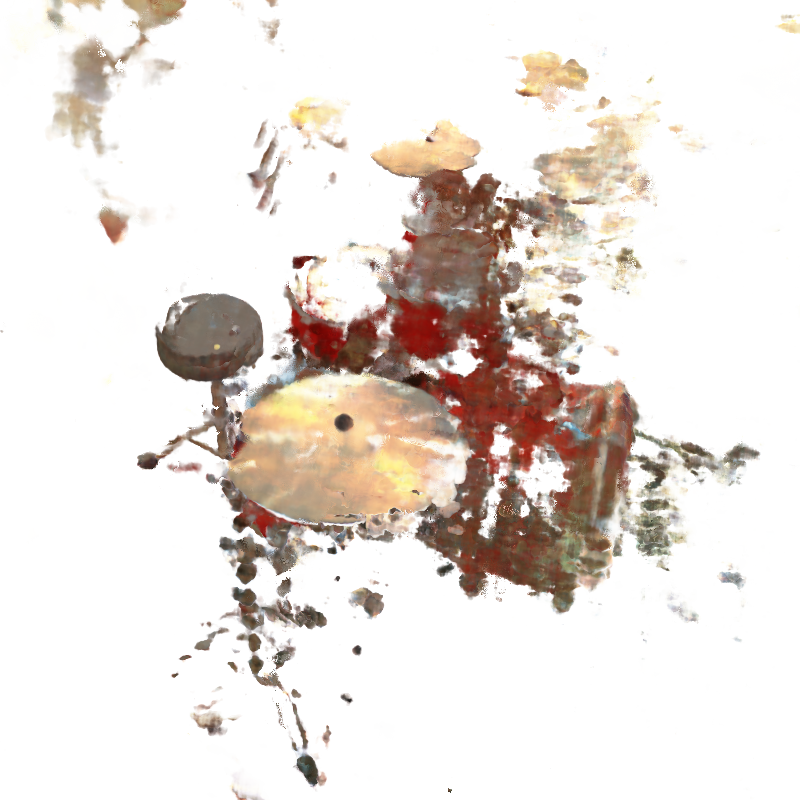}
    \end{subfigure}
        \begin{subfigure}[h]{0.162\linewidth}
        \centering
        \includegraphics[width=\textwidth]{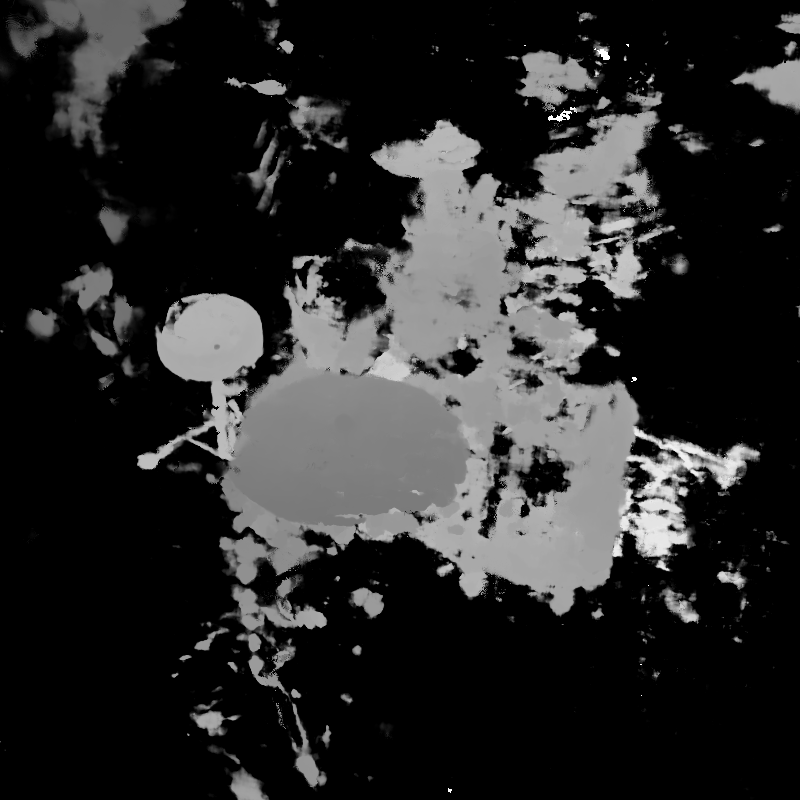}
    \end{subfigure}
    \begin{subfigure}[h]{0.162\linewidth}
        \centering
        \includegraphics[width=\textwidth]{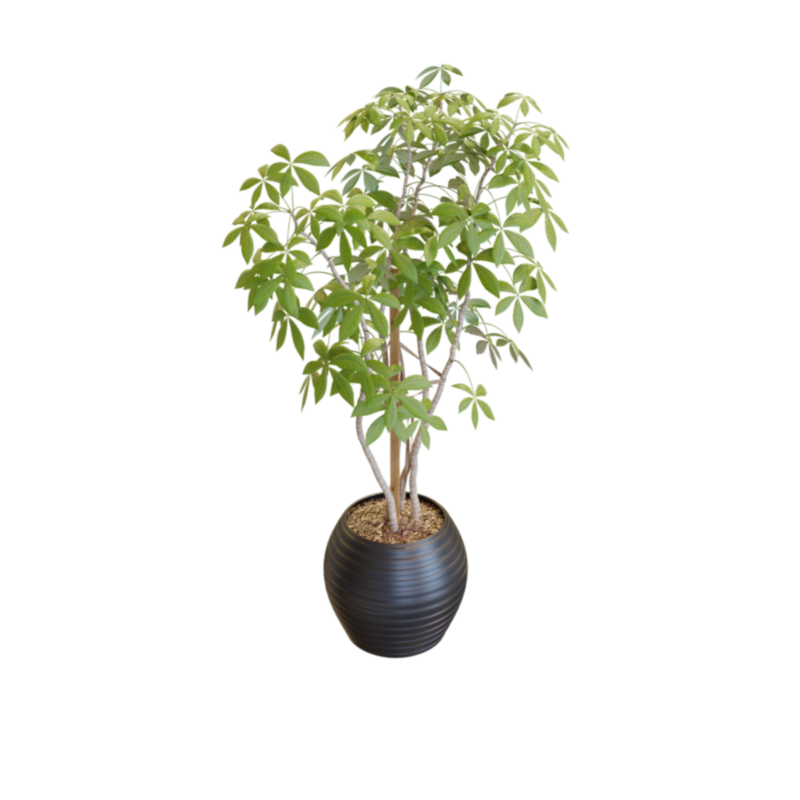}
    \end{subfigure}
    \begin{subfigure}[h]{0.162\linewidth}
        \centering
        \includegraphics[width=\textwidth]{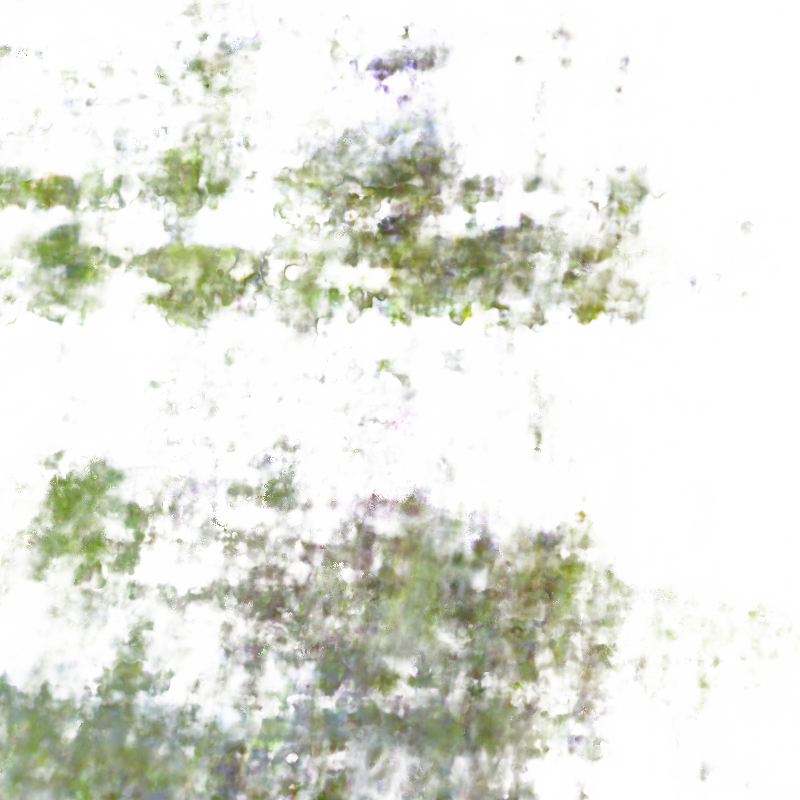}
    \end{subfigure}
    \begin{subfigure}[h]{0.162\linewidth}
        \centering
        \includegraphics[width=\textwidth]{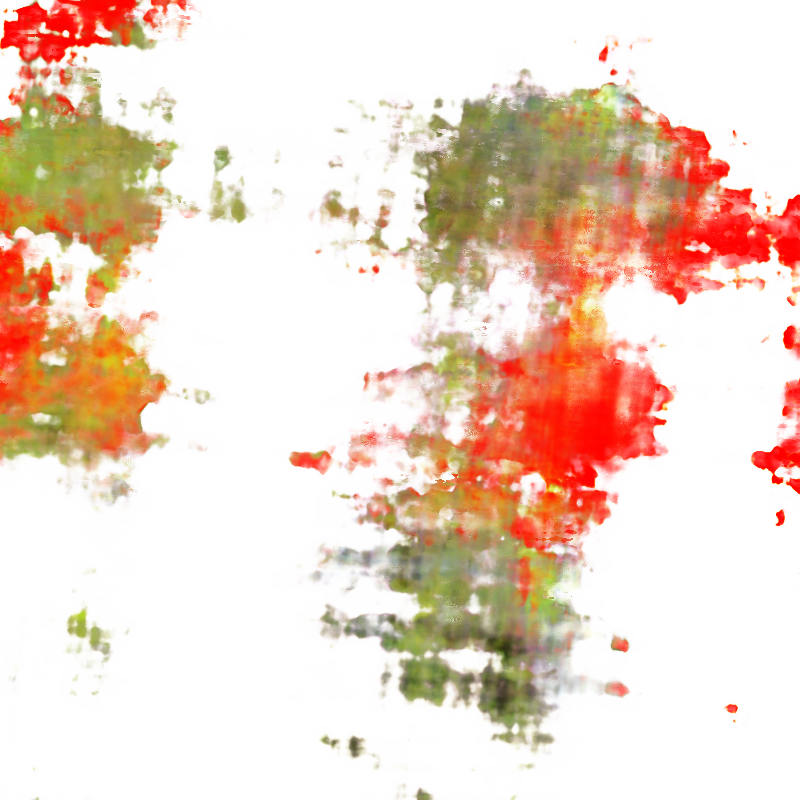}
    \end{subfigure}
    \begin{subfigure}[h]{0.162\linewidth}
        \centering
        \includegraphics[width=\textwidth]{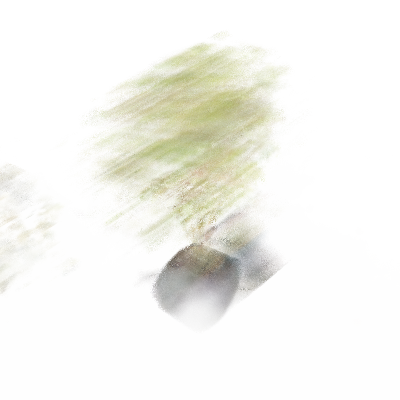}
    \end{subfigure}
    \begin{subfigure}[h]{0.162\linewidth}
        \centering
        \includegraphics[width=\textwidth]{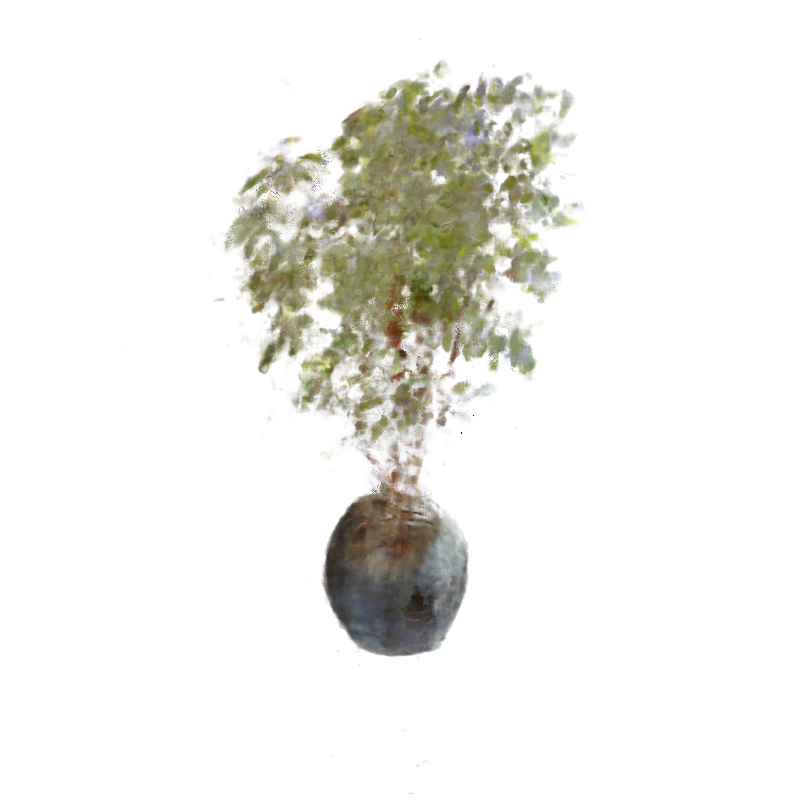}
    \end{subfigure}
    \begin{subfigure}[h]{0.162\linewidth}
        \centering
        \includegraphics[width=\textwidth]{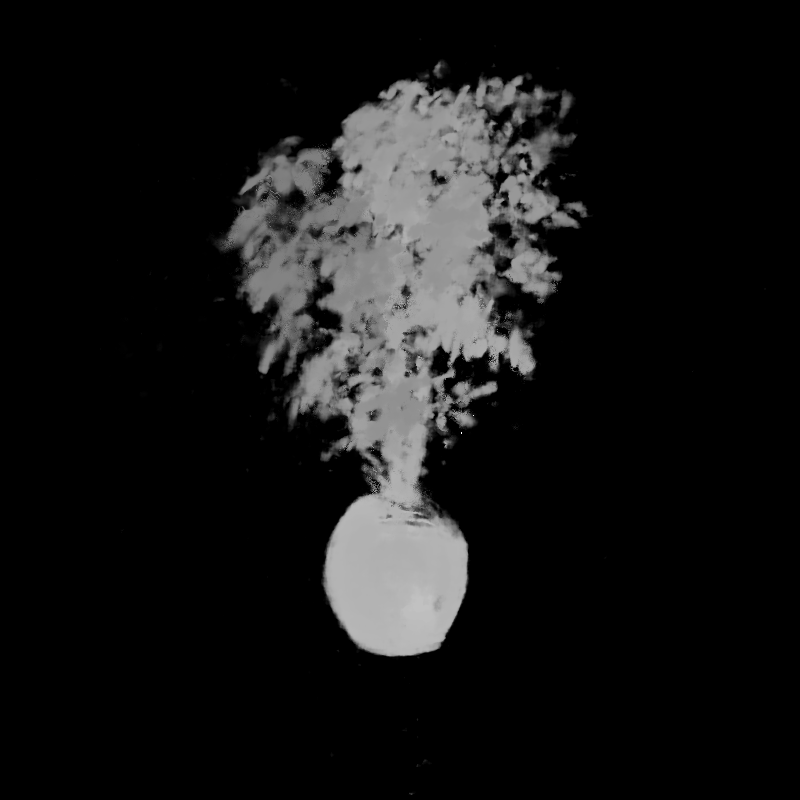}
    \end{subfigure}
           \begin{subfigure}[h]{0.162\linewidth}
        \centering
        \includegraphics[width=\textwidth]{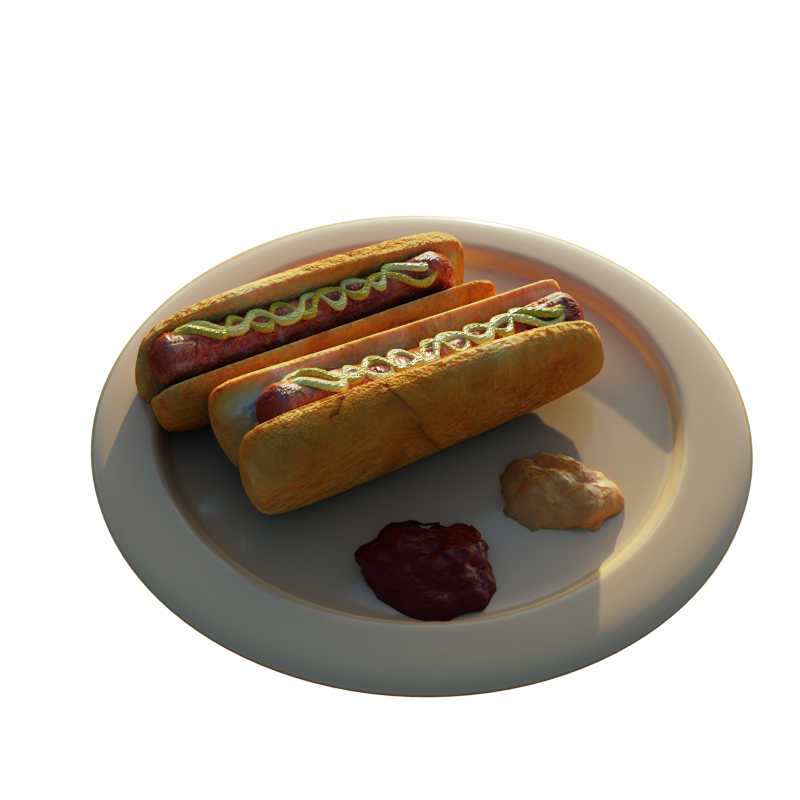}
    \end{subfigure}
    \begin{subfigure}[h]{0.162\linewidth}
        \centering
        \includegraphics[width=\textwidth]{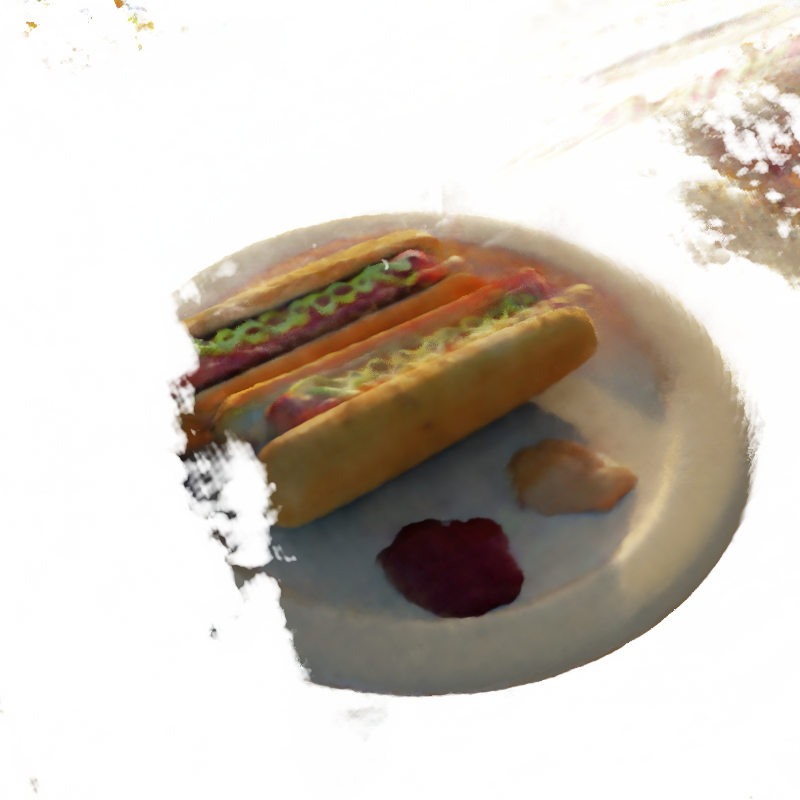}
    \end{subfigure}
    \begin{subfigure}[h]{0.162\linewidth}
        \centering
        \includegraphics[width=\textwidth]{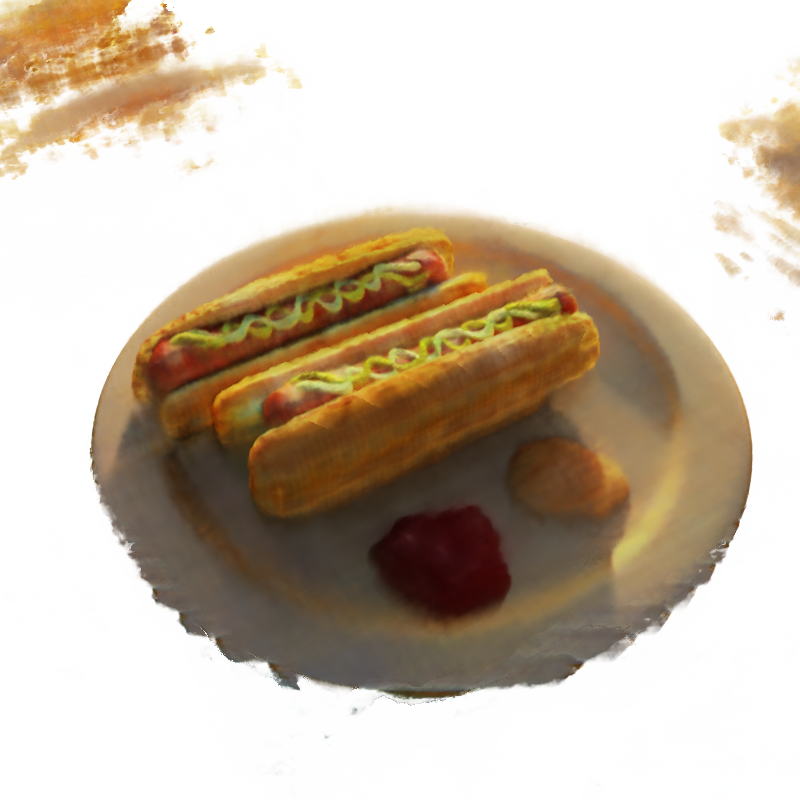}
    \end{subfigure}
    \begin{subfigure}[h]{0.162\linewidth}
        \centering
        \includegraphics[width=\textwidth]{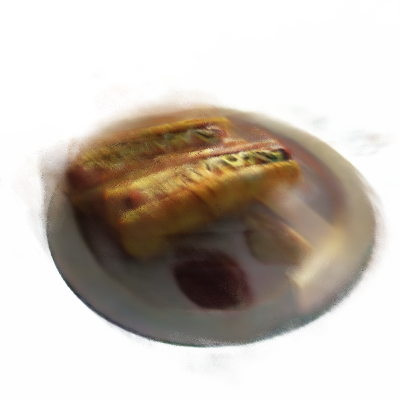}
    \end{subfigure}
    \begin{subfigure}[h]{0.162\linewidth}
        \centering
        \includegraphics[width=\textwidth]{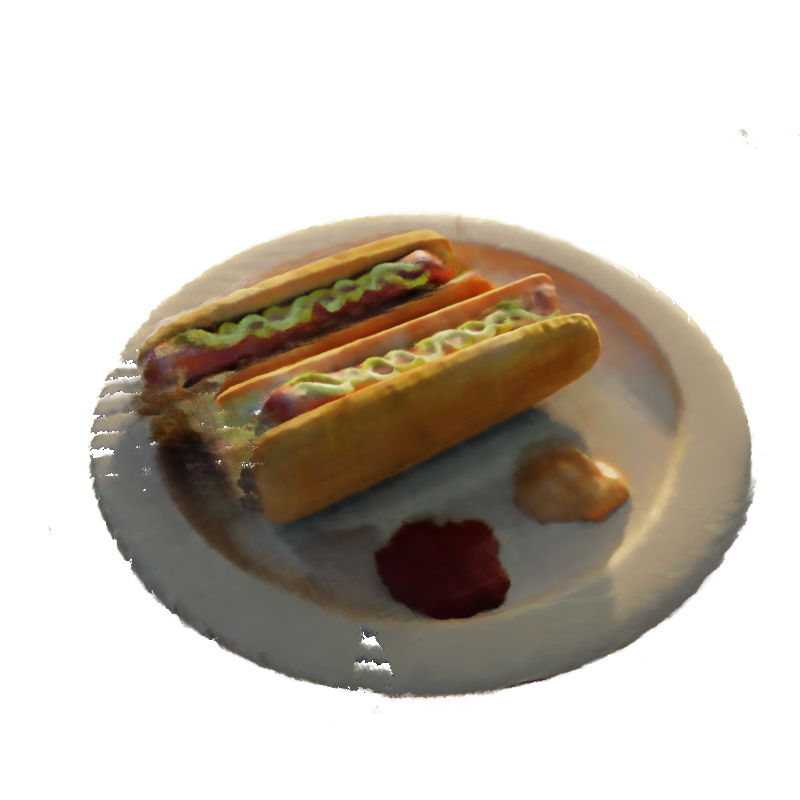}
    \end{subfigure}
    \begin{subfigure}[h]{0.162\linewidth}
        \centering
        \includegraphics[width=\textwidth]{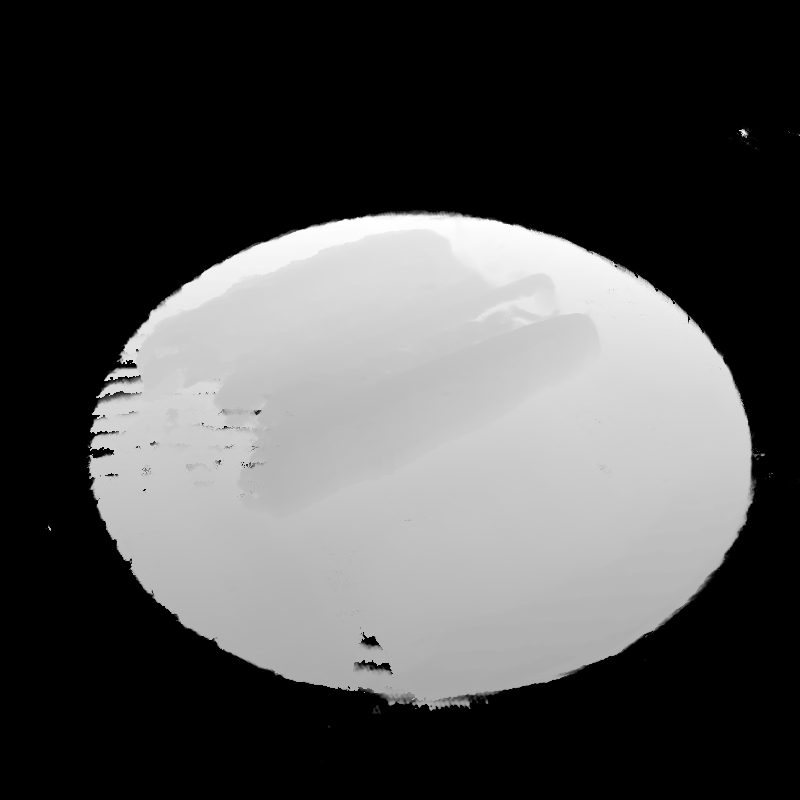}
    \end{subfigure}
            \begin{subfigure}[h]{0.162\linewidth}
        \centering
        \includegraphics[width=\textwidth]{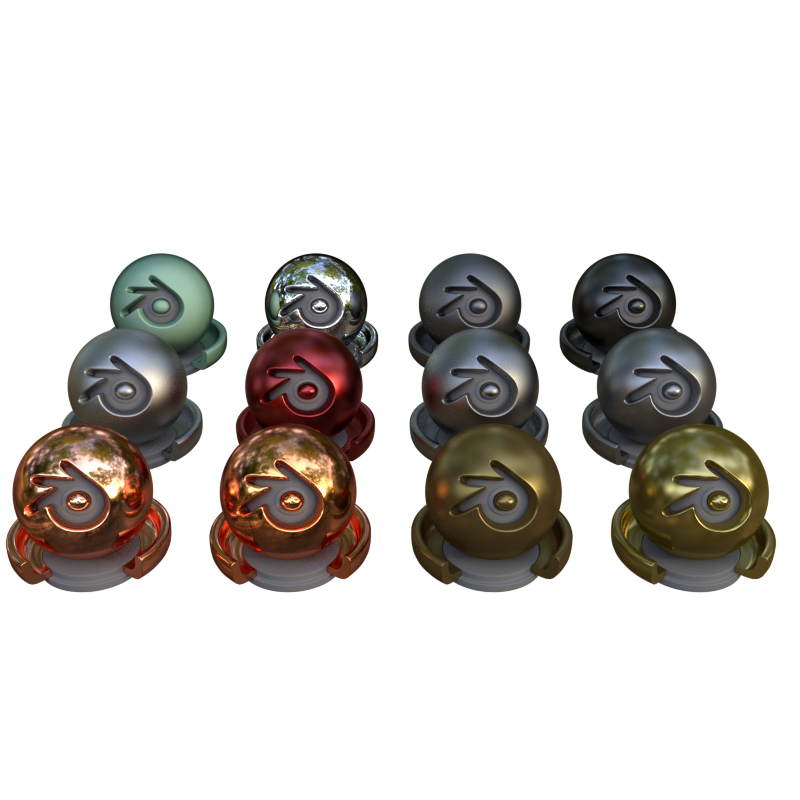}
        \caption{Ground-truth}
    \end{subfigure}
    \begin{subfigure}[h]{0.162\linewidth}
        \centering
        \includegraphics[width=\textwidth]{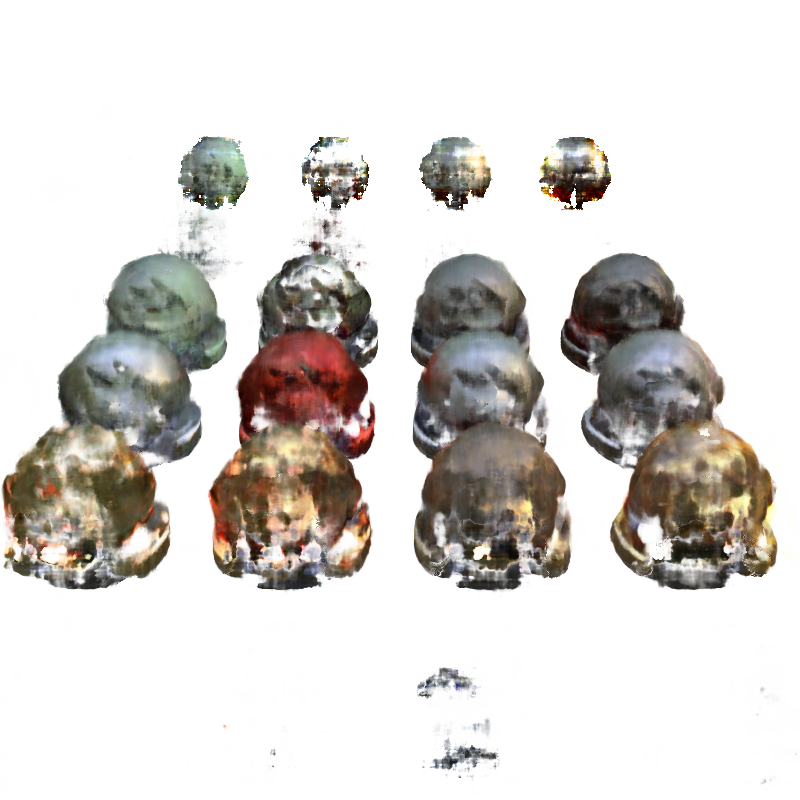}
        \caption{NeRF~\cite{mildenhall2020nerf}}
    \end{subfigure}
    \begin{subfigure}[h]{0.162\linewidth}
        \centering
        \includegraphics[width=\textwidth]{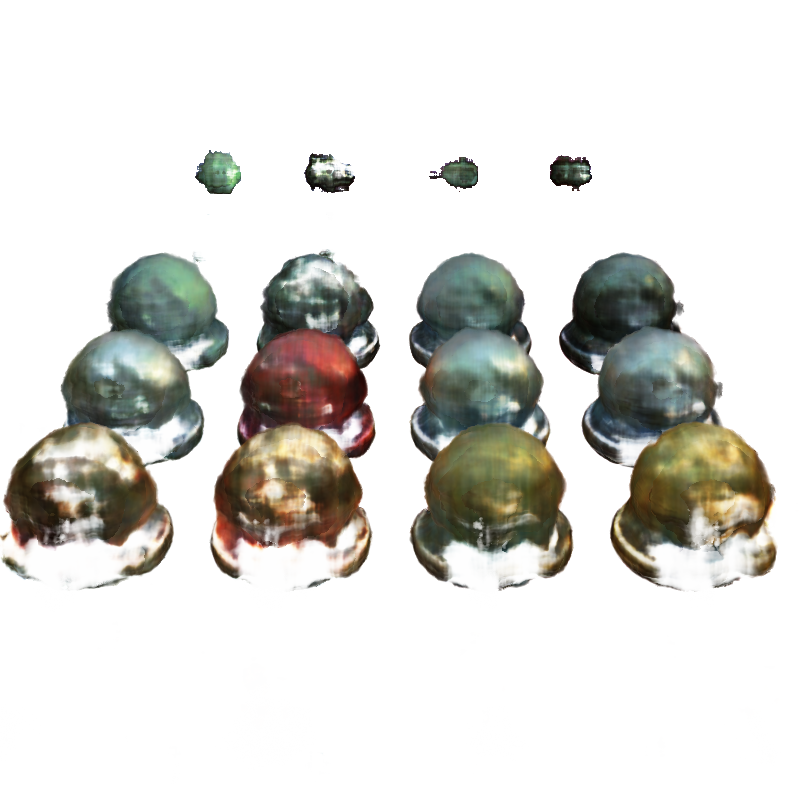}
        \caption{DietNeRF~\cite{jain2021putting}}
    \end{subfigure}
    \begin{subfigure}[h]{0.162\linewidth}
        \centering
        \includegraphics[width=\textwidth]{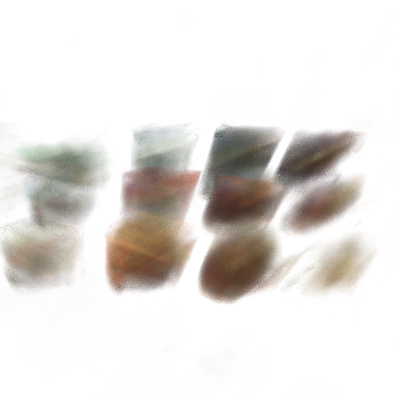}
         \caption{PixelNeRF~\cite{yu2021pixelnerf}}
    \end{subfigure}
    \begin{subfigure}[h]{0.162\linewidth}
        \centering
        \includegraphics[width=\textwidth]{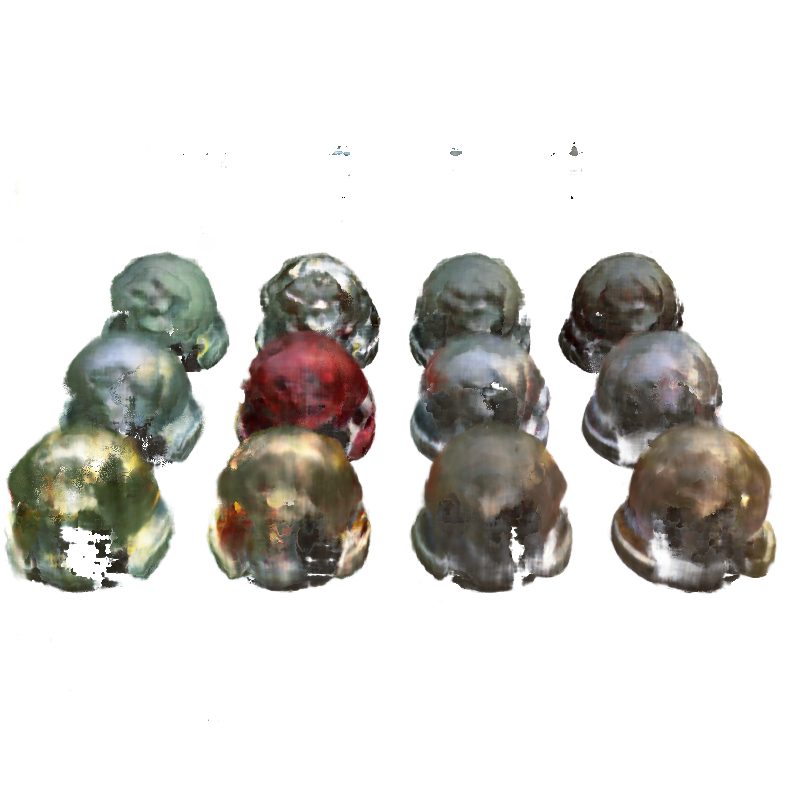}
                        \caption{InfoNeRF}
    \end{subfigure}
        \begin{subfigure}[h]{0.162\linewidth}
        \centering
        \includegraphics[width=\textwidth]{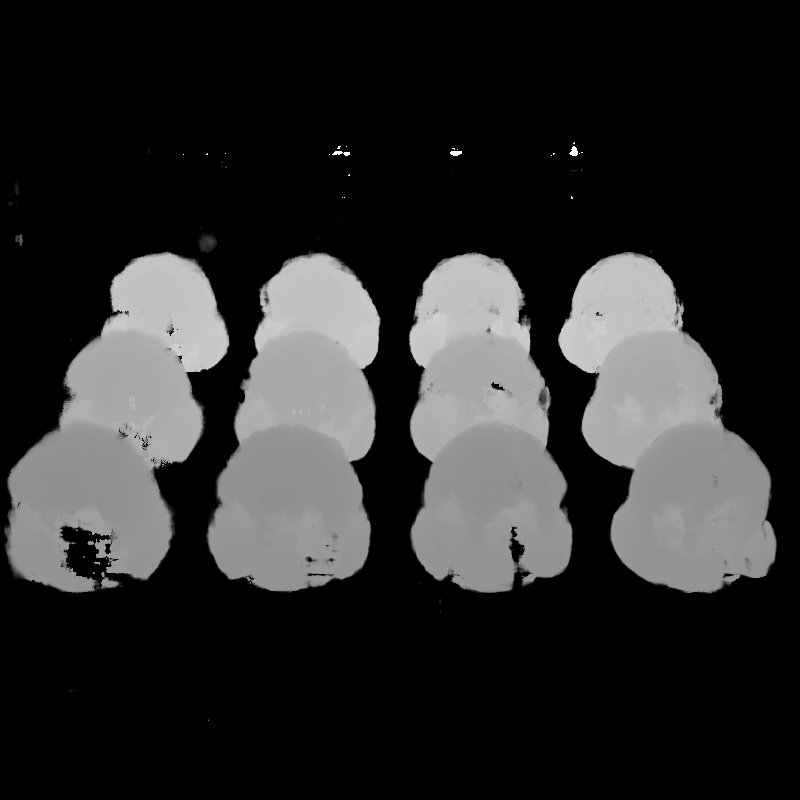}
        \caption{InfoNeRF (depth)}
    \end{subfigure}    
    \caption{
	Qualitative comparison of our method with other NeRF-based models on the \textit{Lego}, \textit{Chair}, \textit{Drums}, \textit{Ficus}, \textit{HotDog} and \textit{Materials} scenes of the Realistic Synthetic 360$^{\circ}$ dataset in the 4-view setting.
    Existing works often suffer from noise (b), color distortion (c), or blur effect (d), while InfoNeRF provides distinguished rendering quality.
    The column (f) visualizes depth maps estimated by InfoNeRF, which provide clear boundaries and fine details.
    }
    \label{fig:qualitiative_synthetic}
\end{figure*}

\begin{figure*}[h]
    \centering
        \begin{subfigure}[h]{0.162\linewidth}
        \centering
        \includegraphics[width=\textwidth]{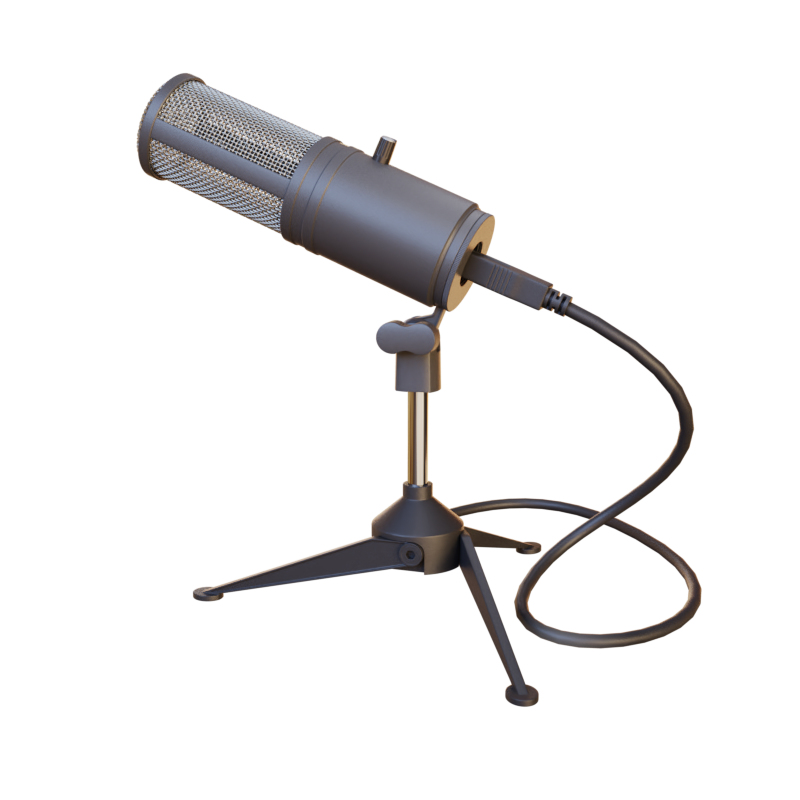}
    \end{subfigure}
    \begin{subfigure}[h]{0.162\linewidth}
        \centering
        \includegraphics[width=\textwidth]{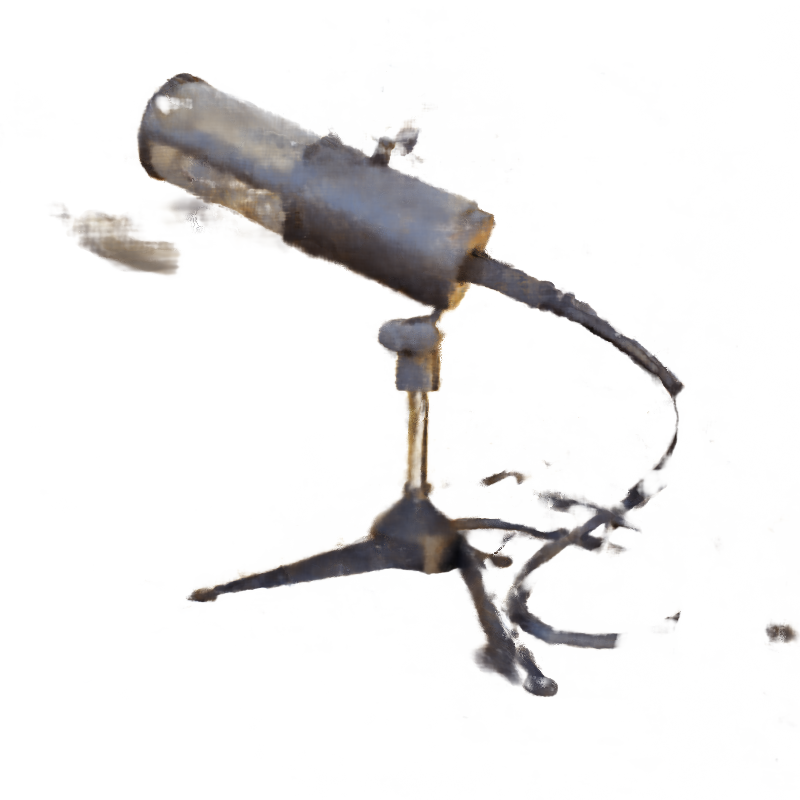}
    \end{subfigure}
    \begin{subfigure}[h]{0.162\linewidth}
        \centering
        \includegraphics[width=\textwidth]{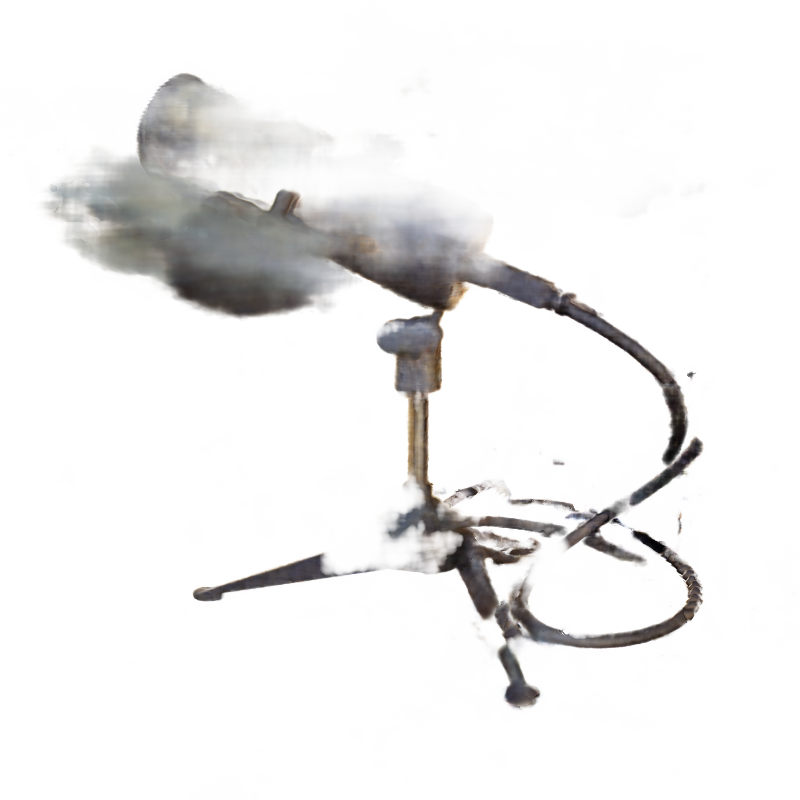}
    \end{subfigure}
    \begin{subfigure}[h]{0.162\linewidth}
        \centering
        \includegraphics[width=\textwidth]{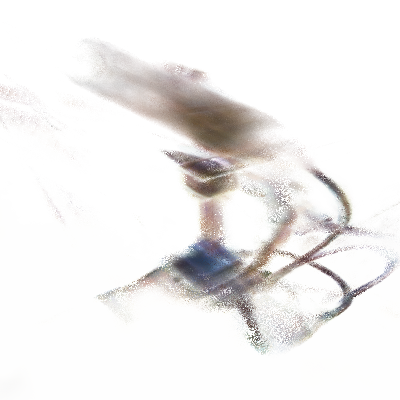}
    \end{subfigure}
    \begin{subfigure}[h]{0.162\linewidth}
        \centering
        \includegraphics[width=\textwidth]{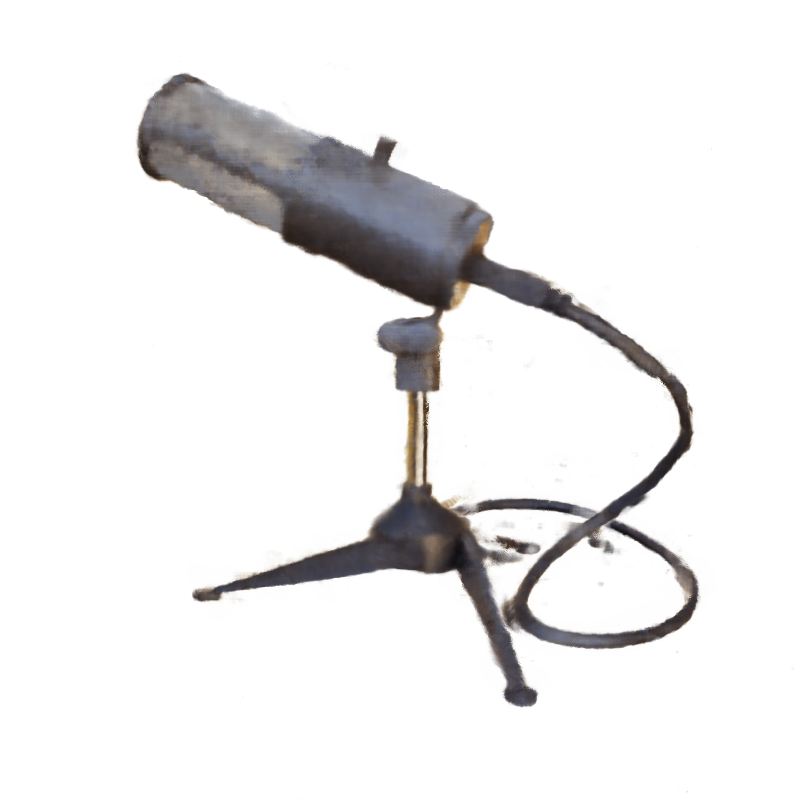}
    \end{subfigure}
    \begin{subfigure}[h]{0.162\linewidth}
        \centering
        \includegraphics[width=\textwidth]{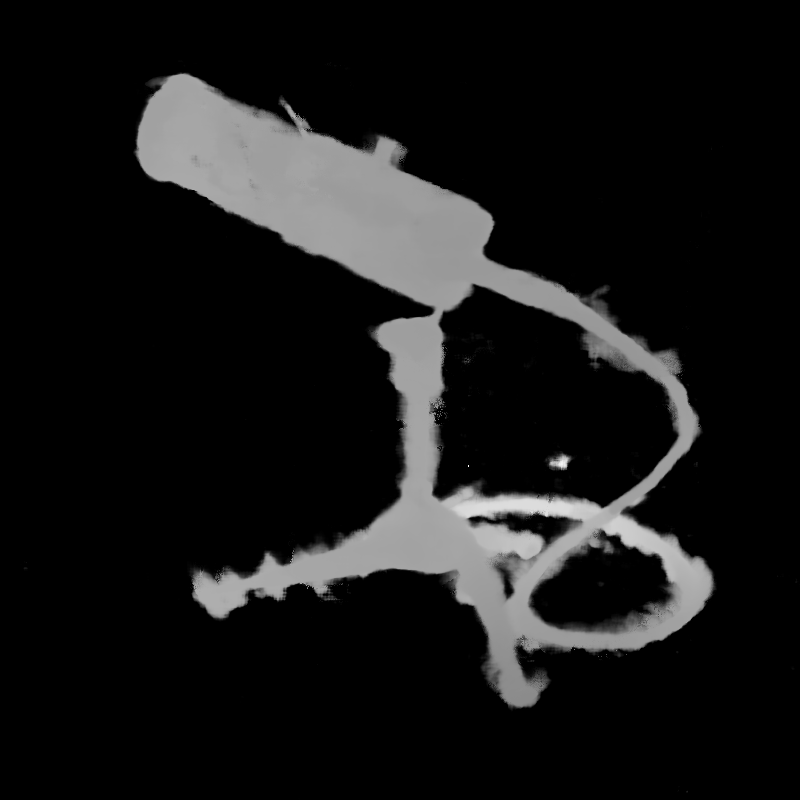}
    \end{subfigure}
    \begin{subfigure}[h]{0.162\linewidth}
        \centering
        \includegraphics[width=\textwidth]{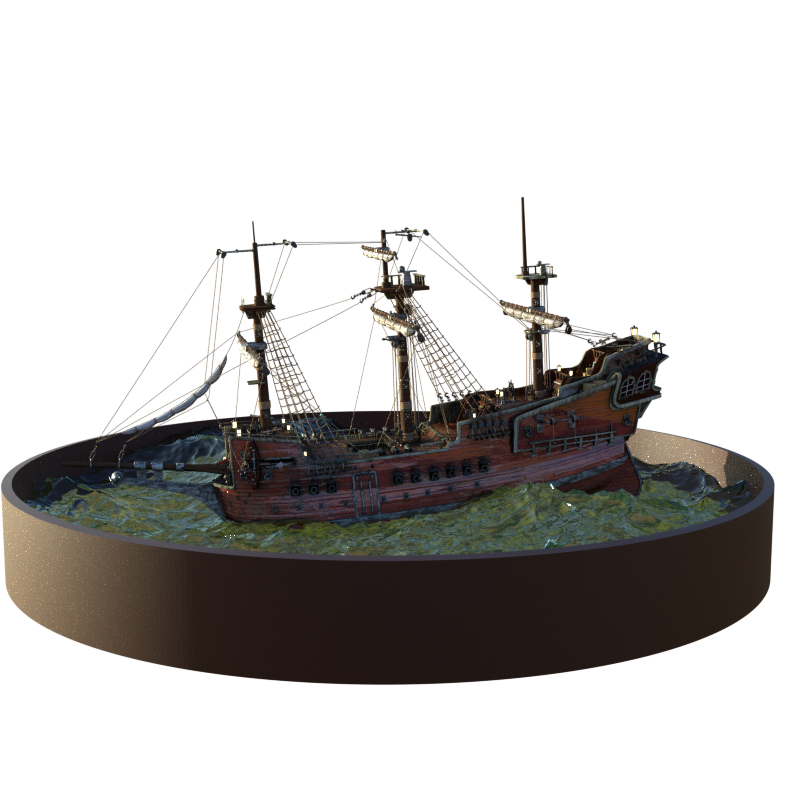}
        \caption{Ground-truth}
    \end{subfigure}
    \begin{subfigure}[h]{0.162\linewidth}
        \centering
        \includegraphics[width=\textwidth]{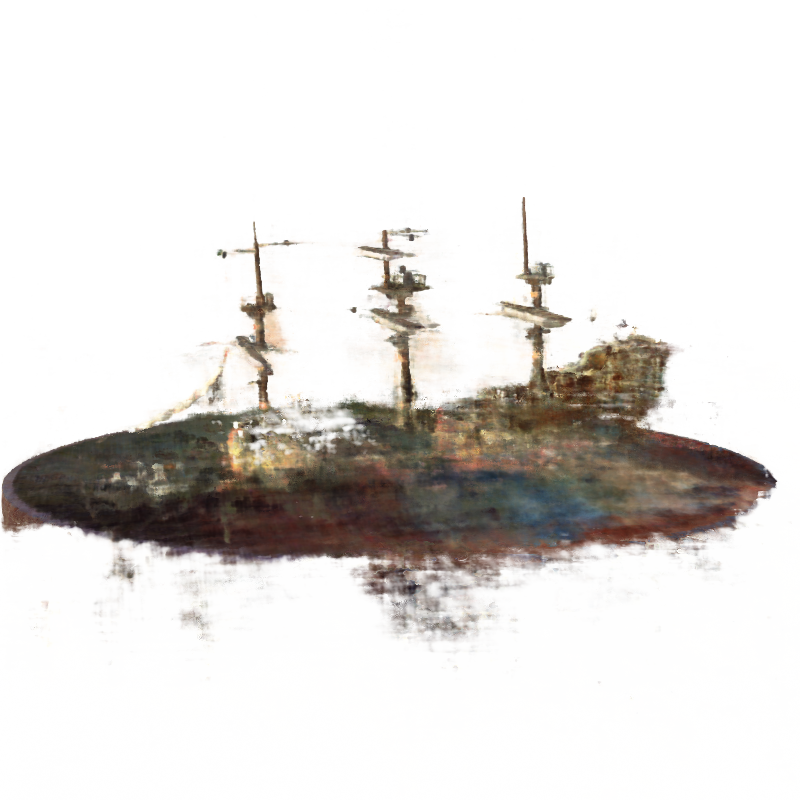}
        \caption{NeRF~\cite{mildenhall2020nerf}}
    \end{subfigure}
    \begin{subfigure}[h]{0.162\linewidth}
        \centering
        \includegraphics[width=\textwidth]{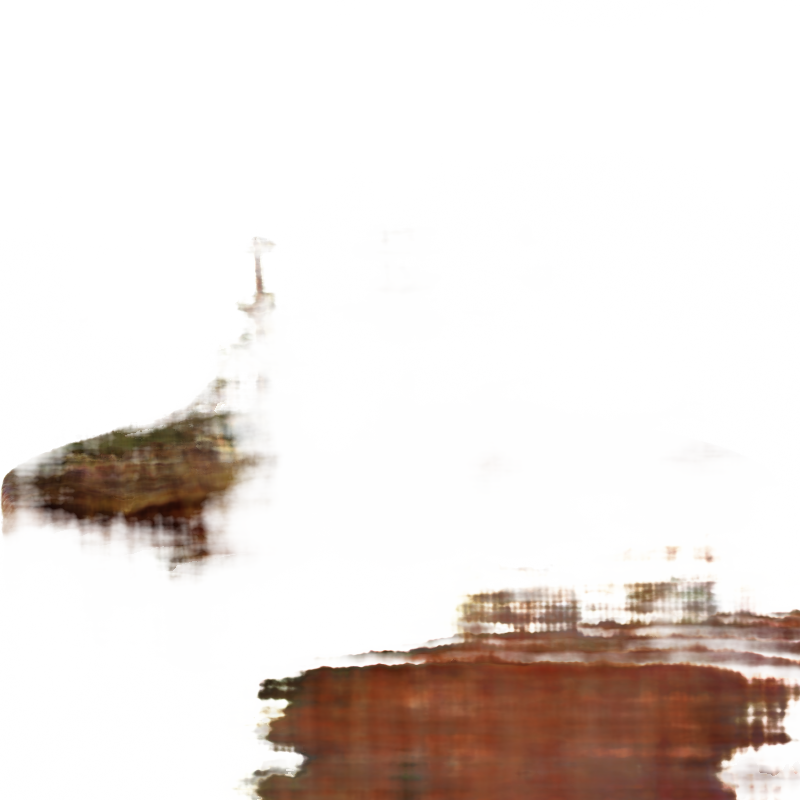}
        \caption{DietNeRF~\cite{jain2021putting}}
    \end{subfigure}
    \begin{subfigure}[h]{0.162\linewidth}
        \centering
        \includegraphics[width=\textwidth]{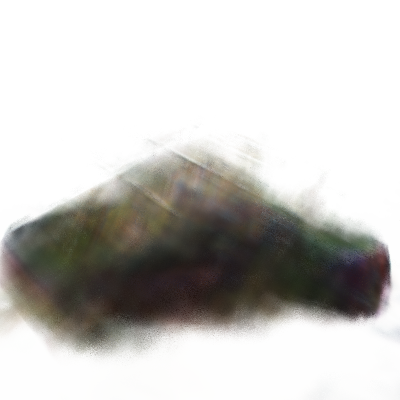}
        \caption{PixelNeRF~\cite{yu2021pixelnerf}}
    \end{subfigure}
    \begin{subfigure}[h]{0.162\linewidth}
        \centering
        \includegraphics[width=\textwidth]{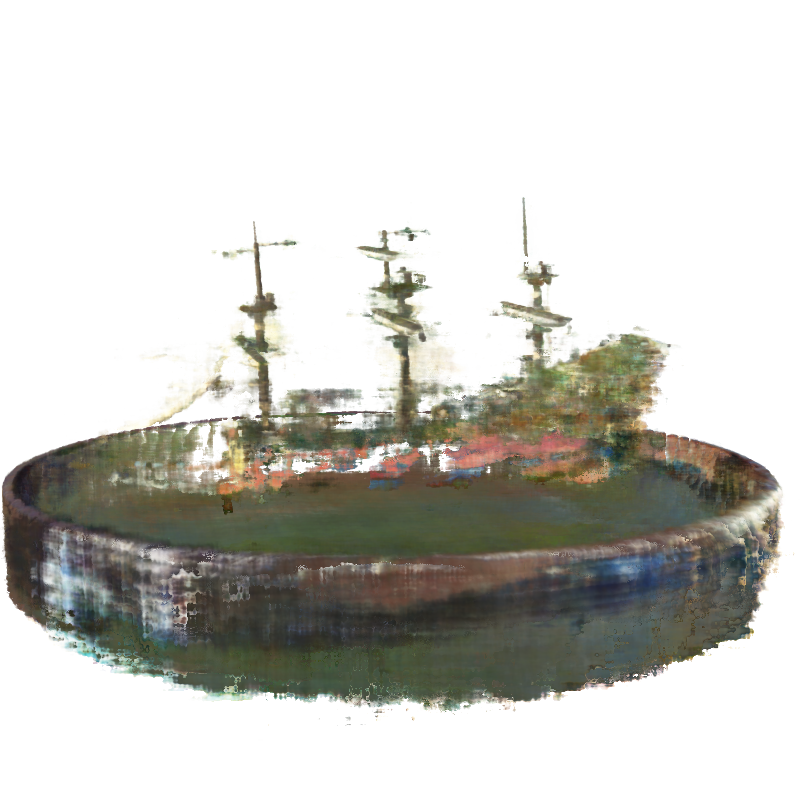}
        \caption{InfoNeRF}
    \end{subfigure}
    \begin{subfigure}[h]{0.162\linewidth}
        \centering
        \includegraphics[width=\textwidth]{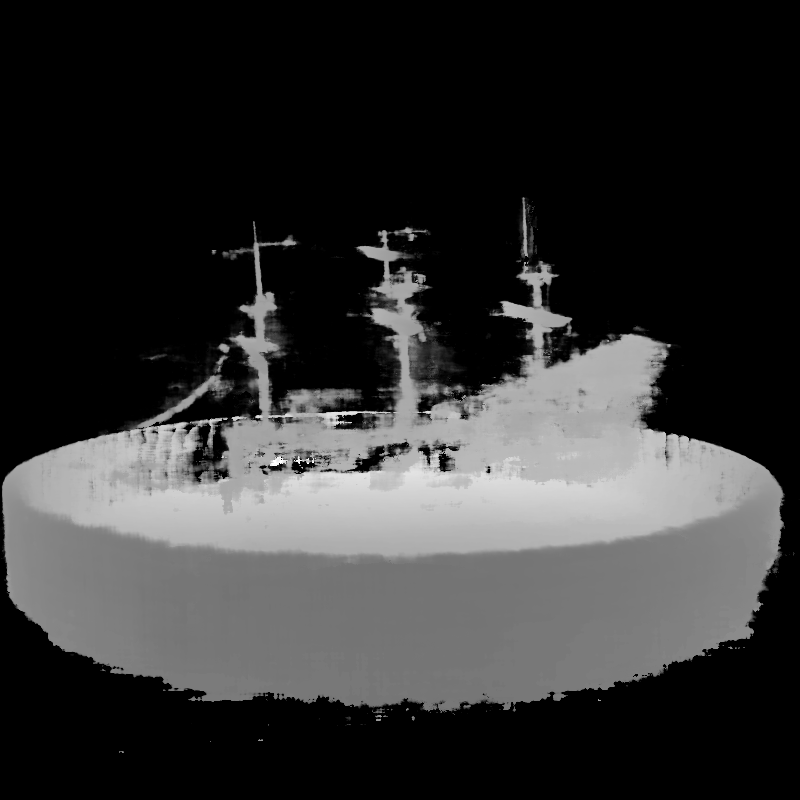}
        \caption{InfoNeRF (depth)}
    \end{subfigure}
    \caption{
	Additional qualitative comparisons of our method with other NeRF-based models on the \textit{Mic} and \textit{Ship} scenes of Realistic Synthetic 360$^{\circ}$ in the 4-view setting.
	As in Figure~\ref{fig:qualitiative_synthetic}, InfoNeRF provides clear boundaries and fine details compared to other works.}
\end{figure*}

\subsection{ZJU-MoCap}
\begin{figure*}[h]
    \centering
        \begin{subfigure}[h]{0.16\linewidth}
         \centering
         \includegraphics[width=\textwidth]{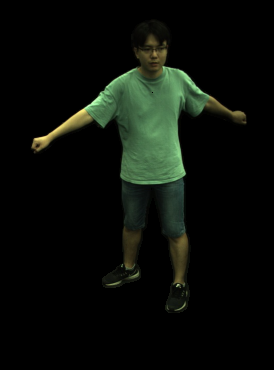}
     \end{subfigure} \hspace{0.3cm}
    \begin{subfigure}[h]{0.16\linewidth}
         \centering
         \includegraphics[width=\textwidth]{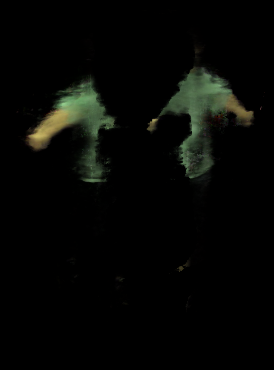}
     \end{subfigure} \hspace{0.3cm}
     \begin{subfigure}[h]{0.16\linewidth}
         \centering
      \includegraphics[width=\textwidth]{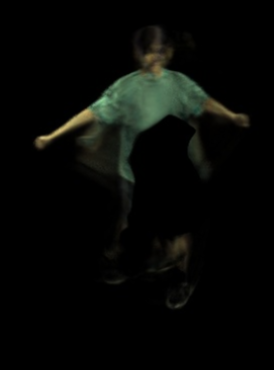}
     \end{subfigure} \hspace{0.3cm}
    \begin{subfigure}[h]{0.16\linewidth}
         \centering
         \includegraphics[width=\textwidth]{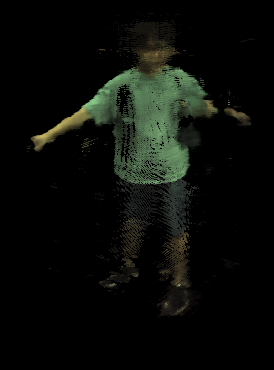}
     \end{subfigure} \hspace{0.3cm}
    \begin{subfigure}[h]{0.16\linewidth}
         \centering
         \includegraphics[width=\textwidth]{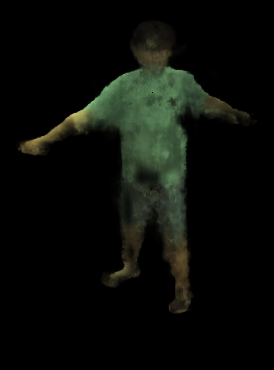}
     \end{subfigure}
    \begin{subfigure}[h]{0.16\linewidth}
         \centering
         \includegraphics[width=\textwidth]{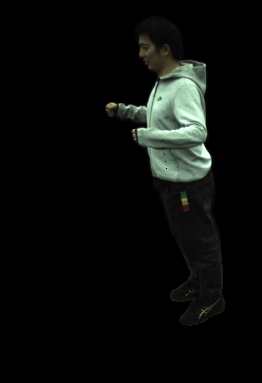}
            \caption{Ground-truth}
     \end{subfigure} \hspace{0.3cm}
    \begin{subfigure}[h]{0.16\linewidth}
         \centering
         \includegraphics[width=\textwidth]{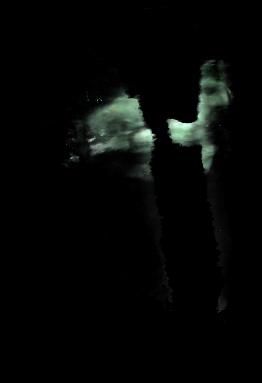}
        \caption{NeRF~\cite{mildenhall2020nerf}}
     \end{subfigure} \hspace{0.3cm}
     \begin{subfigure}[h]{0.16\linewidth}
         \centering
      \includegraphics[width=\textwidth]{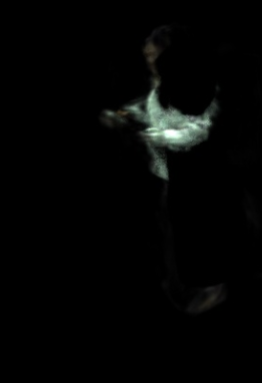}
      \caption{NV~\cite{lombardi2019neural}}
     \end{subfigure} \hspace{0.3cm}
    \begin{subfigure}[h]{0.16\linewidth}
         \centering
         \includegraphics[width=\textwidth]{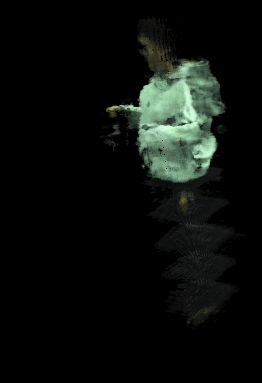}
           \caption{InfoNeRF}
     \end{subfigure} \hspace{0.3cm}
    \begin{subfigure}[h]{0.16\linewidth}
         \centering
         \includegraphics[width=\textwidth]{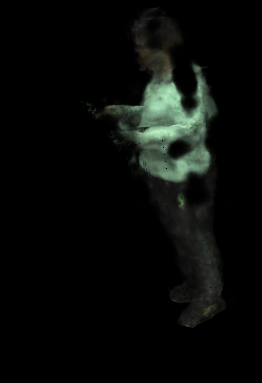}
         \caption{NB~\cite{peng2021neural}}
     \end{subfigure}
        \caption{Qualitative comparison on the ZJU-MoCap dataset in the 4-view setting.
    We visualize the rendering results of prior-free algorithms (b)--(d), including ours, and a prior-based algorithm (e).
    While existing prior-free algorithms, NeRF and NV, often suffer from inconsistent reconstructions and missing parts of the human body, InfoNeRF render most human bodies with competitive quality to the prior-based algorithm, NB.}
        \label{fig:mocap_supple}
\end{figure*}

\section{Potential Negative Societal Impact \& Limitations}
The proposed few-shot view synthesis algorithm works well with a very small number of views, and it is more vulnerable to adversarial attacks, which is problematic in real-world scenarios, \eg, AR or VR systems.
InfoNeRF shows outstanding results on the few-shot volume rendering, but is still struggling from the need for calibrated cameras albeit few in number, which we leave as our future work.

\end{document}